\documentclass[10pt,twocolumn,letterpaper]{article}

\usepackage[pagenumbers]{cvpr}  

\usepackage{tabularx}
\usepackage{multirow}
\usepackage[per-mode=symbol, group-digits=integer, group-minimum-digits = 4, detect-all=true, input-symbols={()}]{siunitx}
\DeclareSIUnit{\nothing}{\relax}

\usepackage[percent]{overpic} 
\usepackage{fp}
\usepackage{graphicx}
\usepackage{tikz}
\usetikzlibrary{positioning}
\usetikzlibrary{tikzmark, positioning}
\usetikzlibrary{shapes.geometric, shapes, arrows, arrows.meta, bending}
\usetikzlibrary{backgrounds}
\usetikzlibrary{calc}
\usepackage{algorithm}
\usepackage{algpseudocode}
\usetikzlibrary{spy}
\usepackage{pgfplots}
\usepackage{bbding}
\usepackage{dsfont}
\usepackage{fontawesome5}
\usepackage{calc}
\pgfplotsset{compat=1.17}
\usepackage[outline]{contour} 
\contourlength{0.02em}
\usepackage[table]{xcolor}
\usepackage{makecell}
\usepackage{adjustbox}
\usepackage{pifont}
\newcommand{\xmark}{\ding{55}}%
\usepackage[scaled=0.97]{newtxtt}
\newcommand\ours{MARCO}
\newcommand\pckTen{$\text{PCK@}0.10$}
\newcommand\pckFive{$\text{PCK@}0.05$}
\newcommand\pckOne{$\text{PCK@}0.01$}
\newcommand\pckFifteen{$\text{PCK@}0.15$}

\newcommand*{\inparagraph}[1]{\smallskip\noindent\textbf{#1}\hspace{0.4em}}

\newcommand{\unc}[1]{\hspace{1pt}{\scriptsize\textcolor{black!60}{$\pm$\,#1}}}

\usepackage{utfsym}
\usepackage[super]{nth}

\usepackage{setspace}
\definecolor{cvprcolor}{RGB}{127,127,255}
\definecolor{softrow}{RGB}{245,245,245}

\definecolor{tud0d}{RGB}{83,83,83}
\definecolor{tud0c}{RGB}{137,137,137}
\definecolor{tud0b}{RGB}{181,181,181}
\definecolor{tud0a}{RGB}{220,220,220}
\definecolor{tud1a}{RGB}{93,133,195}
\definecolor{tud2a}{RGB}{0,156,218}
\definecolor{tud3a}{RGB}{80,182,149}
\definecolor{tud4a}{RGB}{175,204,80}
\definecolor{tud5a}{RGB}{221,223,72}
\definecolor{tud6a}{RGB}{255,224,92}
\definecolor{tud7a}{RGB}{248,186,60}
\definecolor{tud8a}{RGB}{238,122,52}
\definecolor{tud9a}{RGB}{233,80,62}
\definecolor{tud10a}{RGB}{201,48,142}
\definecolor{tud11a}{RGB}{128,69,151}
\definecolor{tud1b}{RGB}{0,90,169}
\definecolor{tud2b}{RGB}{0,131,204}
\definecolor{tud3b}{RGB}{0,157,129}
\definecolor{tud4b}{RGB}{153,192,0}
\definecolor{tud5b}{RGB}{201,212,0}
\definecolor{tud6b}{RGB}{253,202,0}
\definecolor{tud7b}{RGB}{245,163,0}
\definecolor{tud8b}{RGB}{236,101,0}
\definecolor{tud9b}{RGB}{230,0,26}
\definecolor{tud10b}{RGB}{166,0,132}
\definecolor{tud11b}{RGB}{114,16,133}
\definecolor{tud1c}{RGB}{0,78,138}
\definecolor{tud2c}{RGB}{0,104,157}
\definecolor{tud3c}{RGB}{0,136,119}
\definecolor{tud4c}{RGB}{127,171,22}
\definecolor{tud5c}{RGB}{177,189,0}
\definecolor{tud6c}{RGB}{215,172,0}
\definecolor{tud7c}{RGB}{210,135,0}
\definecolor{tud8c}{RGB}{204,76,3}
\definecolor{tud9c}{RGB}{185,15,34}
\definecolor{tud10c}{RGB}{149,17,105}
\definecolor{tud11c}{RGB}{97,28,115}
\definecolor{tud1d}{RGB}{36,53,114}
\definecolor{tud2d}{RGB}{0,78,115}
\definecolor{tud3d}{RGB}{0,113,94}
\definecolor{tud4d}{RGB}{106,139,55}
\definecolor{tud5d}{RGB}{153,166,4}
\definecolor{tud6d}{RGB}{174,142,0}
\definecolor{tud7d}{RGB}{190,111,0}
\definecolor{tud8d}{RGB}{169,73,19}
\definecolor{tud9d}{RGB}{156,28,38}
\definecolor{tud10d}{RGB}{115,32,84}
\definecolor{tud11d}{RGB}{76,34,106}

\usepackage{arydshln}  
\usepackage{booktabs}
\DeclareMathOperator*{\SoftArgmax}{\text{\small soft-argmax}}
\DeclareMathOperator*{\softmax}{softmax}

\newcommand{\Cow}[1][]{\includegraphics[width=10pt,trim={6cm 9cm 5cm 6cm},clip]{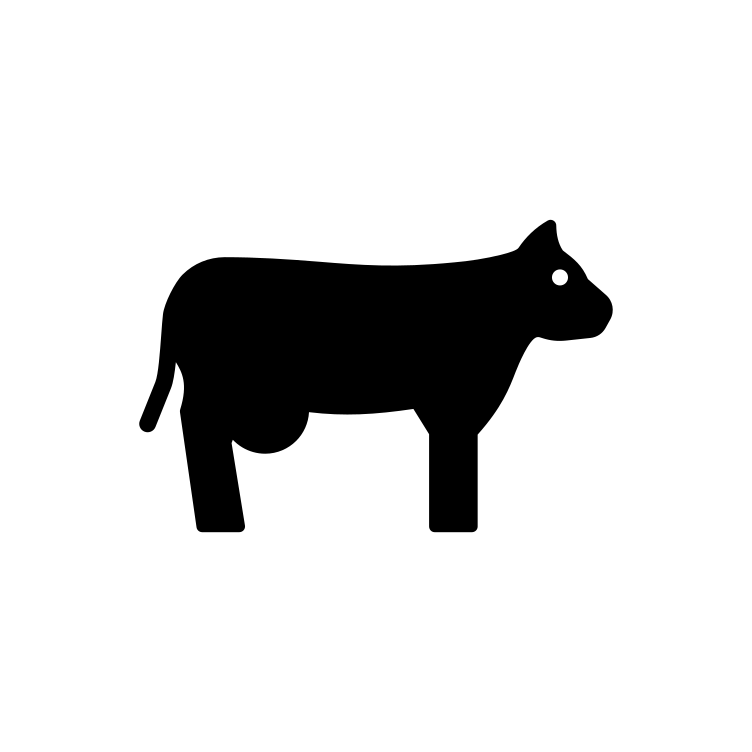}}
\newcommand{\Plant}[1][]{\includegraphics[width=10pt,trim={7cm 6cm 5cm 2cm},clip]{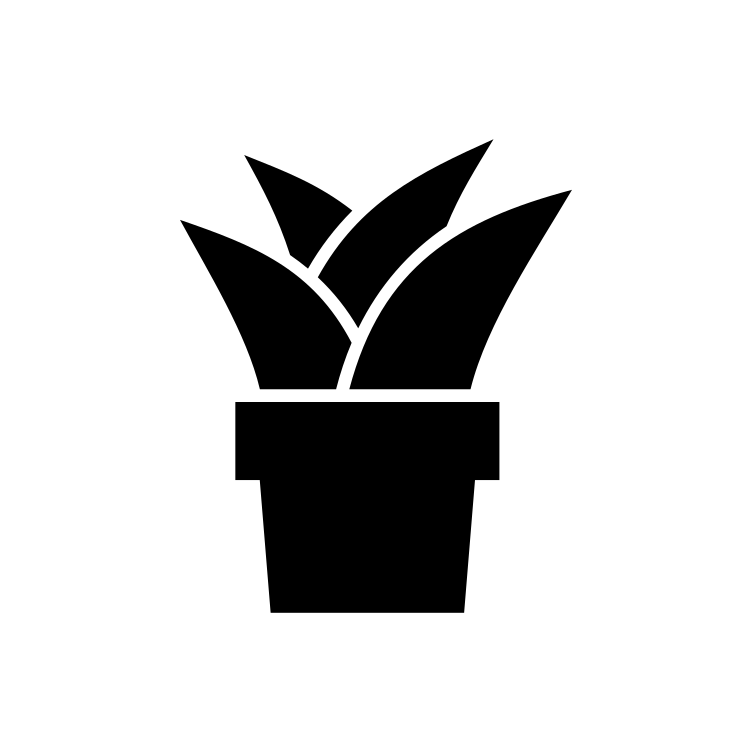}}
\newcommand{\Sheep}[1][]{\includegraphics[width=10pt,trim={6cm 7cm 5cm 6cm},clip]{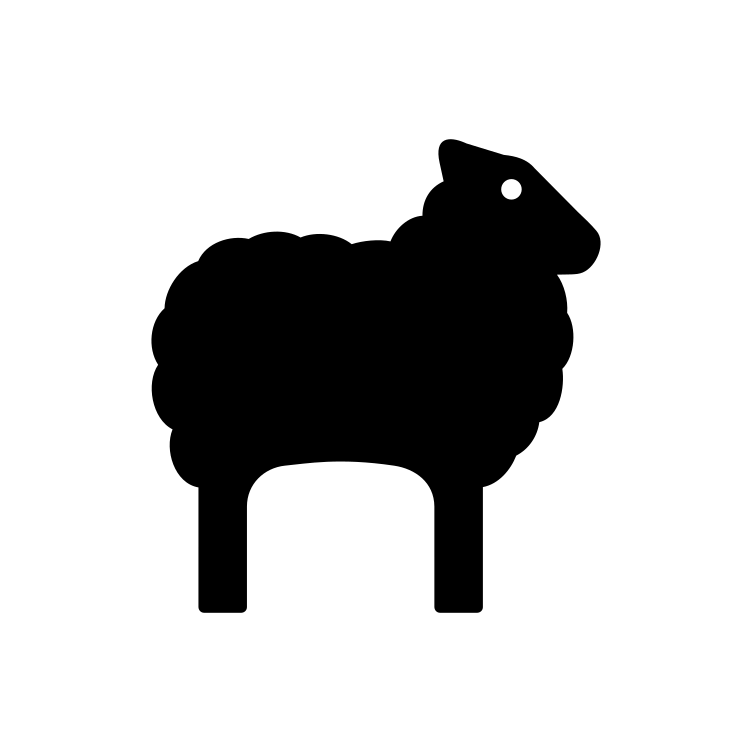}}

\newcommand{\HumanFace}[1][]{%
  \includegraphics[width=10pt]{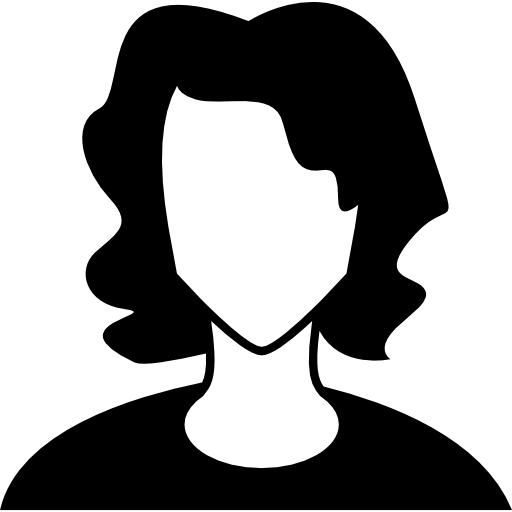}%
}
\newcommand{\Dress}[1][]{%
  \includegraphics[width=10pt]{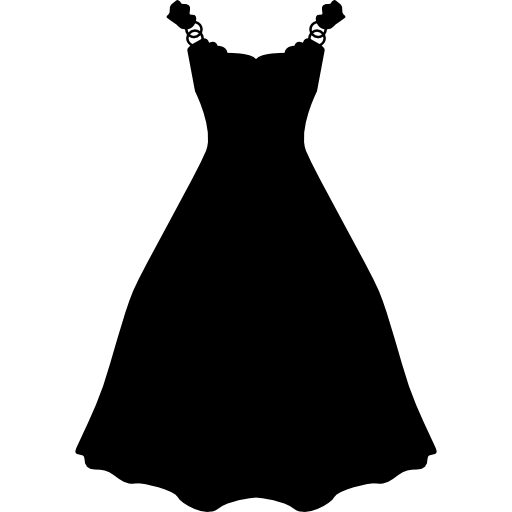}%
}

\newcommand{\TableIcon}[1][]{%
  \includegraphics[width=10pt]{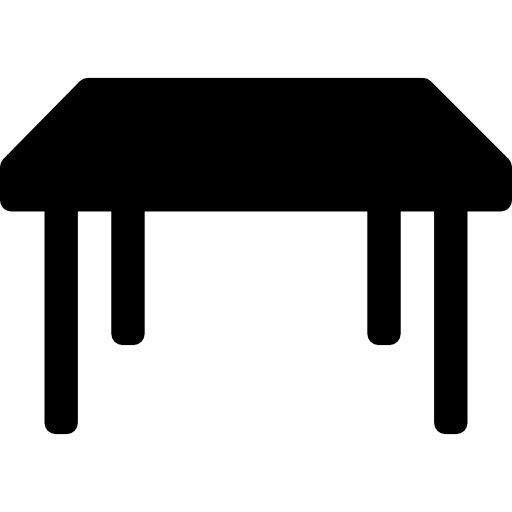}%
}
\newcommand{\Elephant}[1][]{%
  \includegraphics[width=10pt]{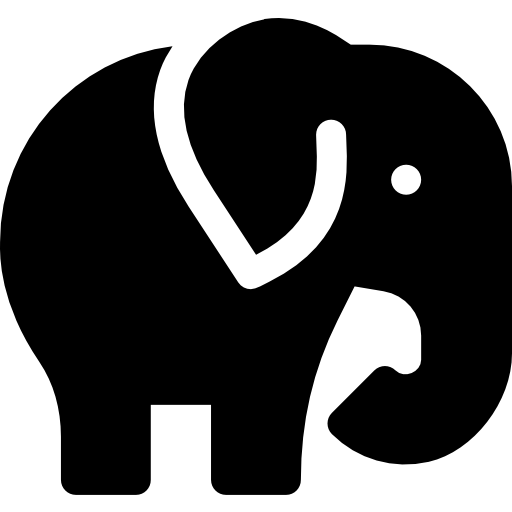}%
}

\newcommand{\AnimalFace}[1][]{%
  \includegraphics[width=10pt]{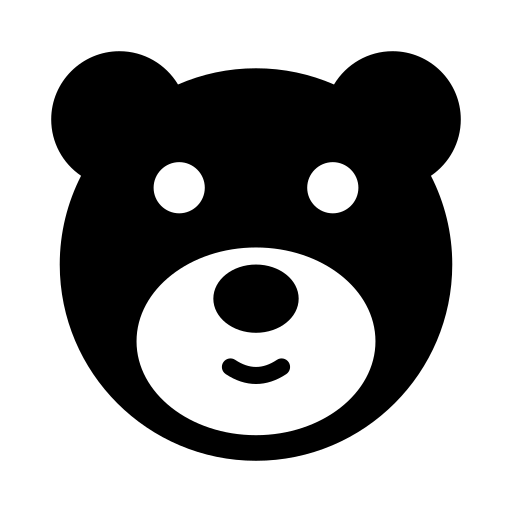}%
}

\newcommand{\tablesize}{\fontsize{8.2pt}{10pt}\selectfont}

\definecolor{cvprblue}{rgb}{0.21,0.49,0.74}
\usepackage[pagebackref,breaklinks,colorlinks,allcolors=cvprblue]{hyperref}

\usepackage{pifont}
\usepackage[table]{xcolor}
\usepackage{subcaption} 
\usepackage{tikz}
\usetikzlibrary{calc}

\usepackage{booktabs}   
\usepackage{array}      
\usepackage{makecell}   
\usepackage{multirow}
\usepackage{graphicx} 
\usepackage{textcomp} 
\usepackage{colortbl}

\usepackage{fontawesome5} 
\usepackage[normalem]{ulem}
\usepackage{xspace}  

\def\confName{CVPR}
\def\confYear{2026}

\title{MARCO: Navigating the Unseen Space of Semantic Correspondence}

\newcommand{\myparagraph}[1]{\smallskip\noindent\textbf{#1}\hspace{0.4em}}

\newcommand{\authorstep}{\hspace{0.75cm}}
\newcommand{\affiliationstep}{\hspace{0.95cm}}

\author{
Claudia Cuttano\textsuperscript{\normalfont{}\,1,2}
\authorstep Gabriele Trivigno\textsuperscript{\normalfont{}\,1}
\authorstep Carlo Masone\textsuperscript{\normalfont{}\,1}
\authorstep Stefan Roth\textsuperscript{\normalfont{}\,2,3,4}\\[0pt]
\small{\textsuperscript{1}Politecnico di Torino\affiliationstep \textsuperscript{2}TU Darmstadt
\affiliationstep \textsuperscript{3}hessian.AI\affiliationstep \textsuperscript{4}ELIZA}\\[-2pt]\small {\url{https://visinf.github.io/MARCO}}}

\usepackage{fancyhdr}
\usepackage{setspace}

\fancypagestyle{firststyle}
{

	\fancyhf{}
	\lfoot{{\footnotesize\begin{spacing}{.5}\parbox{\linewidth}{\vspace{-0.5em}%
					To appear in \emph{Proceedings of the IEEE/CVF Conference on Computer Vision and Pattern Recognition}, Denver, CO, USA, 2026.%
					\\\hrule\vspace{\baselineskip}
					\copyright~2026 IEEE. Personal use of this material is permitted. Permission from IEEE must be obtained for all other uses, in any current or future media, including reprinting/republishing this material for advertising or promotional purposes, creating new collective works, for resale or redistribution to servers or lists, or reuse of any copyrighted component of this work in other works. 
	}\end{spacing}}}
}

\begin{document}

\twocolumn[{%
\renewcommand\twocolumn[1][]{#1}%
\maketitle
\vspace{-1.4em}
\centering
\includegraphics[width=0.99\textwidth]{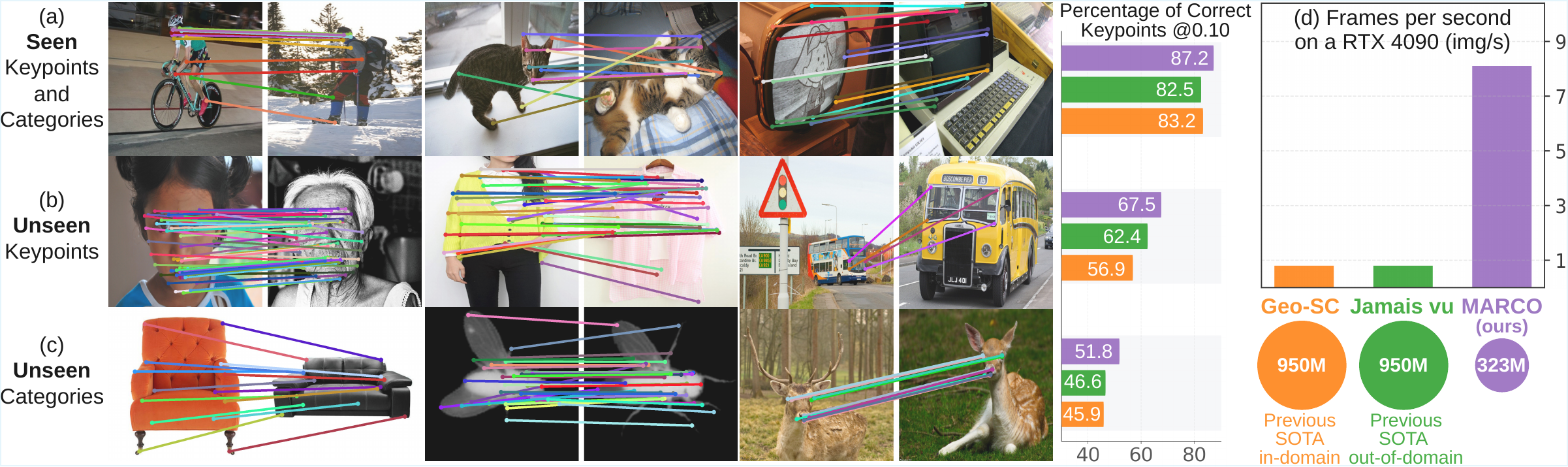}
\vspace{-0.5em}
\captionof{figure}{\textbf{MARCO, a model for generalizable correspondences.}
Built on DINOv2, \ours{} explores the unseen space of semantic correspondence by inferring structure that lies beyond the sparsity of keypoint annotations. During training, it discovers reliable matches across instances and propagates them smoothly across the object surface, transforming limited keypoint supervision into a dense training signal. Compared to prior dual-encoder approaches that pair DINOv2 with diffusion backbones, \ours{} achieves state-of-the-art accuracy on standard benchmarks \emph{(a)}, stronger generalization to unseen keypoints and categories \emph{(b–c)}, and remains $3\times$ smaller and $10\times$ faster \emph{(d)}.
\label{fig:teaser}}
\vspace{1.3\baselineskip}
}]

\begin{abstract}
Recent advances in semantic correspondence rely on dual-encoder architectures, combining DINOv2 with diffusion backbones. While accurate, these billion-parameter models generalize poorly beyond training keypoints, revealing a gap between benchmark performance and real-world usability, where queried points rarely match those seen during training.
Building upon DINOv2, we introduce \ours{}, a unified model for generalizable correspondence driven by a novel training framework that enhances both fine-grained localization and semantic generalization. By coupling a coarse-to-fine objective that refines spatial precision with a self-distillation framework, which expands sparse supervision beyond annotated regions, our approach transforms a handful of keypoints into dense, semantically coherent correspondences.
\ours{} sets a new state of the art on SPair-71k, AP-10K, and PF-PASCAL, with gains that amplify at fine-grained localization thresholds (+8.9 PCK@0.01), strongest generalization to unseen keypoints (+5.1, SPair-U) and categories (+4.7, MP-100), while remaining 3× smaller and 10× faster than diffusion-based approaches.
\end{abstract}

\thispagestyle{firststyle}
\section{Introduction\label{sec:introduction}}

Semantic correspondence estimation aims to establish pixel-level matches between semantically equivalent object regions~\cite{Min:2019:Spair, Zhang:2023:Tale, Lee:2019:SFNet, Liu:2011:SIFTflow, Cho:2023:CATS, Tang:2023:Dift}.
Accurate correspondences are essential for applications such as image editing~\cite{Wang:2024:COVE, Ofri:2023:Neural, Cohen:2011:Image}, pose estimation~\cite{Xu:2022:MP100, Nguyen:2024:Escape}, style transfer~\cite{Kim:2019:Transfer}, and affordance understanding~\cite{Lai:2021:Affordance}.
The task is challenging as corresponding parts often appear under significant variations in pose, texture, and viewpoint (\cf \cref{fig:teaser}), while supervision is typically available only at sparse locations.\\
Recent advances in large-scale pre-trained vision models have pushed this frontier forward. Self-supervised encoders, \eg DINOv2 \cite{Oquab:2023:Dinov2}, provide robust semantic alignment, while diffusion models \cite{Rombach:2022:SD} supply rich local structure and spatial detail \cite{Zhang:2023:Tale, Tang:2023:Dift, Elbanani:2024:Probing}. Unsurprisingly, their combination has become the dominant recipe for semantic correspondence \cite{Zhang:2024:Telling, Zhang:2023:Tale, Zhang:2024:Telling, Dunkel:2025:DIY}. However, this trend has led to computationally heavy architectures, requiring feature extraction from two encoders, and approaching one billion parameters.
More critically, we find that models trained with sparse keypoints generalize poorly, failing on both novel keypoints and unseen categories.
This limitation, also noted by the concurrent work Jamais Vu~\cite{Mariotti:2025:Jamais} for the case of unseen keypoints, highlights a broader gap between benchmark performance and real-world usability, where queried points rarely correspond to annotated ones. Our work aims to bridge this gap by introducing a unified correspondence model that \emph{(i)} achieves \emph{higher spatial precision} on supervised keypoints (\cf \cref{fig:teaser}a), \emph{(ii) generalizes better} to unseen keypoints and categories (\cref{fig:teaser}b-c), and \emph{(iii) remains compact and efficient} (\cref{fig:teaser}d).

Building upon DINOv2~\cite{Oquab:2023:Dinov2}, we pursue a minimalist architectural design, adding two components: bottleneck adapters \cite{Chen:2022:AdaptFormer} and a compact upsampling head to restore sub-patch resolution, incurring a parameter overhead of less than \SI{5}{\%}.
At the core of our approach lies a novel \textit{training framework} that guides the model toward fine-grained localization, while simultaneously leveraging reliable correspondences to propagate semantics across the entire object surface, extending supervision \textit{beyond the annotated regions}.
This is achieved through two complementary objectives.

First, we introduce a coarse-to-fine correspondence objective in which the spatial support of a Gaussian target distribution is gradually narrowed during training. This encourages the model to first learn region-level alignment and then progressively guides it toward subpatch-accurate localization. 
This improves correspondence accuracy across standard localization thresholds (\cref{fig:teaser}a) and achieves a substantial leap in precision at tighter ones: prior methods typically align coarse semantic parts (\eg, the \textit{eye}), whereas our model better resolves finer subregions (\eg, the \textit{pupil}).

\begin{figure}[t]
  \centering
  \includegraphics[width=\linewidth]{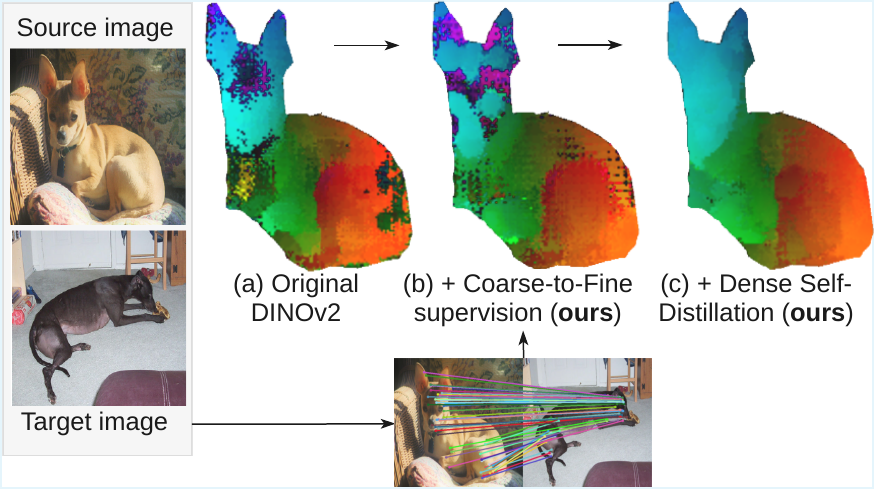}
    \vspace{-1.5em}
    \caption{\textbf{Flow consistency in DINOv2.} Semantic flow (in HSV space) from raw feature matches between two objects. Fine-tuning on sparse keypoints improves only the landmarks' representation, reducing geometric coherence (\emph{b}). Our self-supervised objective produces smooth, object-consistent flow across the surface (\emph{c}).}
    \label{fig:flow}
    \vspace{-0.5em}
\end{figure}

While this coarse-to-fine objective improves precision, it inherits the bias of sparse supervision. \cref{fig:flow} shows this effect: the frozen DINOv2 encoder produces a partially coherent dense correspondence field across the object, that, after supervised fine-tuning, contracts around the annotated landmarks (\eg, eyes, nose). 
Concurrent to our work, Jamais Vu \cite{Mariotti:2025:Jamais} mitigates this effect by mapping object points to 3D canonical templates, which remain inherently tied to the training categories, struggle to model highly deformable objects, and depend on a monocular depth model to estimate 3D geometry.
Instead, we introduce a general framework that discovers dense correspondences during training directly from the evolving representation, without depending on predefined 3D structures or category priors. 

Our formulation builds on the observation that the feature space of DINOv2, despite its limited spatial consistency, contains \textit{sparse yet reliable} correspondence cues. We exploit this property to \textit{densify} the sparse keypoint annotations across the full object surface, allowing the representation to serve as a source of self-supervision. During training, we mine reliable mutual matches across instances and densify them via piecewise-affine interpolation over Delaunay triangles, obtaining a continuous correspondence field that maps each source point to its target location. In this field, each set of reliable correspondences defines local geometric relationships that \textit{transport semantics smoothly across the object surface}. Since the resulting field inevitably contains erroneous matches due to model inaccuracies, symmetries, or occlusions, we retain only those that are geometrically consistent with annotated keypoints, which act as anchors.
Our solution effectively expands supervision from a handful of annotated keypoints to thousands of reliable correspondences, yielding features that vary smoothly across the object surface rather than collapsing around keypoints. Ultimately, this produces a smoother correspondence field than the frozen encoder (\cref{fig:flow}c).

Finally, to assess generalization, we propose a novel benchmark based on MP\text{-}100 \cite{Xu:2022:MP100}, containing novel keypoint definitions and categories never seen during training. Our final model, \ours{}, sets a new state of the art: it improves over Geo-SC \cite{Zhang:2024:Telling} by +4.0 \pckTen{} on SPair\text{-}71k and +2.9 on AP\text{-}10K, with substantial gains at strict matching thresholds, reaching +8.9 \pckOne. In generalization,
\ours{} outperforms Jamais Vu \cite{Mariotti:2025:Jamais} by +5.1 on SPair\text{-}U and by +5.6 on our MP-100 benchmark. 
Notably, \ours{} uses a single backbone, making it $3\times$ smaller and $10\times$ faster than dual-encoder methods (\cref{fig:teaser}d).

\noindent Summarizing, we make the following key contributions:
\begin{itemize}
\item We present a unified model for generalizable correspondence that attains state-of-the-art pixel-level accuracy and robust generalization to unseen keypoints and categories, with minimal computational overhead. 
\item We show that sparse keypoint supervision can be expanded into dense correspondences over the entire object surface by leveraging reliable matches emerging in the DINOv2 feature space, without relying on 3D templates, category priors, or external depth information.
\item We introduce a new benchmark to measures correspondence performance on \emph{(i)} unseen keypoints within known categories and \emph{(ii)} unseen keypoints on novel categories, offering a challenging testbed for future research.
\end{itemize}

\section{Related Work}
\label{sec:related}

\paragraph{Emergent correspondences in foundation models.}
Early work \cite{Tumanyan:2022:Splicing, Ofri:2023:Neural, Amir:2021:Deep} leveraged emergent properties in DINO~\cite{Caron:2021:DINO} for correspondence discovery. Recent studies \cite{Zhang:2023:Tale, Tang:2023:Dift} identified compelling properties in DINOv2~\cite{Oquab:2023:Dinov2}, providing semantically rich features, and in Stable Diffusion (SD)~\cite{Rombach:2022:SD}, offering fine-grained spatial detail.
Their complementarity has led to dual-encoder correspondence models \cite{Zhang:2023:Tale, Zhang:2024:Telling, Xue:2025:MATCHA, Mariotti:2025:Jamais, Dunkel:2025:DIY}.
To avoid the high cost of diffusion models, recent work adopted a single-backbone design based on DINOv2. 
DistillDIFT~\cite{Fundel:2025:Distillation} and GECO~\cite{Hartwig:2025:GECO} adapted DINOv2 with LoRA~\cite{Hu:2022:LoRA}: the former distilled SD+DINO in an unsupervised setting, while the latter introduced an optimal-transport objective, which requires category-aware keypoints (\eg,
labeling symmetric parts such as \emph{left} vs.\ \emph{right} eye). Despite being more efficient, both approaches are less accurate than dual-encoder architectures, particularly at fine-grained evaluation thresholds.  In contrast, we maintain the efficiency of a single-backbone design and introduce a novel coarse-to-fine objective that, coupled with a lightweight upsampling head, leads to more accurate correspondences, outperforming dual-encoder approaches without trading off accuracy for efficiency.

\myparagraph{Self-supervised dense semantic correspondence.}
The lack of densely annotated correspondences has motivated self-supervised alternatives to supervised learning.
Early methods~\cite{Liu:2011:SIFTflow, Rocco:2018:NCNet, Li:2020:Correspondence, Min:2021:Convolutional} processed 4D cost volumes and enforced spatial smoothness constraints, later extended with transformer attention~\cite{Cho:2023:CATS, Hong:2022:VAT} or implicit neural fields~\cite{Hong:2022:Neural}.
To avoid costly 4D reasoning, a different approach is to project the 4D correlations into 2D flow fields that are refined through correlation networks~\cite{Truong:2020:GLUNet} or hierarchical attention~\cite{Sun:2024:LPM}.
Other methods, like ours, employed self-supervised objectives exploiting augmented masks~\cite{Lee:2019:SFNet}, temporal consistency in videos~\cite{Jiang:2023:DuoDuo}, or distilled models trained on synthetic data~\cite{Li:2021:Probabilistic}.
Closer to our approach, SCorrSAN~\cite{Huang:2022:SCorrSAN} adopted a self-distillation strategy, where pseudo-labels are mined locally \textit{around annotated keypoints} using a small-loss criterion. Unlike prior work that learns correspondence from scratch, we \textit{adapt} a pre-trained encoder, whose global semantic structure degrades under sparse supervision.
To counter this collapse, we introduce a dense self-distillation objective that uses keypoints as anchors, mines additional reliable matches from the frozen encoder, and propagates them to obtain pseudo-correspondences over the entire object surface. This preserves, and ultimately enhances, the \textit{global} structure of the encoder beyond the supervised keypoints.

\myparagraph{Generalization in semantic correspondence.}
Supervised approaches \cite{Cho:2023:CATS, Mariotti:2025:Jamais, Zhang:2024:Telling, Hartwig:2025:GECO, Luo:2023:DHF} are typically evaluated \textit{in-domain}, \ie on landmarks seen during training.
Common benchmarks include PF-PASCAL \cite{Ham:2017:PfPascal}, AP-10K \cite{Yu:2021:AP10k}, and SPair-71k \cite{Min:2019:Spair}. PF-WILLOW \cite{Ham:2017:PfPascal} is often used for evaluation, but its limited difficulty \cite{Min:2019:Spair} makes it unsuitable to assess generalization.
Concurrently to our work, Jamais Vu \cite{Mariotti:2025:Jamais} highlighted this limitation and introduced SPair-U, adding a few \textit{unseen} keypoints (4 out of 20) to SPair-71k.
They learn a category-specific canonical representation by lifting keypoints to 3D via a monocular depth model, which improves transfer to novel keypoints but remains tied to the training taxonomy. In this work, we advocate for more challenging benchmarks to evaluate the generality of learned representations. We thus build a new benchmark from MP-100~\cite{Xu:2022:MP100} to assess generalization across (\textit{i}) {novel-keypoint splits}, introducing unseen landmarks within known categories, and
(\textit{ii}) {novel-category splits}, requiring transfer to object classes never observed during training.
\begin{figure*}
    \centering
    \includegraphics[width=0.975\linewidth]{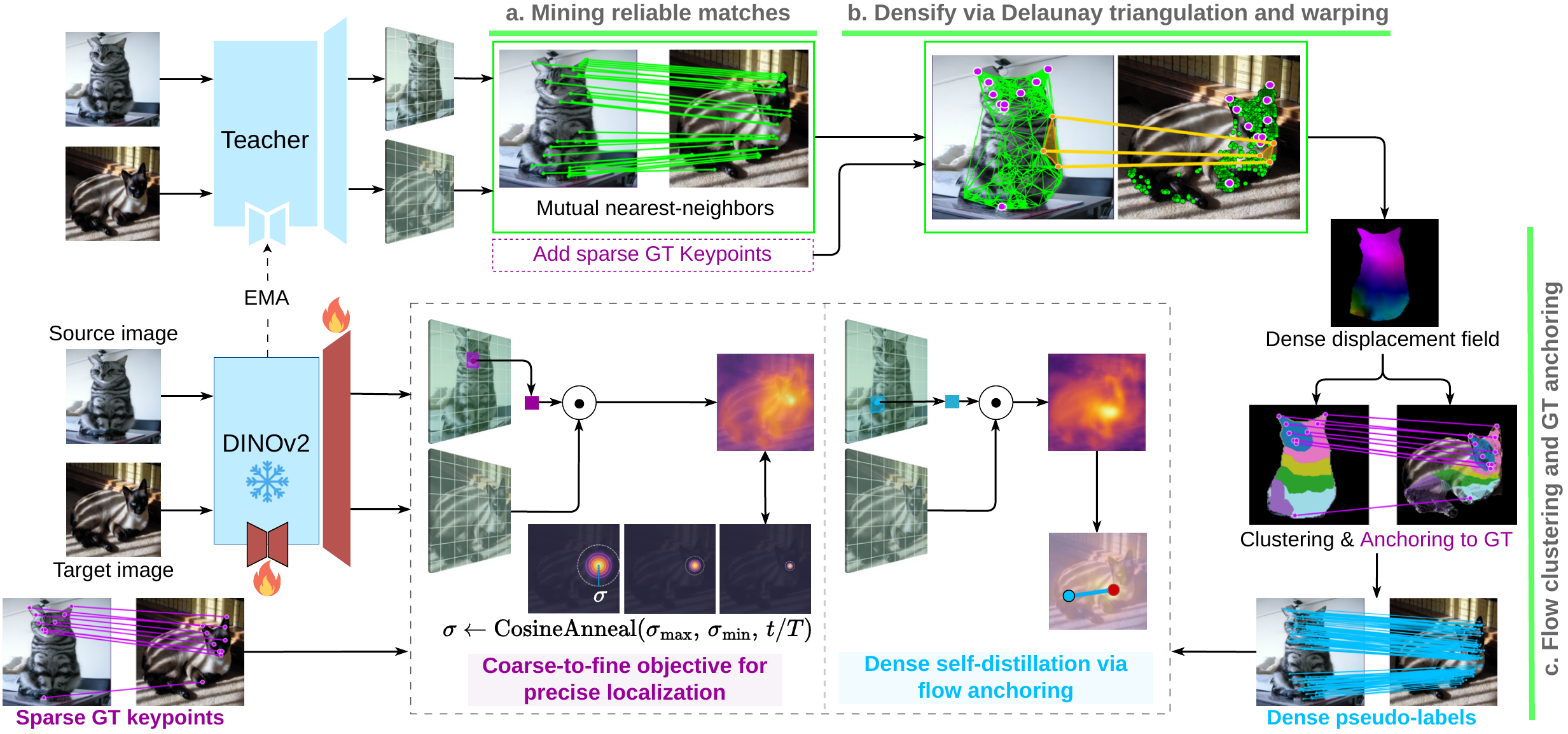}
    \vspace{-0.5em}
\caption{\textbf{Overview of \ours{}.} We insert lightweight adapters into DINOv2 and add a compact upsampling layer \emph{(red)}. At training time, we propose a \textit{coarse-to-fine} Gaussian RBF loss that progressively sharpens peaks on annotated keypoints and a self-distillation objective that exploits the pre-existing structure of DINOv2 features.
Given source and target images, we extract features from an EMA teacher and identify reliable mutual nearest-neighbor matches \emph{(a)}. These sparse correspondences are densified via piecewise-affine interpolation over a Delaunay triangulation, producing an initial flow field \emph{(b)}. Coherent motion regions are then obtained by clustering in the displacement space and anchored to sparse ground-truth keypoints, yielding geometrically consistent pseudo-labels \emph{(c)}. The student is trained to reproduce these pseudo-correspondences under a self-distillation objective, strengthening geometric coherence and improving generalization.}
    \label{fig:method}
    \vspace{-0.6em}
\end{figure*}

\section{Semantic Correspondence with MARCO}
\label{sec:method}
\paragraph{Problem statement.} \emph{Semantic correspondence} estimation is the task of establishing pixel-level matches between semantically equivalent keypoints.
Given source and target images $\mathbf{I}^s, \mathbf{I}^t \in \mathbb{R}^{H \times W \times 3}$ defined over the lattice $\Lambda=\{1,\ldots,H\}\times\{1,\ldots,W\}$, we assume that a learned function \( \Phi_\theta \) extracts feature maps $\mathbf{F}^s, \mathbf{F}^t \in \mathbb{R}^{D \times H' \times W'}$ of lower resolution $H'\times W'$.
Hence, for any pixel $\mathbf{p}^s \in \Lambda$ in $\mathbf{I}^s$, we can extract a local descriptor $\mathbf{F}^s[\mathbf{p}^s]\in\mathbb{R}^D$ using a suitable interpolation in the lower-resolution feature map.
From this descriptor, the task is then to identify \( \mathbf{p}^t \in \Lambda\) in $\textbf{I}^t$ such that both locations refer to the same landmark (\eg, \emph{front-left leg} of a chair). During training, the model is provided with image pairs alongside a sparse set of correspondences. For notational simplicity, we omit the source and target images and denote pairs as $\mathcal{E} = \{\, (\mathbf{p}_i^{s}, \mathbf{p}_i^{t})|_{i=1}^{K}\}$, where $K$ is the number of correspondences for the pair.

\myparagraph{Overview of \ours{}.} 
We leverage the semantic structure of DINOv2 features while enhancing their spatial consistency and fine-grained localization abilities. To this end, our method comprises \textit{architectural} elements (\cref{sec:architecture}) and a supervised \textit{coarse-to-fine} strategy (\cref{sec:supervised}). Together, these contributions give us strong in-domain results, consistently outperforming prior methods, particularly at fine-grained evaluation thresholds.
However, sparse keypoint supervision leads to overfitting on annotated regions. We introduce a dense self-distillation objective (\cref{sec:self_supervised}) that exploits the \textit{partially consistent} local structure in DINOv2 features. By identifying and propagating reliable matches across the object, our approach converts them into dense supervision, enhancing correspondence in unseen regions.

\subsection{Architecture}
\label{sec:architecture}
Our architecture (\cf \cref{fig:method}) builds upon a pre-trained DINOv2 \cite{Oquab:2023:Dinov2} backbone, which provides strong semantic matches \cite{Mariotti:2024:SphericalMaps, Hartwig:2025:GECO} but lacks spatial coherence compared to diffusion models \cite{Zhang:2023:Tale, Tang:2023:Dift, Cuttano:2025:Sansa}.
Recent semantic correspondence methods address this by {combining DINOv2 with Stable Diffusion}~\cite{Zhang:2024:Telling, Li:2024:Sd4match, Mariotti:2025:Jamais, Xue:2025:MATCHA, Dunkel:2025:DIY}, leveraging their complementary strengths.
In contrast, we pursue a \emph{minimalist single-backbone route}, enhancing the representations of DINOv2 \emph{directly inside the network} through
\emph{(i)} lightweight adapter modules for feature enrichment, and
\emph{(ii)} a compact upsampling layer for sub-patch refinement.
This strategy enriches DINOv2's geometric awareness and preserves its representation while keeping the architecture efficient.

\myparagraph{Adapter-based feature enrichment.}  We insert AdaptFormer modules~\cite{Chen:2022:AdaptFormer} into the higher layers of the transformer backbone.
Each adapter operates token-wise and consists of a bottleneck with learnable down- and up-projections 
$\mathbf{W}_{\text{down}} \in \mathbb{R}^{D \times d}$,
$\mathbf{W}_{\text{up}} \in \mathbb{R}^{d \times D}$, with embedding dimension $D$, bottleneck dimensionality $d$, and $d \ll D$.
Given token embeddings \( \mathbf{x} \in \mathbb{R}^D \), the adapter is defined as
\begin{equation}
\mathcal{A}(\mathbf{x}) = \mathrm{GELU}(\mathbf{x} \mathbf{W}_{\text{down}})\mathbf{W}_{\text{up}}.
\end{equation}
The adapted representation is summed in a residual manner:
\begin{align}
\mathbf{x}_{\text{self}} &= \mathrm{Attention}(\mathbf{x}), \\
\mathbf{x}' &= \mathrm{MLP}(\mathbf{x}_{\text{self}}) + \mathbf{x}_{\text{self}} + \mathcal{A}(\mathbf{x}_{\text{self}}).
\end{align}
The backbone weights are kept frozen; only the adapter matrices \(\mathbf{W}_{\text{down}}\) and \(\mathbf{W}_{\text{up}}\) are learned.
This allows us to refine high-level features with minimal parameter overhead.

\myparagraph{Feature upsampling.} Each token in DINOv2 features represents a 14$\times$14 image patch. This coarse granularity limits the achievable localization accuracy.
Many existing correspondence pipelines refine coarse matches through additional modules such as CNN matchers or cross-attention blocks, as common in image matching~\cite{Sun:2021:Loftr, Edstedt:2024:Roma} and point tracking~\cite{Tumanyan:2024:DINOTracker}.
These refinement stages improve matching precision but introduce substantial computational overhead. Instead, we use a lightweight upsampling layer that increases the feature resolution by a factor of \( \times 4 \). Formally, given  $\mathbf{F} = \Phi_\theta(\mathbf{I})$, the upsampled dense features $\hat{\mathbf{F}}$ on lattice $\hat{\Lambda}=\{1,\ldots,4H'\}\times \{1,\ldots,4W'\}$ are obtained as:
\begin{align}
\mathbf{F}_1 &= \mathrm{ConvTranspose}_{2 \times 2}(\mathbf{F}), \\
\hat{\mathbf{F}} &= \mathrm{DepthwiseConv}_{3 \times 3}(\mathrm{GELU}(\mathbf{F}_1)).
\end{align}
This refinement layer enhances localization accuracy while preserving efficiency and architectural simplicity.

\subsection{Coarse-to-fine objective for precise localization
}
\label{sec:supervised}
Our goal is to leverage available supervision to enhance the spatial precision of the predicted correspondences.
Recent methods \cite{Zhang:2024:Telling, Mariotti:2025:Jamais, Xue:2025:MATCHA, Dunkel:2025:DIY} regress keypoint coordinates using a soft-argmax operator, supervised with an $\mathcal{L}_2$ loss.
This objective is known to be unstable under multi-modal distributions \cite{Chen:2019:Over, Pan:2024:Sampling, Yang:2025:Heatmap}:
by driving predictions toward the mean of competing modes, rather than enforcing a uniform sharp peak, it produces over-smoothed responses that degrade localization accuracy. Therefore, we replace coordinate regression with a distributional matching objective ~\cite{Gu:2023:BCIR, Zhang:2020:DARK, Yang:2025:Heatmap}, where the predicted probability map is supervised to match a \textit{Gaussian RBF kernel centered at the ground-truth keypoint} via a cross-entropy (CE) loss \cite{Li:2023:SimSC}. 

Given the correspondences
$\mathcal{E} = \{\, (\mathbf{p}_{i}^{s}, \mathbf{p}_{i}^{t})|_{i=1}^{K}\}$,
we extract the source descriptor $\hat{\mathbf{F}}^s[\mathbf{p}^s_i]$ and compute its dense similarity map with respect to the target feature grid:
\begin{equation}
S(\mathbf{p}_i^s,\mathbf{u}) = \langle \hat{\mathbf{F}}^s[\mathbf{p}^s_i], \hat{\mathbf{F}}^t[\mathbf{u}] \rangle, \quad \forall \mathbf{u} \in \hat{\Lambda}.
\end{equation}
The RBF kernel and the CE loss are defined as
\begin{align}
G_\sigma(\mathbf{u}; \mathbf{p}_i^t) \propto & \exp\left(-\frac{\|\mathbf{u} - \mathbf{p}_i^t\|_2^2}{2\sigma^2}\right), \\
\label{eq:ce_loss}
\mathcal{L}_{\text{sup}} = - \frac{1}{K} \sum_{i=1}^K \sum_{\mathbf{u}\in \hat{\Lambda}} &G_\sigma(\mathbf{u}; \mathbf{p}_i^t) \log \softmax S(\mathbf{p}_i^s,\mathbf{u}).\raisetag{3.5em}
\end{align}

The bandwidth $\sigma$ represents the spread of the target kernel around the GT keypoint (\cf \cref{fig:method}). While, in principle, adjusting $\sigma$ controls the sharpness of the distribution without altering the peak location, we find it instead to strongly affect learning dynamics: a large $\sigma$ yields coarse matches, encouraging the model to align broad semantic parts but not to resolve fine-grained details; a small $\sigma$ enforces sharp localization, yielding accurate predictions on a few confident matches while degrading overall correspondence accuracy.

We expose this inherent trade-off and propose a \textit{coarse-to-fine} annealing strategy to balance these competing effects. We start training with a wide kernel that promotes stable region-level matching and gradually decrease its bandwidth $\sigma$, adopting a cosine annealing schedule:
\begin{equation}
\sigma(t) = \sigma_{\min} 
+ \frac{1}{2}\,(\sigma_{\max} - \sigma_{\min})
\left( 1 + \cos\Big(\pi\,\frac{t}{T}\Big) \right). 
\end{equation}

This schedule allows learning to gradually progress from \emph{broad semantic alignment} to \emph{precise geometric localization}.

\subsection{Dense self-distillation via flow anchoring}
\label{sec:self_supervised}

The available supervision offers precise, yet limited guidance, typically covering fewer than 20 landmarks \cite{Min:2019:Spair, Yu:2021:AP10k} per image. 
Relying solely on this sparse supervision causes the model to specialize to annotated landmarks~\cite{Mariotti:2025:Jamais}, leading to a \textit{representation collapse} and ultimately to underperforming the unmodified DINOv2 encoder on unseen keypoints.  To this end, we observe that \emph{(i)} object regions in frozen DINOv2 features exhibit partially \textit{consistent flow} across views (\cf \cref{fig:flow}\textit{a}), and that \emph{(ii)} this can be exploited during training to transform its evolving feature space into a source of self-supervision. 
Concretely, we propose a strategy to automatically discover reliable correspondences, densify them over the object, and use annotated keypoints as local anchors to determine trustworthy pseudo-labels. Learning is stabilized through self-distillation, where an exponential moving average (EMA) teacher generates the pseudo-labels for the student.
This strategy enhances spatial consistency of the pre-trained features (\cf \cref{fig:flow}\textit{c}).

\myparagraph{Estimating dense semantic flow.}
Let $\mathbf{F}_T^s, \mathbf{F}_T^t$ denote the upsampled features extracted by the teacher network. For each patch $\mathbf{u} \in \hat{\Lambda}$, we find its most similar target feature
\begin{equation}
    \underset{s \rightarrow t}{\mathrm{NN}}(\mathbf{u}) = \arg\max_{\mathbf{v} \in \hat{\Lambda}} \langle \mathbf{F}_T^s[\mathbf{u}], \mathbf{F}_T^t[\mathbf{v}] \rangle ,
\end{equation}
and symmetrically compute $\mathrm{NN}_{t \rightarrow s}(\mathbf{v})$ for each $\mathbf{v} \in \hat{\Lambda}$.  
The set of mutual nearest neighbors is defined as
\begin{equation}
    \mathcal{P}_{\text{MNN}} = \bigl\{ (\mathbf{u},\mathbf{v}) \mid  \underset{s \rightarrow t}{\mathrm{NN}}(\mathbf{u})=\mathbf{v} \, \wedge \, \underset{t \rightarrow s}{\mathrm{NN}}(\mathbf{v})=\mathbf{u} \bigr\}.
\end{equation}
This set, together with the annotated keypoints $\mathcal{E}$, provides a reliable, yet sparse set of correspondences over the object (\cf \cref{fig:method}\emph{a}). 
We denote this \textit{seed} set as  $\mathcal{P}_{\text{seed}} = \mathcal{E} \cup \mathcal{P}_{\text{MNN}}$.
We restrict $\mathcal{P}_{\text{seed}}$ to pixels inside object masks derived with SAM \cite{Kirillov:2023:SAM} to suppress potential outliers, following previous works \cite{Dunkel:2025:DIY, Fundel:2025:Distillation, Mariotti:2024:SphericalMaps, Mariotti:2025:Jamais}, or alternatively inside bounding boxes derived from available ground-truth keypoints (\cf Supp. Mat. for more details).
The \textit{seed} set serves as starting point to densify correspondences over the whole object (\cf \cref{fig:method}\emph{b}). To do so, we take the source keypoints of the \textit{seed} set $\mathcal{P}_\text{source}=\{\,\mathbf{u}_i \mid (\mathbf{u}_i, \mathbf{v}_i) \in \mathcal{P}_{\text{seed}}\}$ and construct a Delaunay triangulation $\mathcal{T}\equiv\mathcal{T}(\mathcal{P}_{\text{source}})$, which partitions their convex hull into non-overlapping triangles.
For each resulting triangle $\tau \in \mathcal{T}$, a corresponding triangle in the target $\tau'$ is defined by replacing every source vertex with its matched target point.
This operation \textit{lifts} the mined correspondences from discrete points to local triangular regions. To densify correspondences within triangle pairs, we estimate an affine warp between each pair, $\mathcal{W}_{\tau\rightarrow\tau'}$, mapping every pixel lying in the triangle $\mathbf{u}\!\in\!\tau$ to the position $\mathcal{W}_{\tau\rightarrow\tau'}(\mathbf{u})$ in the target image $\mathbf{I}^t$ (\cf \cref{fig:method}).
The union of these transformations forms a continuous piecewise-affine warp $\mathcal{\hat{W}}$ over the convex hull defined by $\mathcal{P}_{\text{source}}$.
We represent the resulting \textit{dense} flow as a displacement map $\mathbf{D} : \hat{\Lambda} \rightarrow \mathbb{R}^{2}$: 
\begin{equation}
    \begin{aligned}
        \mathbf{u} &\mapsto \mathbf{D}(\mathbf{u})= \mathcal{\hat{W}}(\mathbf{u})-\mathbf{u}.
    \end{aligned}
\end{equation}

\begin{figure}[t]
  \centering
  \includegraphics[width=0.99\linewidth]{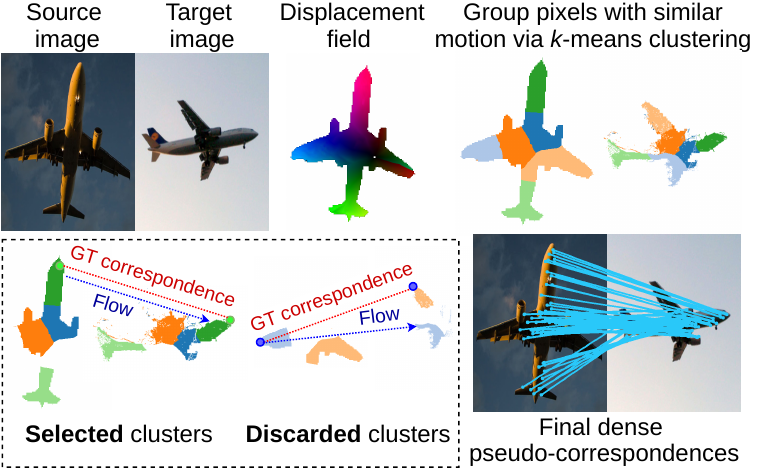}
    \vspace{-0.5em}
    \caption{\textbf{Correspondence mining via flow anchoring.} Given a source and target image, the dense displacement field is estimated from sparse matches using piece-wise affine motion on a Delaunay triangulation. We then identify regions of coherent motion via $k$-means clustering in displacement space. Due to model inaccuracies, occlusion, or object symmetries, coherent motion may lead to incorrect matches: we anchor clusters to GT keypoints and retain those having a coherent flow toward the correct region. For clarity of visualization, final dense correspondences are downsampled.} 
    \label{fig:flow_clusters}
    \vspace{-0.5em}
\end{figure}

\myparagraph{Flow clustering and GT anchoring.}
The estimated flow field maps pixels from the source object onto the target, but often includes erroneous correspondences arising from object symmetries, occlusions, or model inaccuracies (\eg, the \textit{left wing} of a plane mapped to the \textit{right wing}, as in \cref{fig:flow_clusters}). Intuitively, we identify reliable correspondences as those regions that \emph{(i)} exhibit locally consistent flow, and \emph{(ii)} whose flow is also consistent with ground-truth correspondences (\cf \cref{fig:flow_clusters}).
To this end, we group regions with coherent \textit{motion} through clustering in the flow space, and anchor clusters to GT matches to filter out outliers. Specifically, we cluster the flow vectors via $k$-means, avoiding manual selection of $k$ by over-clustering and greedily merging clusters until the Bayesian Information Criterion (BIC) is maximized.
For each flow cluster $\Omega_n, n=\{1,\ldots,k\}$, we can identify a pair of corresponding regions, $C_n^s$ in the source and $C_n^t$ in the target image, which mark the source and destination of the flow. 
The final set of pseudo-labels is obtained by retaining only clusters whose associated regions contain a pair of annotated keypoints, \ie $\mathbf{p}_i^s$ and $\mathbf{p}_i^t$:
\begin{align}
    C_n^s &= \{\mathbf{u} \mid \mathbf{D}(\mathbf{u}) \in \Omega_n \}, \\
    C_n^t &= \{\mathbf{u} + \mathbf{D}(\mathbf{u}) \mid \mathbf{u} \in C_n^s \}, \\
    \begin{split}
    \mathcal{P}_{\text{self}} &= \bigl\{\big( \mathbf{u}, \mathbf{u}+\mathbf{D}(\mathbf{u}) \big) \mid \exists \, n,  (\mathbf{p}_i^s,\mathbf{p}_i^t)\!\in\mathcal{E} : \\
    & \qquad\qquad\mathbf{u}\in C^s_n \wedge \mathbf{p}_i^s \!\in\! C^s_n \wedge \mathbf{p}_i^t \!\in\! C^t_n \bigr\}.
    \end{split}
\end{align}

This formulation preserves clusters that exhibit plausible motion while excluding regions that move consistently, yet correspond to the wrong location.

\myparagraph{Self-distillation.}
Pseudo-labels are generated by a teacher network with parameters $\theta_{\mathrm{T}}$, maintained as an exponential moving average (EMA) of the student parameters $\theta_\text{S}$,
\begin{equation}
\theta_{\mathrm{T}} \leftarrow \beta\,\theta_{\mathrm{T}} + (1-\beta)\,\theta_{\mathrm{S}},
\end{equation}

Given that pseudo-labels are inherently noisy, we avoid enforcing a strict spatial prior as in the CE loss (Eq.\ \ref{eq:ce_loss}) and adopt a standard $\mathcal{L}_2$ regression for the student:
\begin{align}
\mathcal{L}_{\text{self}} =
\frac{1}{|\mathcal{P}_{\text{self}}|}
\sum_{(\hat{\mathbf{u}},\hat{\mathbf{v}}) \in \mathcal{P}_{\text{self}}}
\bigl\|\SoftArgmax_{\mathbf{u} \in \hat{\Lambda}}(S(\hat{\mathbf{u}}, \mathbf{u})) - \hat{\mathbf{v}} \bigr\|_2^2 ,\raisetag{.8em}
\end{align}
where $S(\hat{\mathbf{u}}, \cdot)$ is the similarity between $\hat{\mathbf{u}}$ and the target.  

\definecolor{nomasks}{gray}{0.55}
\newcommand{\ind}[1]{\textcolor{nomasks}{#1}}

\begin{table*}[t]
\centering
\caption{\textbf{Evaluation on standard benchmarks.} 
Per-image PCK (\%, $\uparrow$) at multiple thresholds on SPair-71k, AP-10K (intra-, cross-species, and cross-family), and PF-PASCAL.  
Best results \textbf{bold}, \nth{2} best \underline{underlined}. $\S$ uses depth maps at training; $\dagger$ uses object masks at training; $\ddagger$ uses object masks at inference. For \ours, we report two variants, with and \ind{without} restricting pixels to object masks during training. \textbf{MARCO}, built solely on DINOv2, sets a new state of the art, with strong gains at challenging fine-grained thresholds (PCK@0.01).}
\label{tab:benchmark_spair_ap_pfpascal}
\tablesize
\vspace{-0.75em}
\setlength{\tabcolsep}{4pt}
\begin{tabularx}{\linewidth}{@{}Xcl|S[table-format=2.1]cc|S[table-format=2.1]cc|S[table-format=2.1]cc|S[table-format=2.1]cc|ccc@{}}
\toprule
 & \multicolumn{2}{c}{\textbf{Encoders}} & \multicolumn{3}{c}{\textbf{SPair-71k}} 
& \multicolumn{3}{c}{\textbf{AP-10K (I.S.)}} 
& \multicolumn{3}{c}{\textbf{AP-10K (C.S.)}} 
& \multicolumn{3}{c}{\textbf{AP-10K (C.F.)}} 
& \multicolumn{3}{c@{}}{\textbf{PF-PASCAL}} \\
\cmidrule(lr){2-3}\cmidrule(lr){4-6}\cmidrule(lr){7-9}\cmidrule(lr){10-12}\cmidrule(lr){13-15}\cmidrule(lr){16-18}
\textbf{Method} & DINOv2  &  SD &
0.01 & 0.05 & 0.10 
& 0.01 & 0.05 & 0.10 
& 0.01 & 0.05 & 0.10 
& 0.01 & 0.05 & 0.10 
& 0.05 & 0.10 & 0.15 \\
\midrule
\multicolumn{2}{@{}l}{\textit{Unsupervised / Weakly Supervised}} &&&&&&&&&&&&&& \\
~~DINOv2+NN~\cite{Zhang:2023:Tale} & \checkmark &  &
6.3 & 38.4 & 53.9 
& 6.4 & 41.0 & 60.9 
& 5.3 & 37.0 & 57.3 
& 4.4 & 29.4 & 47.4 
& 63.0 & 79.2 & 85.1 \\
~~DIFT~\cite{Tang:2023:Dift}  &  & \checkmark 
& 7.2 & 39.7 & 52.9 
& 6.2 & 34.8 & 50.3 
& 5.1 & 30.8 & 46.0 
& 3.7 & 22.4 & 35.0 
& 66.0 & 81.1 & 87.2 \\
~~SD{+}DINOv2~\cite{Zhang:2023:Tale} & \checkmark & \checkmark 
& 7.9 & 44.7 & 59.9 
& 7.6 & 43.5 & 62.9 
& 6.4 & 39.7 & 59.3 
& 5.2 & 30.8 & 48.3 
& 71.5 & 85.8 & 90.6 \\
~~DIY-SC \cite{Dunkel:2025:DIY} $^\dagger$ & \checkmark & \checkmark 
& 10.1 & 53.8 & 71.6 & \text{--} & \text{--} & 70.6 & \text{--} & \text{--} & 69.8 & \text{--} & \text{--} & 57.8 & \text{--} & \text{--} & \text{--}\\
\midrule
\multicolumn{2}{@{}l}{\textit{Supervised}} &&&&&&&&&&&&& \\
~~DHF~\cite{Luo:2023:DHF} &  & \checkmark 
& 8.7 & 50.2 & 64.9 
& 8.0 & 45.8 & 62.7 
& 6.8 & 42.4 & 60.0 
& 5.0 & 32.7 & 47.8 
& 78.0 & 90.4 & 94.1 \\
~~SD{+}DINOv2~\cite{Zhang:2023:Tale}  & \checkmark & \checkmark 
& 9.6 & 57.7 & 74.6 
& 9.9 & 57.0 & 77.0 
& 8.8 & 53.9 & 74.0 
& 6.9 & 46.2 & 65.8 
& 80.9 & 93.6 & 96.9 \\
~~GECO~\cite{Hartwig:2025:GECO} $^\dagger$ & \checkmark &  
& 14.2 & 59.6 & 73.6 & 19.2 & 67.1 & 82.5 & 17.4 & 64.9 & 81.2 & 14.5 & 60.4 & 76.6 & 80.5 & 92.3 & 95.7 \\
~~Jamais Vu~\cite{Mariotti:2025:Jamais} $^{\dagger \S}$
& \checkmark & \checkmark 
& 20.5 & 71.9 & 82.5 
& \text{--} & \text{--} & \text{--} 
& \text{--} & \text{--} & \text{--} 
& \text{--} & \text{--} & \text{--} 
& \text{--} & \text{--} & \text{--} \\
~~Geo\text{-}SC ~\cite{Zhang:2024:Telling} $^\ddagger$ & \checkmark & \checkmark & 21.7 &  72.8 &  83.2 &  23.2 &  73.2 &  87.7 &  21.7 &  70.3 &  85.9 & 18.3 &  63.2 &  78.5&  85.3&  95.0&  97.4 \\

~~\textbf{MARCO} \emph{(ours)} & \checkmark & & \ind{26.6} & \ind{75.5} & \ind{86.7} & \ind{31.1} & \ind{77.0} & \ind{88.7} & \ind{31.3} & \ind{74.7} & \ind{87.5} & \ind{27.4} & \ind{69.9} & \ind{82.5} & \ind{90.3} & \ind{96.8} & \ind{98.6}\\
~~\textbf{MARCO} \emph{(ours)} $^\dagger$ & \checkmark &  
& \textbf{27.0} & \textbf{77.6} & \textbf{87.2} 
& \textbf{32.6} & \textbf{77.4} & \textbf{89.1} 
& \textbf{32.2} & \textbf{76.6} & \textbf{88.3} 
& \textbf{28.5} & \textbf{71.1} & \textbf{83.4} 
& \textbf{91.1} & \textbf{96.9} & \textbf{98.6} \\
\bottomrule
\end{tabularx}
\vspace{-1.1em}
\end{table*}

\section{Experiments}
\begin{figure}[h]
  \centering

  \newlength{\colgap}\setlength{\colgap}{2pt}               
  \setlength{\tabcolsep}{\colgap}                          
  \newlength{\colw}\setlength{\colw}{\dimexpr(\linewidth - 8\colgap)/5\relax}
  \newlength{\tile}\setlength{\tile}{0.5\colw}              
  \newlength{\twobytwo}\setlength{\twobytwo}{2\tile}        
\newlength{\singlesize}\setlength{\singlesize}{\dimexpr 1.25\tile\relax} 

  \newcommand{\headsize}{\scriptsize}               
  \newlength{\headh}\setlength{\headh}{1.2\baselineskip}    
  \newcommand{\Header}[1]{\headsize \rule{0pt}{\headh}#1}   
  \newcolumntype{C}{>{\centering\arraybackslash}m{\colw}}   

  \newcommand{\TileBox}[1]{%
    \begin{minipage}[c][\tile][c]{\tile}\centering #1 \end{minipage}%
  }

  \newcommand{\TileImg}[1]{\TileBox{\includegraphics[width=\tile,height=\tile,keepaspectratio]{#1}}}
  \newcommand{\TileDots}{\TileBox{\small$\cdots$}}

  \newlength{\innergap}\setlength{\innergap}{1pt}
  \newcommand{\GridTwoByTwo}[4]{
    \begingroup
      \setlength{\tabcolsep}{0pt}\renewcommand{\arraystretch}{1.7}
      \begin{tabular}{@{}c@{\hspace{\innergap}}c@{}}
        #1 & #2 \\[\innergap]
        #3 & #4
      \end{tabular}
    \endgroup
  }

\newcommand{\CenterSingleTileImg}[1]{%
  \begin{minipage}[c][\twobytwo][c]{\twobytwo}\centering
    \includegraphics[width=\singlesize,height=\singlesize,keepaspectratio]{#1}%
  \end{minipage}%
}

\renewcommand{\arraystretch}{0.5}
  \begin{tabular}{@{}C|C|C|C|C@{}}
    \toprule
    \vspace{-.75em}
    \Header{Human face} &
    \vspace{-.75em}
    \Header{Apparel item} &
    \vspace{-.75em}
    \Header{Animal body} &
    \vspace{-.75em}
    \Header{Home furniture} &
    \vspace{-.75em}
    \Header{Animal face} \\
    \midrule
    \parbox{\colw}{\centering
    \CenterSingleTileImg{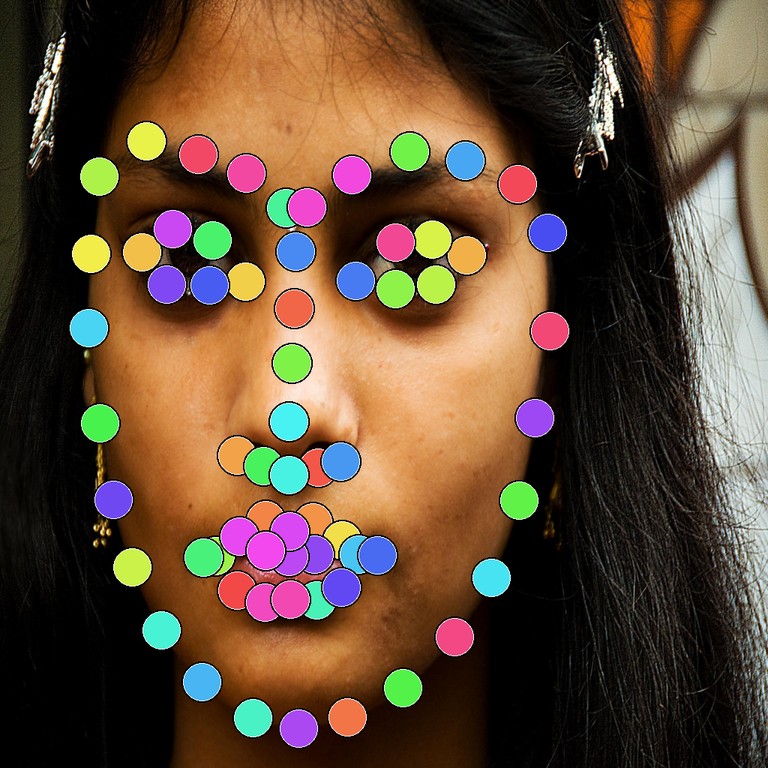}
    } &
    \parbox{\colw}{\centering
      \GridTwoByTwo
        {\TileImg{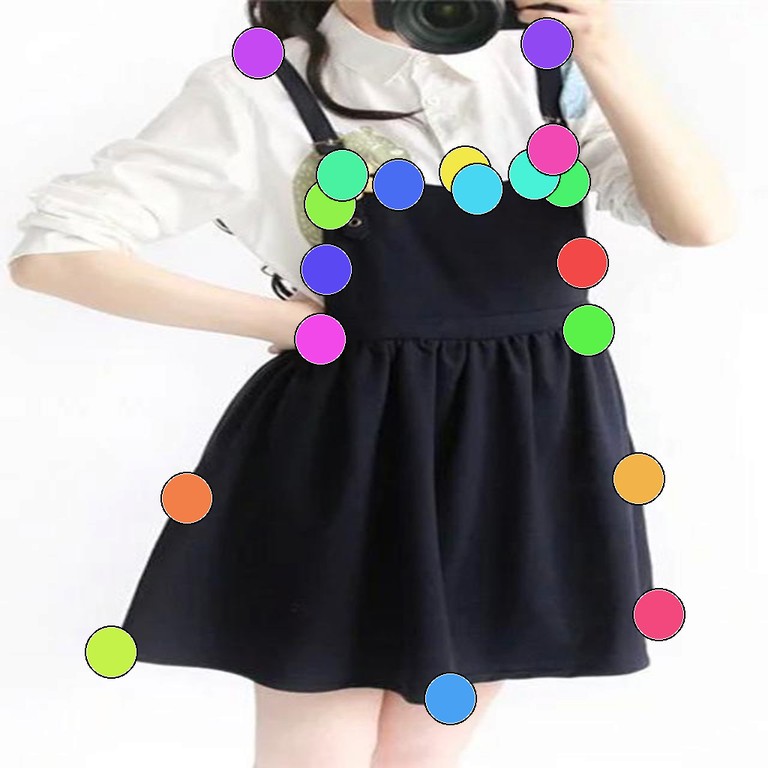}}{\TileImg{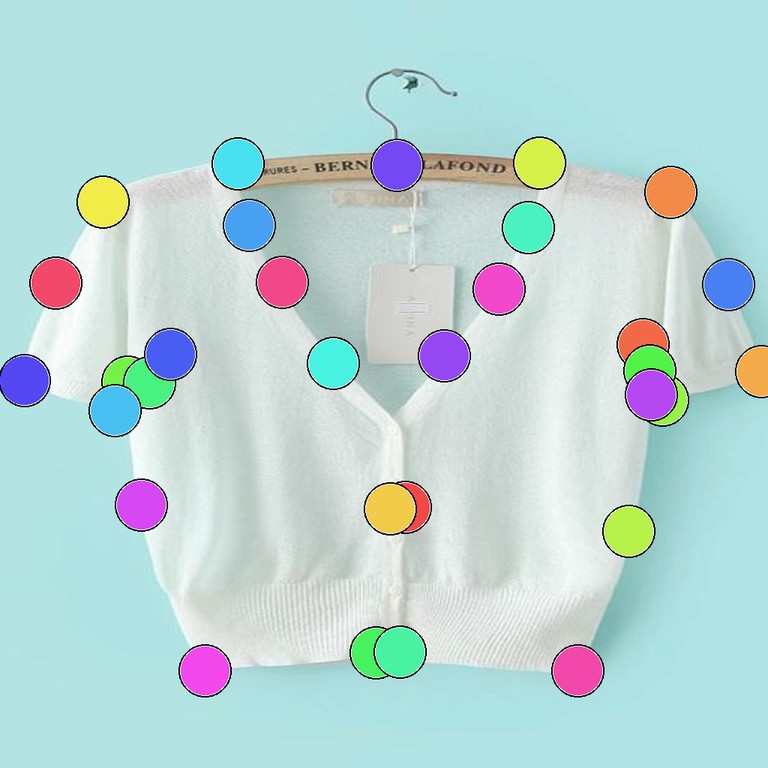}}
        {\TileImg{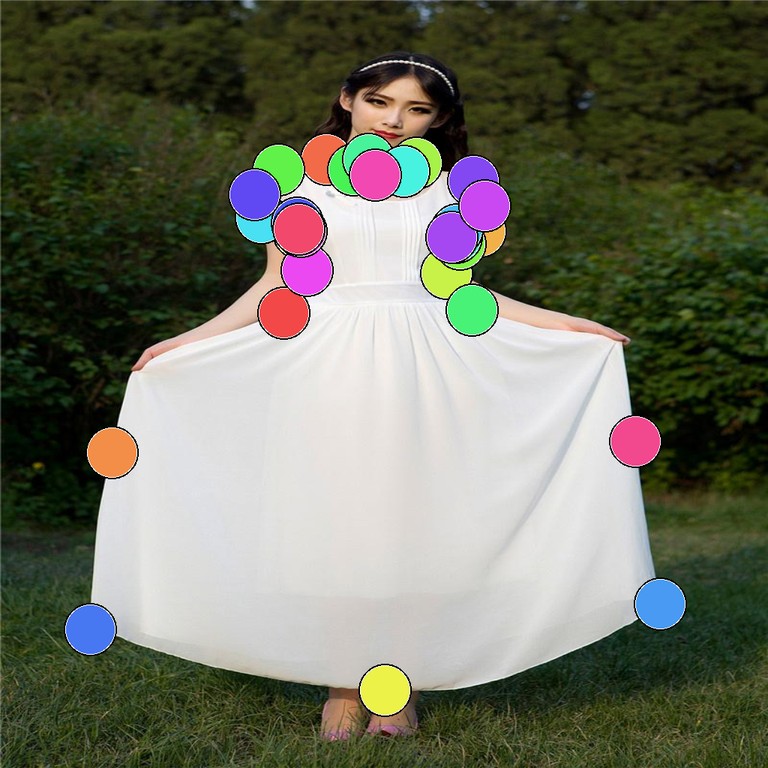}}{\TileDots}
    } &
    \parbox{\colw}{\centering
      \GridTwoByTwo
        {\TileImg{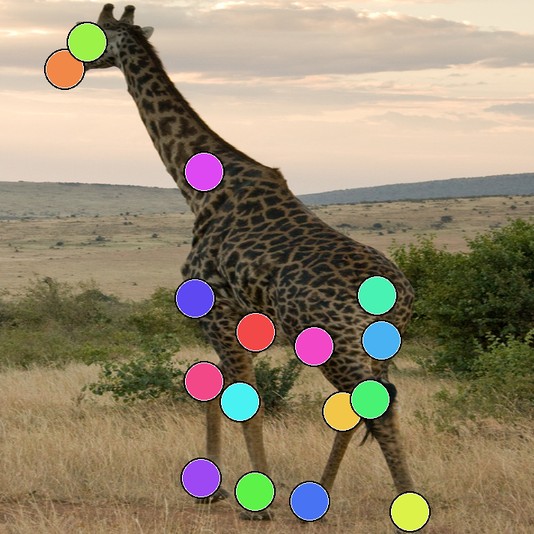}}{\TileImg{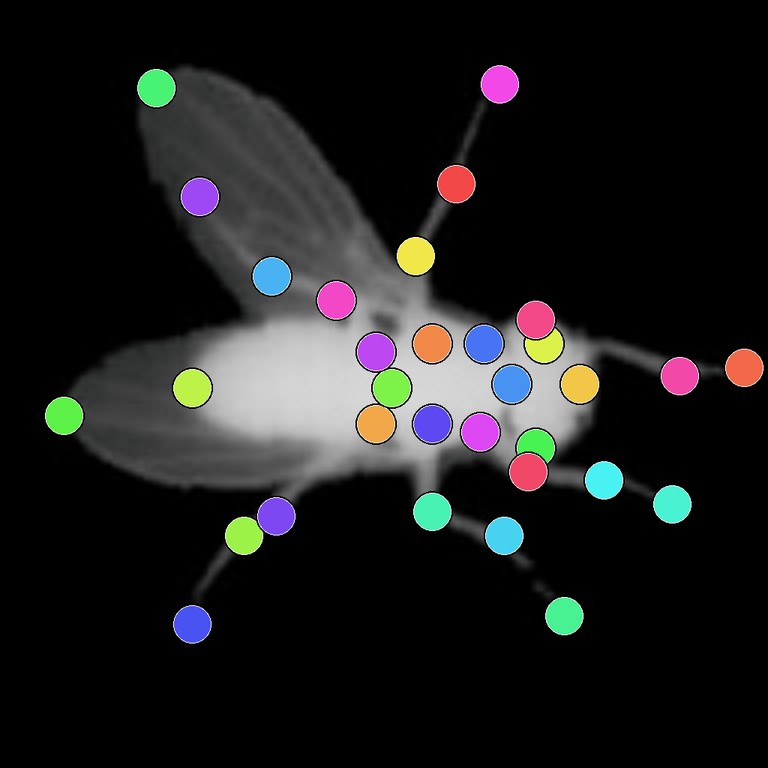}}
        {\TileImg{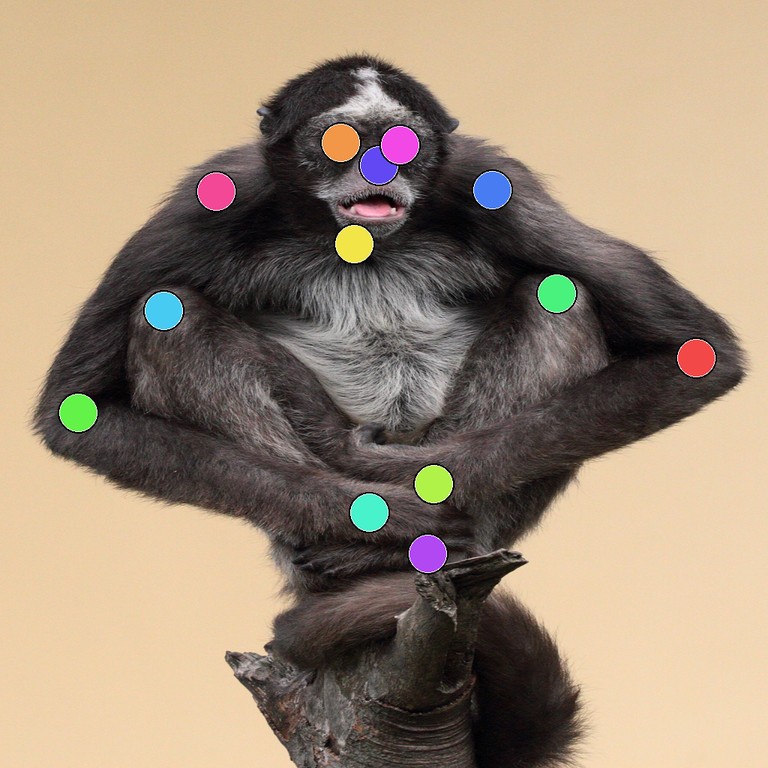}}{\TileDots}
    } &
    \parbox{\colw}{\centering
      \GridTwoByTwo
        {\TileImg{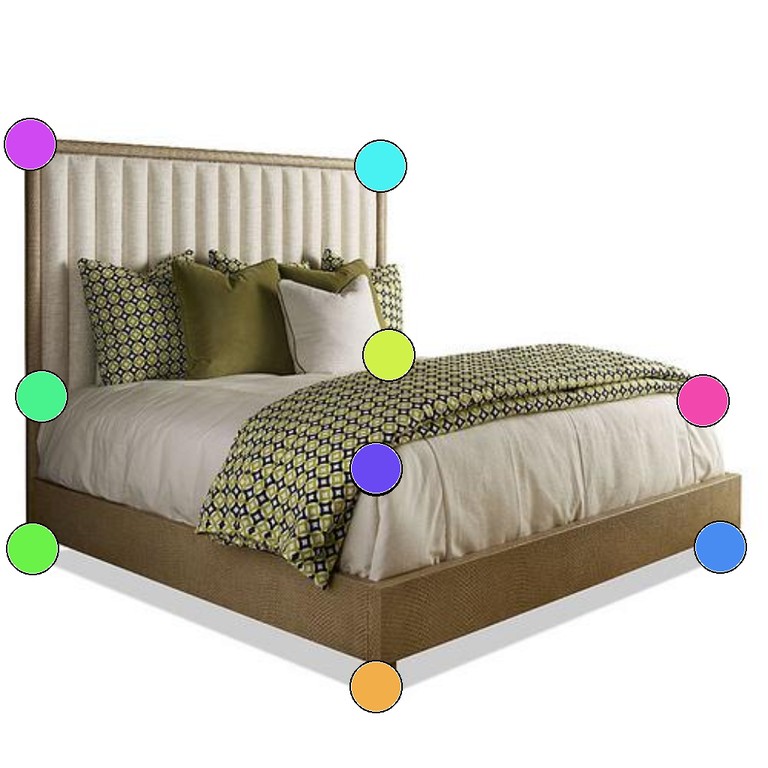}}{\TileImg{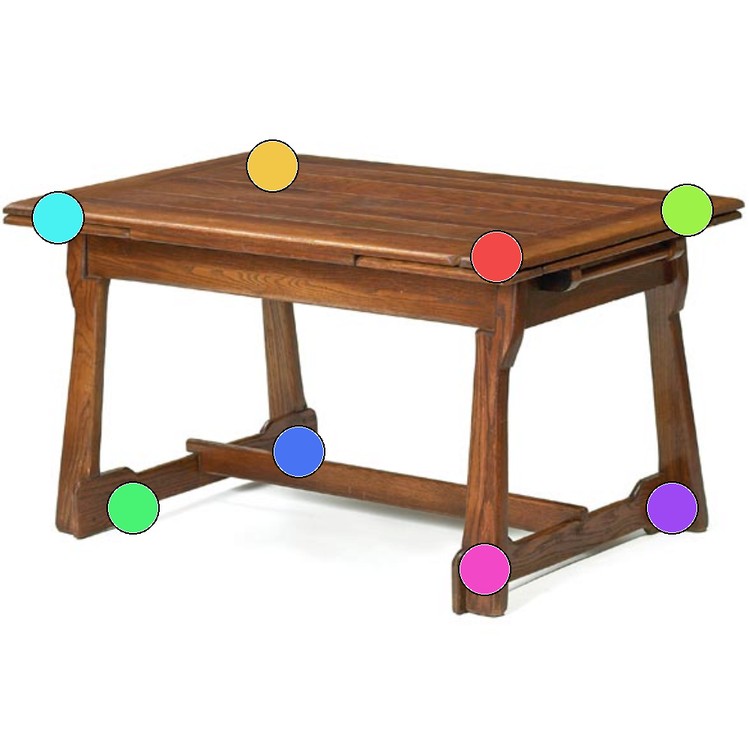}}
        {\TileImg{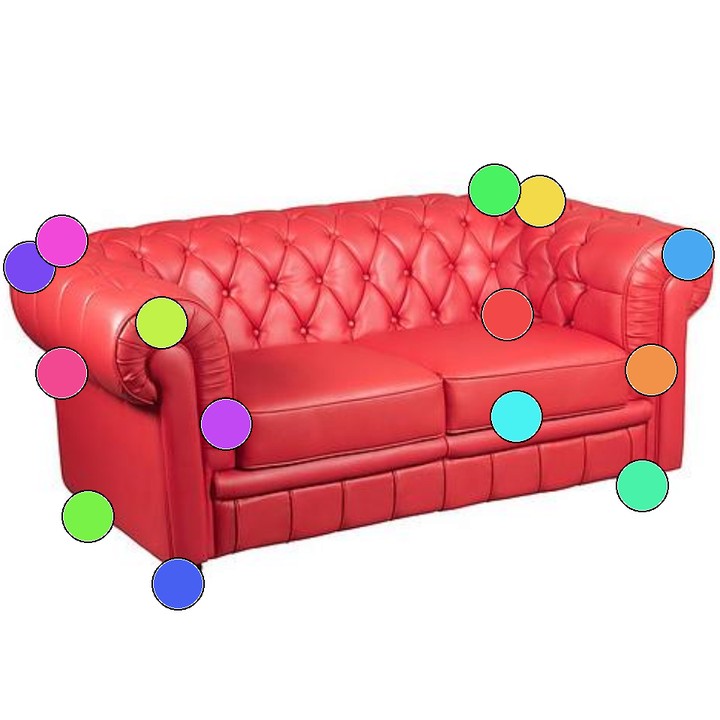}}{\TileImg{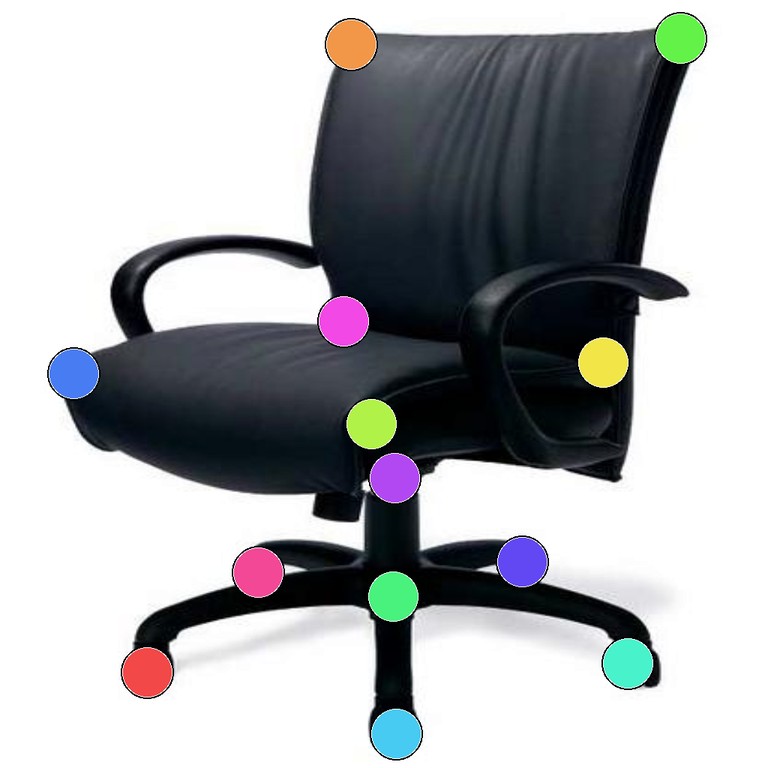}}
    } &
    \parbox{\colw}{\centering
      \GridTwoByTwo
        {\TileImg{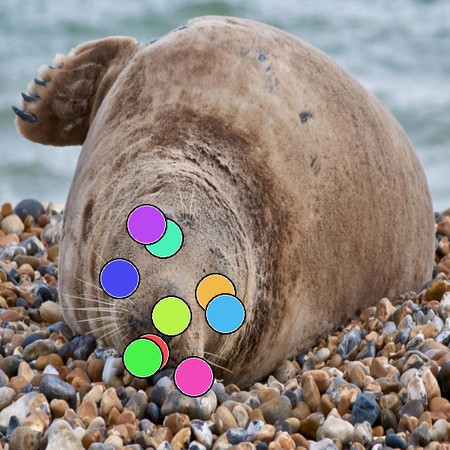}}{\TileImg{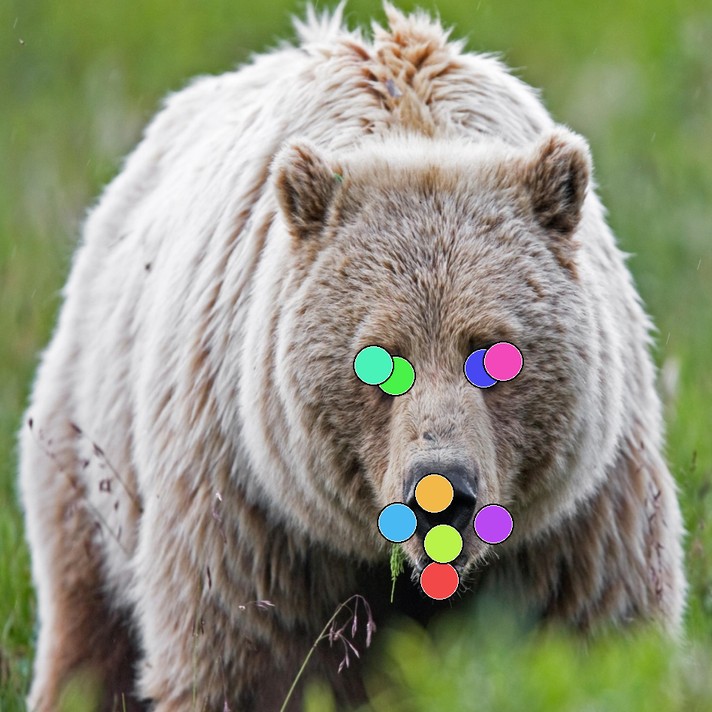}}
        {\TileImg{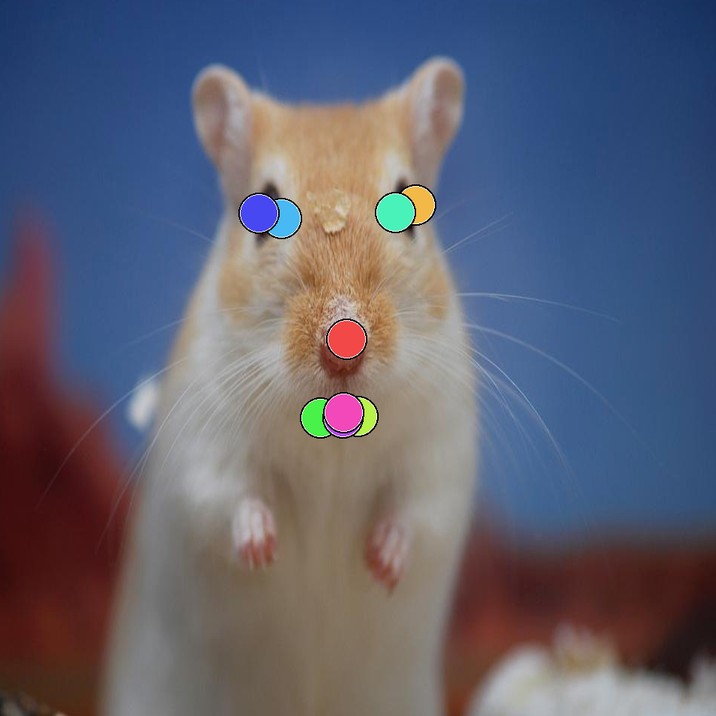}}{\TileDots}
    } \\
    \midrule
    \parbox{\colw}{\centering \scriptsize Categories: 1\\ \scriptsize Avg.\ kpts.: 68} &
    \parbox{\colw}{\centering \scriptsize Categories: 12\\ \scriptsize Avg.\ kpts.: 26} &
    \parbox{\colw}{\centering \scriptsize Categories: 32\\ \scriptsize Avg.\ kpts.: 15} &
    \parbox{\colw}{\centering \scriptsize Categories: 4\\ \scriptsize Avg.\ kpts.: 12} &
    \parbox{\colw}{\centering \scriptsize Categories: 27\\ \scriptsize Avg.\ kpts.: 9} \\
    \bottomrule
  \end{tabular}

  \vspace{-0.7em}
  \caption{\textbf{MP\textendash100 macro-domains at a glance.}}
  \label{fig:mp100-benchmark}
\end{figure}

\paragraph{Benchmarks.}
We first evaluate \ours{} on standard semantic correspondence datasets and benchmarks.
\textbf{SPair-71k}~\cite{Min:2019:Spair} contains with 53.4k train, 5.4k val, and 12.2k test image pairs, across 18 categories with 20 keypoints per image.  
\textbf{PF-PASCAL}~\cite{Ham:2017:PfPascal} provides 2.9k train, 308 test, and 299 val pairs across 20 categories.  
\textbf{AP-10K}~\cite{Yu:2021:AP10k} is a large animal pose dataset with 17 keypoints shared across 54 species, comprising 261k train, 17k val, and 36k test pairs spanning intra-species, cross-species, and cross-family splits.  
To assess generalization, we further evaluate models trained on SPair-71k in a zero-shot setting.  
\textbf{SPair-U}~\cite{Mariotti:2025:Jamais} extends SPair-71k with an average of 4 additional unseen keypoints per category, providing a first test of keypoint-level novelty but remaining limited in scale. To enable a broader evaluation, we introduce a new protocol based on \textbf{MP-100}~\cite{Xu:2022:MP100}, a large-scale pose dataset with $\sim$150k annotated keypoints over 100 object classes.  
We group categories into five macro-domains (\cf \cref{fig:mp100-benchmark}) to evaluate two axes of generalization: \textit{i) unseen keypoints}, with seen categories annotated with new landmarks (\ie, \textit{human face}, which extends the \textit{person} category from 7 annotated points in SPair-71k to 68 facial landmarks, or \textit{apparel items}); and  \textit{ii) unseen categories}, with object types not seen during training (\eg, \textit{home furniture}, \textit{animal face}, \textit{animal body}), with overlaps against SPair-71k removed. Full statistics are provided in the Supplementary Material.

\myparagraph{Evaluation metrics.}  
Following \cite{Zhang:2024:Telling, Zhang:2023:Tale, Mariotti:2025:Jamais, Li:2024:Sd4match}, we use the Percentage of Correct Keypoints (PCK@$\alpha$) as metric, for which a prediction is considered correct if it lies within a distance of \(\alpha \cdot \max(h, w)\) from the ground-truth keypoint; here, \(h\) and \(w\) denote the object bounding-box dimensions. We report per-image PCK averaged over the test set.

\myparagraph{Implementation details.}  
We use DINOv2 ViT-L/14 \cite{Oquab:2023:Dinov2} and train only the upsampling head and AdaptFormer \cite{Chen:2022:AdaptFormer} blocks in the last 12 transformer layers. We optimize with Adam \cite{Kingma:2015:Adam}, use a learning rate of \num{1E-4} and teacher EMA momentum $\beta = \num{0.999}$. We set $\sigma_{\text{max}} = 3$ and $\sigma_{\text{min}} = 1$, and train for 10 epochs with batch size 16. Unless stated otherwise, during training we restrict pixels to object masks derived via SAM~\cite{Kirillov:2023:SAM}, following previous work \cite{Fundel:2025:Distillation, Mariotti:2024:SphericalMaps, Mariotti:2025:Jamais}. At inference, we use Window soft-argmax as in \cite{Zhang:2024:Telling, Mariotti:2025:Jamais}.

\subsection{Comparison on standard benchmarks}
\label{sec:exp_standard}

Table~\ref{tab:benchmark_spair_ap_pfpascal} reports results on the standard in-domain benchmarks: SPair-71k, AP-10K, and PF-PASCAL. \ours{} achieves the highest accuracy, establishing a new state of the art while relying on a \emph{single} DINOv2 encoder. The advantage is striking at the challenging threshold of \pckOne{}, where we outperform Geo-SC \cite{Zhang:2024:Telling} by +\SI{5.3}{} and +\SI{10.0}{} on SPair-71k and AP-10K, respectively. Under the coarser metric \pckTen{}, \ours{} reaches 87.2 on {SPair-71k}, surpassing Geo-SC~\cite{Zhang:2024:Telling} by +4.0 pts.  
On {AP-10K}, \ours{} outperforms prior methods in the intra-species, cross-species, and cross-family splits, with the largest gains in the most challenging cross-family case (+4.9 \pckTen{}).
On {PF-PASCAL}, although close to saturation, \ours{} improves by +5.8 \pckFive{}, indicating consistent fine-grained accuracy. Note that replacing object masks with bounding boxes derived from ground-truth keypoints during training yields virtually identical results. Finally, compared to the previous state-of-the-art Geo-SC, \ours{} remains 3$\times$ smaller and 10$\times$ faster (\cref{fig:teaser}d).

\subsection{Generalization}
\label{sec:exp_generalization}
In \cref{tab:spair_seen_unseen_single,tab:mp100_generalization}, we evaluate models trained with SPair-71k on SPair-U and MP-100, where either the queried keypoints or the object categories have not been seen during training.  
A consistent trend emerges: \textbf{Geo-SC}, the strongest in-domain method (\cf \cref{tab:benchmark_spair_ap_pfpascal}), suffers substantial drops out-of-domain, whereas \textbf{Jamais Vu}~\cite{Mariotti:2025:Jamais}, designed to preserve generalization, degrades less but remains weaker in-domain.  
\textbf{\ours{}} sets state-of-the-art results in \emph{both} scenarios, outperforming Jamais Vu on unseen keypoints and unseen categories, and Geo-SC \cite{Zhang:2024:Telling} on traditional benchmarks.  
This shows that \ours{} does not merely trade in-domain strength for generalization, but \emph{improves both}.
We report \textbf{qualitative examples} of our predictions in \cref{fig:teaser}\textit{a-c}.

\begin{table}[b]
  \centering
  \vspace{-0.5em}
  \caption{\textbf{Generalization on MP-100}~\cite{Xu:2022:MP100}. 
  Per-image \pckTen{} (\%, $\uparrow$) across \emph{unseen keypoints} and semantic \emph{unseen categories}. $^*$~unsupervised methods. Best results \textbf{bold}, \nth{2} best \underline{underlined}.}
  \label{tab:mp100_generalization}
  \tablesize
  \vspace{-0.75em}
  \setlength{\tabcolsep}{2pt}
  \begin{tabularx}{\linewidth}{@{}Xccccc@{}}
    \toprule
     &
    \multicolumn{2}{c}{\textbf{Unseen keypoints}} &
    \multicolumn{3}{c}{\textbf{Unseen categories}} \\
    \cmidrule(lr){2-3}\cmidrule(lr){4-6}
    & \begin{tabular}{@{}c@{}}\HumanFace\\[-1pt]\footnotesize Human\\[-2pt]\footnotesize face\end{tabular}
    & \begin{tabular}{@{}c@{}}\Dress\\[-1pt]\footnotesize Apparel\\[-2pt]\footnotesize items\end{tabular}
    & \begin{tabular}{@{}c@{}}\Elephant \\[-1pt]\footnotesize Animal\\[-2pt]\footnotesize body\end{tabular}
    & \begin{tabular}{@{}c@{}}\TableIcon\\[-1pt]\footnotesize Home\\[-2pt]\footnotesize furniture\end{tabular}
    & \begin{tabular}{@{}c@{}}\AnimalFace\\[-1pt]\footnotesize Animal\\[-2pt]\footnotesize face\end{tabular} \\
    \midrule
    DINOv2 $^*$~\cite{Oquab:2023:Dinov2}     
    & 66.2 & 44.7 & 36.1 & 44.2 & 33.3 \\
    DIFT $^*$~\cite{Tang:2023:Dift}  
    & \underline{87.3} & 48.2 & 31.0 & 46.9 & 26.5 \\
    SD $+$ DINO $^*$~\cite{Zhang:2023:Tale}  
    & 85.3 & \underline{50.2} & 36.1 & 49.2 & 39.6 \\
    GECO \cite{Hartwig:2025:GECO} &
    82.9 & 41.7 & 31.9 & 48.1 & 38.2 \\ 
    Geo-SC~\cite{Zhang:2024:Telling}     
    & 85.2 & 42.9 & 38.9 & 49.6 & \underline{49.2} \\
    Jamais Vu~\cite{Mariotti:2025:Jamais}   
    & 85.5 & 45.7 & \underline{39.3} & \underline{52.7} & 47.7 \\
    \textbf{MARCO} \emph{(ours)}                   & \textbf{87.5} & \textbf{55.9} & \textbf{42.3} & \textbf{60.4} & \textbf{52.6} \\
    \bottomrule
  \end{tabularx}
\end{table}

\begin{table*}[t]
    \centering
    \caption{\textbf{Generalization on SPair-U (Unseen keypoints)} in terms of per-image \pckTen{} (in \%, $\uparrow$). 
    Methods marked with $^*$ are unsupervised. All methods are trained on SPair-71k. Best results are shown in \textbf{bold}; \nth{2} best are \underline{underlined}.}
    \label{tab:spair_seen_unseen_single}
    \vspace{-0.75em}
    \tablesize
    \setlength{\tabcolsep}{3.7pt}
    \begin{tabularx}{\linewidth}{@{}Xcccccccccccccccccc@{\hspace{1.3em}}c@{}}
        \toprule
        \textbf{Method}  &\faIcon{plane} & \faIcon{bicycle} & \faIcon{crow} & \faIcon{ship} & \faIcon{wine-bottle} &
        \faIcon{bus} & \faIcon{car} & \faIcon{cat}
        &\faIcon{chair} & \Cow & \faIcon{dog} &
        \faIcon{horse} & \faIcon{motorcycle} & \faIcon{walking} & \Plant & \Sheep &
        \faIcon{train} & \faIcon{tv} & avg \\
        \midrule
\noalign{
\begin{tikzpicture}[overlay, remember picture]
\draw[line width=0.4pt] (8.03,-3) -- ++(0,3.8cm);
\end{tikzpicture}
}
        DIFT $^*$~\cite{Tang:2023:Dift}  
        & 73.2 & 71.8 & 48.8 & 37.7 & 43.0 & 55.1 & 47.2 & 25.4 & 35.9 & 60.4 & 46.2 & 41.6 & 59.9 & 53.1 & 57.8 & 36.1 & 50.6 & 19.5 & 47.4 \\
        DINOv2 $^*$~\cite{Oquab:2023:Dinov2} 
        & \textbf{88.2} & 75.6 & \textbf{79.0} & 52.9 & 39.8 & 54.1 & 60.0 & 43.9 & 34.8 & 67.2 & 64.6 & 53.6 & \underline{75.8} & \underline{79.1} & 37.8 & 45.6 & \underline{53.3} & 8.4 & 54.9 \\
        DINOv2+SD $^*$~\cite{Zhang:2023:Tale} 
        & \underline{88.0} & \textbf{80.4} & \underline{72.3} & 48.2 & \textbf{47.9} & 62.3 & 61.5 & 44.8 & 45.0 & \textbf{73.0} & 64.7 & 58.2 & \textbf{75.5} & \textbf{80.0} & {62.7} & 46.1 & \textbf{55.9} & {16.9} & 59.4 \\
        \midrule
        DHF~\cite{Luo:2023:DHF} 
        & 71.4 & 58.1 & 39.1 & 35.8 & \underline{44.7} & 74.0 & 40.2 & 33.5 & 27.4 & 52.0 & 50.4 & 41.6 & 56.5 & 51.6 & 41.6 & 30.0 & 42.5 & 14.5 & 43.3 \\
        GECO~\cite{Hartwig:2025:GECO} 
        & 59.8 & 63.3 & 50.8 & 57.7 & 31.5 & 80.6 & 53.7 & \underline{61.1} & 38.6 & \underline{72.4} & \textbf{80.7} & 72.7 & 56.2 & 58.9 & 33.5 & \textbf{59.7} & 40.0 & \underline{41.7} & 55.2 \\
        Geo-SC~\cite{Zhang:2024:Telling} 
        & 80.9 & 71.4 & 51.8 & {65.3} & 36.9 & 91.0 & 70.8 & 55.7 & 36.9 & 55.7 & 79.2 & 53.7 & 66.5 & 62.3 & 61.1 & 39.0 & 39.0 & 17.4 & 56.9 \\
        Jamais Vu~\cite{Mariotti:2025:Jamais} 
        & 80.3 & 74.5 & 70.6 & \underline{67.1} & {40.2} & \textbf{92.9} & \underline{72.7} & 53.8 & \underline{45.8} & {68.5} & 75.3 & \underline{62.0} & 67.8 & 65.4 & \underline{68.1} & 45.4 & {47.9} & 30.5 & \underline{62.4} \\
        \textbf{MARCO} \emph{(ours)} & 86.6 &      \underline{78.2} &      55.8 &      \textbf{71.7} &      36.8 &      \underline{92.5} &      \textbf{79.2} &      \textbf{62.7} &      \textbf{62.9} &      63.9 &      \underline{79.3} &      \textbf{69.6} &      73.7 &      78.1 &      \textbf{74.2} &      \underline{55.4} &      35.8 &      \textbf{51.9} &  \textbf{67.5} \\
    \bottomrule
    \end{tabularx}
    \vspace{-0.5em}
\end{table*}

\myparagraph{SPair-U (unseen keypoints).}
Table~\ref{tab:spair_seen_unseen_single} evaluates transfer to novel keypoints defined on the SPair-71k images.  
Jamais Vu is the strongest baseline, as it learns category-specific 3D canonical templates on the training set.  
\ours{} outperforming Jamais Vu by +5.1, reaching 67.5 \pckTen{}.  

\newcommand{\step}{\hspace*{0.9em}\texttt{+}\;}
\newcommand{\stepp}{\hspace*{1.8em}\texttt{+}\;}  
\newcommand{\steppp}{\hspace*{2.7em}\texttt{+}\;}
\begin{table}[t]
\centering
\caption{\textbf{Ablation studies.} 
Per-image $\text{PCK}$ (\%, $\uparrow$) on \textbf{SPair-71k} and \textbf{SPair-U}. 
SPair-71k evaluates the effect of \textit{architecture} and \textit{sparse supervision}, whereas SPair-U (unseen keypoints) analyzes the generalization effect of our \textit{dense self-supervision}. 
Each component group is assessed independently, keeping the others fixed.}
\label{tab:ablations_unified}
\vspace{-0.75em}

{\tablesize
\setlength{\tabcolsep}{4.5pt}

\begin{tabularx}{\linewidth}{@{}l
  X
  S[table-format=2.1]
  S[table-format=2.1]
  >{\columncolor{black!6}[5pt][0pt]}S[table-format=2.1]@{}}
\toprule
 &  & \multicolumn{2}{c}{\textbf{SPair-71k}} & \cellcolor{white} \textbf{SPair-U} \\
\cmidrule(lr){3-4}\cmidrule(lr){5-5}
& & {@0.01} & {@0.10} & \cellcolor{white} {@0.10} \\
\midrule
\multirow{3}{*}{\makecell[l]{Architecture}}
& DINOv2 (frozen)                & 6.3  & 53.9 & \cellcolor{white} 54.9 \\
& \step Adapter                  & 14.4 & 76.9 & \cellcolor{white} 60.7 \\ 
& \cellcolor{black!6} \stepp Upsampling \emph{(ours)} & \cellcolor{black!6} \bfseries 27.0 & \cellcolor{black!6} \bfseries 87.2 & \bfseries 67.5 \\
\midrule
\noalign{
  \begin{tikzpicture}[overlay, remember picture]
    \draw[line width=0.4pt] (-2.6,-3) -- ++(0,4.25cm);
  \end{tikzpicture}
}
\multirow{3}{*}{\makecell[l]{Supervision\\with GT}}
& Fixed $\sigma{\,=\,}3$             & 19.5 & 86.8 & \cellcolor{white} 66.6 \\
& Fixed $\sigma{\,=\,}1$             & \bfseries 27.8 & 81.0 & \cellcolor{white} 62.2 \\
 & \cellcolor{black!6} Coarse-to-fine \emph{(ours)} & \cellcolor{black!6} 27.0 & \cellcolor{black!6} \bfseries 87.2 & \bfseries 67.5 \\
\midrule
\multirow{5}{*}{\makecell[l]{Dense\\self-\\distillation}}
& Ours w/o dense loss            & 26.8 & 85.6 & \cellcolor{white} 41.8 \\
& \step Raw flow                 & 25.2 & 84.6 & \cellcolor{white} 49.6 \\
& \step MNN matches              & 26.5 & 85.9 & \cellcolor{white} 52.5 \\
& \stepp Delaunay \& warp        & 26.7 & 86.2 & \cellcolor{white} 64.7 \\
& \cellcolor{black!6} \steppp GT anchor \emph{(ours)} & \cellcolor{black!6} \bfseries 27.0 & \cellcolor{black!6} \bfseries 87.2 & \bfseries 67.5 \\
\bottomrule
\end{tabularx}
} 

\vspace{-0.5em}
\end{table}

\myparagraph{MP-100 (unseen keypoints and unseen categories).}
\cref{tab:mp100_generalization} evaluates models trained on SPair-71k under the challenging MP-100 protocol, where both the keypoints and the categories differ from those seen during training.
On the \textit{Human face} split, featuring 68 novel, densely annotated landmarks, \ours{} surpasses Jamais Vu by +2.0 PCK.
On \textit{Apparel items} and \textit{Home furniture}, we further improve over Jamais Vu by +10.2 and +7.7 PCK, respectively.
Finally, on \textit{Animal body} and \textit{Animal face}, which together span more than 30 unseen species, \ours{} achieves gains of +3.0 and +4.9 PCK. These results show that the representation learned by our dense self-supervision transfers reliably to new landmark vocabularies and entirely new semantic domains. Methods relying on category-specific structure (\eg, Jamais Vu) lose accuracy once the taxonomy shifts, and models optimized purely for in-domain results (\eg, Geo-SC) degrade even more sharply. In contrast, \ours{} maintains strong generalization on both axes, unseen keypoints and unseen categories, beyond standard benchmarks.

\subsection{Ablation studies}
We conduct ablations on SPair-71k and SPair-U to disentangle the contribution of each component. On SPair-71k, we consider both coarse (\pckTen{}) and fine (\pckOne{}) thresholds to assess the effect of our \textit{coarse-to-fine} supervision strategy. This joint evaluation mirrors the two goals of MARCO: precise localization and strong generalization. For more ablation studies, please refer to Supp. Material.

\myparagraph{Architecture.}
Starting from a frozen DINOv2 encoder, lightweight adapters provide strong accuracy improvements (+8.1$\text{@}0.01$, +23.0$\text{@}0.10$ on SPair-71k). Similarly, the upsampling head provides substantial gains (+12.6$\text{@}0.01$, +10.3$\text{@}0.10$), highlighting the value of restoring sub-patch structure. Together, these two lightweight components turn DINOv2 into a precise correspondence backbone.

\myparagraph{Supervision with GT.}
A fixed narrow target ($\sigma{\,=\,}1$) maximizes fine localization (27.8$\text{@}0.01$) but degrades coarse accuracy (81.0$\text{@}0.10$) and unseen generalization (62.2$\text{@}0.10$). A wide target ($\sigma{\,=\,}3$) does the opposite (19.5$\text{@}0.01$, 86.8$\text{@}0.10$). Our coarse-to-fine schedule balances these regimes, reaching 27.0$\text{@}0.01$ (near $\sigma{\,=\,}1$), 87.2$\text{@}0.10$ (near $\sigma{\,=\,}3$), trading ${\sim}1$ point relative to the best fixed settings while avoiding their respective collapses. 

\myparagraph{Dense self-distillation.}
Removing dense self-supervision causes a severe collapse in generalization (-25.7 \pckTen{} on SPair-U).
Using raw semantic flow partially recovers accuracy (+7.8 pts.), but is vulnerable to erroneous matches, ultimately harming in-domain accuracy. Our solution instead builds \textit{reliable pseudo-labels}. 
Using only mutual nearest neighbors as pseudo-labels recovers part of the missing structure (52.5 \pckTen{}), yet still underperforms the frozen encoder (-2.4 pts.). Delaunay-based densification then converts sparse matches into coherent, locally smooth correspondence fields, yielding a further +12.2 \pckTen{} (64.7).
Finally, anchoring clusters with ground-truth keypoints removes symmetric and implausible motions, reaching 67.5 \pckTen{}, without affecting in-domain results.

\section{Conclusion}
\label{sec:conclusion}
We revisited semantic correspondence from the dual perspective of precision and generalizability. We show that a single DINOv2 backbone, trained to progressively refine spatial detail, can deliver accurate pixel-level correspondences. At the same time, leveraging the reliable structure that emerges in feature space during learning enables the model to generalize robustly to unseen landmarks and novel taxonomies.
Looking forward, \ours{} suggests a path toward correspondence models that remain accurate and efficient while adapting to the open-ended variety of keypoints required in tasks such as image editing and pose estimation. A promising next step is to further reduce the dependence on sparse supervision to better exploit large-scale, unconstrained data, ultimately allowing models to infer geometric relationships directly from visual data at scale.

{\small \inparagraph{Acknowledgments.} 
Claudia Cuttano was supported by the Sustainable Mobility Center (CNMS), which received funding from the European Union Next Generation EU (Piano Nazionale di Ripresa e Resilienza (PNRR), Missione 4 Componente 2 Investimento 1.4 ``Potenziamento strutture di ricerca e creazione di `campioni nazionali di R\&S' su alcune Key Enabling Technologies'') with grant agreement no.\ CN\_00000023.  
Stefan Roth has received funding from the European Research Council (ERC) under the European Union's Horizon 2020 research and innovation programme (grant agreement No.~866008).
Further, he was supported by the DFG under Germany's Excellence Strategy (EXC-3057/1 ``Reasonable Artificial Intelligence'', Project No.~533677015). We acknowledge the CINECA award under the ISCRA initiative, for the availability of high performance computing resources.
We acknowledge the support of the European Laboratory for Learning and Intelligent Systems (ELLIS).

{
    \small
    \bibliographystyle{ieeenat_fullname}
    \bibliography{bibtex/short, bibtex/references}
}

\clearpage
\setcounter{section}{0}
\renewcommand\thesection{\Alph{section}}
\setcounter{page}{1}
\pagenumbering{roman}
\maketitlesupplementary

In this appendix, we provide additional insights into \ours{}, together with further experimental analyses. Specifically:
\begin{itemize}
    \item \textbf{General applicability of our approach}. In \cref{sec:geo_flow}, we show that our dense self-distillation loss can produce coherent and generalizable representations when applied to previous state-of-the-art methods.
    \item \textbf{Results with pre-training}. In \cref{sec:ap_pretrain}, we present a variant of our model pre-trained on the AP-10K dataset, following recent approaches~\cite{Zhang:2024:Telling, Mariotti:2025:Jamais}.
    \item \textbf{Additional experiments}. \Cref{sec:ablations_supp} provides further analyses of \ours{}, including studies on adapter placement and dimensionality, comparisons with full fine-tuning, a progressive ablation of the proposed components, and analyses of pseudo-label coverage, pseudo-label noise, and hyperparameter robustness.
    \item \textbf{Use of object masks}. In \cref{sec:training}, we discuss the role of object masks in our pipeline, show that they can be removed with negligible impact on results, and compare the supervision assumptions of different methods.
    \item \textbf{Compute details}. In \cref{sec:compute}, we report model size and inference speed, and describe the protocol used to measure computational performance.
    \item \textbf{Pseudo-code of dense self-distillation}. In \cref{sec:pseudocode}, we summarize the proposed flow-anchoring procedure and self-distillation loss in algorithmic form.
    \item \textbf{Details on MP-100}. In \cref{sec:benchmark_mp100}, we provide details of the proposed benchmark, including the curation protocol, evaluation splits, and extended quantitative results.
    \item \textbf{Qualitative examples}. In \cref{sec:qualitatives}, we present additional qualitative results for \ours{}.
\end{itemize}

\section{Broad Applicability of our Dense Self-Distillation via Flow Anchoring}
\label{sec:geo_flow}

A central motivation behind \ours{} is to enhance the \emph{geometric coherence} of learned feature representations, producing features that vary smoothly across the object surface to generalize to \emph{unseen keypoints and categories}, despite the sparsity of landmark supervision. Our dense self-distillation via flow-anchoring is designed for this purpose: by mining reliable correspondences emerging in the feature space and propagating them across the object via piecewise-affine interpolation, the training objective encourages the backbone to maintain smooth, consistent geometry beyond the annotated regions. We raise the question of \emph{whether our self-distillation strategy is tied to the MARCO architecture, or whether it constitutes a general training principle that can benefit other correspondence pipelines}. To assess its generality, we apply our dense self-distillation loss to the state-of-the-art Geo-SC model \cite{Zhang:2024:Telling}. We compare three variants:
\emph{(i)} the original Geo-SC,
\emph{(ii)} Jamais Vu \cite{Mariotti:2025:Jamais}, which augments Geo-SC with a loss mapping object points to a learned canonical manifold, and
\emph{(iii)} Geo-SC augmented with our flow-anchoring loss.
Because all variants share the same backbone, inference pipeline, and base loss, this experiment isolates the contribution of the auxiliary loss.

\begin{table}[t]
\centering
\caption{\textbf{General applicability of our self-distillation loss.} By training the state-of-the-art Geo-SC model with our dense self-distillation objective, we markedly improve its generalization to unseen keypoints on SPair-U. In \ours{}, we integrate this objective with a coarse-to-fine supervised loss, reaching state-of-the-art results while being smaller and faster. Per-image PCK (in \%, $\uparrow$) on SPair-71k, and SPair-U (unseen keypoints); models trained on SPair-71k. Best results \textbf{bold}, \nth{2} best \underline{underlined}.}
\label{tab:geo_flow}
\tablesize
\vspace{-0.65em}
\setlength{\tabcolsep}{2.8pt}
\begin{tabularx}{\linewidth}{@{}XcS[table-format=2.1]cccccc@{}}
\toprule
 &  &   \multicolumn{3}{c}{\textbf{SPair-71k}} 
& \multicolumn{3}{c}{\textbf{SPair-U}}  \\
\cmidrule(lr){3-5}\cmidrule(lr){6-8}
\textbf{Method} & \textbf{Encoders} &
0.01 & 0.05 & 0.10 
& 0.01 & 0.05 & 0.10 \\
\midrule
\noalign{
  \begin{tikzpicture}[overlay, remember picture]
    \draw[line width=0.4pt] (-0.03,-2.05) -- ++(0.,2.15cm);
    \draw[line width=0.4pt] (2.1,-2.05) -- ++(0.,2.15cm);
  \end{tikzpicture}
}
Jamais Vu~\cite{Mariotti:2025:Jamais}& SD+DINO 
& 20.5 & 71.9 & 82.5  & 4.2 & 37.8 & 62.4 \\
Geo\text{-}SC ~\cite{Zhang:2024:Telling}& SD+DINO  
& \underline{21.7} & 72.8 & 83.2 & 3.9 & 35.4 & 56.9 \\
~\tikz[baseline=1ex]{\draw[->, thick] (0., 2.8ex) -- (0,1.4ex) -- (1em,1.4ex);} {\small{+ our dense self-distillation loss}}& SD+DINO  
& 20.8 & \underline{73.0} & \underline{83.6} &  4.2 & \underline{38.1} & \underline{63.4} \\
\midrule
\textbf{MARCO} \emph{(ours)} & DINOv2  
& \textbf{27.0} & \textbf{77.6} & \textbf{87.2} &  \textbf{4.7} & \textbf{41.7} & \textbf{67.5} \\
\bottomrule
\end{tabularx}
\vspace{-0.5em}
\end{table}

\myparagraph{Effect of adding our loss to Geo-SC.}
As shown in \cref{tab:geo_flow}, adding our flow-anchoring loss consistently improves Geo-SC. On SPair-U, which evaluates generalization to unseen keypoints, Geo-SC improves from 56.9 to 63.4 \pckTen{}, outperforming Jamais Vu, which requires monocular depth prediction from an external model for lifting keypoints in 3D. Importantly, our loss does not degrade in-domain accuracy: on SPair-71k, Geo-SC increases from 83.2 to 83.6 \pckTen{}, whereas Jamais Vu slightly impairs results (82.5). Despite these improvements, Geo-SC enhanced with our loss remains below the accuracy levels of \ours{}, which achieves 87.2 \pckTen{} on SPair-71k and 67.5 \pckTen{} on SPair-U. This gap highlights the importance of coupling flow anchoring with the full \ours{} framework: coarse-to-fine supervision, adapter-based feature enrichment, and efficient sub-patch refinement, which together strengthen spatial fidelity in DINOv2. Finally, \ours{} attains these improvements with a \emph{single} DINOv2 encoder (323M parameters), compared to the 950M-parameter SD{+}DINO dual-encoder used by Geo-SC and Jamais Vu, being approximately $10\times$ faster.

\section{Pre-training on AP-10k}
\label{sec:ap_pretrain}

Recent correspondence works have increasingly adopted an additional pre-training stage on the AP-10K dataset. This strategy was first introduced by Geo-SC~\cite{Zhang:2024:Telling}, which also proposed the AP-10K benchmark itself. Subsequent methods, including Jamais Vu~\cite{Mariotti:2025:Jamais}, followed this practice and incorporated AP-10K pre-training into their pipelines, making it a common component of recent evaluation protocols. To ensure a fair and direct comparison with these approaches, we therefore train a variant of \ours{} that includes the same AP-10K pre-training stage. The corresponding results are reported in \cref{tab:ap_pretrain}.
Like prior works, our method benefits from pre-training. Notably, \emph{(i)} simply adding more data does not aid generalization on SPair-U (\cf GECO, Geo-SC), and \emph{(ii)} our SPair-only model outperforms previous baselines pre-trained on AP-10k. 

\begin{table}[t]
\centering
\caption{\textbf{Impact of pre-training.} Following recent works, we show a variant of our model pretrained on AP-10k and compare to other works in the same setting. $\dagger$ indicates a model jointly trained on SPair-71k and AP-10k, rather than pre-trained and then fine-tuned.
Per-image PCK (in \%, $\uparrow$) on SPair-71k, and SPair-U (unseen keypoints). Best results \textbf{bold}, \nth{2} best \underline{underlined}}
\label{tab:ap_pretrain}
\tablesize
\vspace{-0.65em}
\setlength{\tabcolsep}{3.2pt}
\begin{tabularx}{\linewidth}{@{}Xccccccc@{}}
\toprule
 &  &   \multicolumn{3}{c}{\textbf{SPair-71k}} 
& \multicolumn{3}{c}{\textbf{SPair-U}}  \\
\cmidrule(lr){3-5}\cmidrule(lr){6-8}
\textbf{Method} & \textbf{Encoders} &
0.01 & 0.05 & 0.10 
& 0.01 & 0.05 & 0.10 \\
\midrule
\noalign{
  \begin{tikzpicture}[overlay, remember picture]
    \draw[line width=0.4pt] (-0.08,-3.2) -- ++(0.,3.3cm);
    \draw[line width=0.4pt] (2.12,-3.2) -- ++(0.,3.3cm);
  \end{tikzpicture}
}
\multicolumn{2}{@{}l}{\textit{Train on SPair-71k}} &&&& \\
~~Jamais Vu~\cite{Mariotti:2025:Jamais}& SD+DINO 
& 20.5 & 71.9 & 82.5  & 4.2 & 37.8 & 62.4 \\
~~Geo\text{-}SC ~\cite{Zhang:2024:Telling}& SD+DINO  
& 21.7 & 72.8 & 83.2 & 3.9 & 35.4 & 56.9 \\
~~\textbf{MARCO} \emph{(ours)} & DINOv2  & 
\underline{27.0} & \underline{77.6} & \underline{87.2} & \underline{4.7} & \underline{41.7} & 67.5 \\

\multicolumn{2}{@{}l}{\textit{w/ Pre-train on AP-10k}} &&&&\\
~~GECO~\cite{Hartwig:2025:GECO} $\dagger$ & DINOv2 & 14.2 & 59.6 & 73.6 & 3.2 & 32.1 & 55.2 \\ 
~~Jamais Vu~\cite{Mariotti:2025:Jamais}& SD+DINO 
& 20.9 & 73.1 & 85.4 & 4.2 & 37.8 & \underline{66.1} \\
~~Geo\text{-}SC ~\cite{Zhang:2024:Telling}& SD+DINO  
& 22.0 & 75.3 & 85.6 &  4.1 & 35.5 & 57.1 \\
~~\textbf{MARCO} \emph{(ours)} & DINOv2  & 
\textbf{28.5} & \textbf{78.3} & \textbf{87.3} & \textbf{5.0} & \textbf{44.2} & \textbf{69.7}  \\
\bottomrule
\end{tabularx}
\vspace{-0.5em}
\end{table}

\section{Additional Ablations}
\label{sec:ablations_supp}

\paragraph{Adapter design study.}
We first analyze the architectural design choices underlying \ours{}. 
\cref{tab:ablations} reports controlled ablations studying 
\emph{(i)} different fine-tuning strategies, 
\emph{(ii)} alternative parameter-efficient adaptation mechanisms, 
\emph{(iii)} adapter placement across the transformer depth, and 
\emph{(iv)} the dimensionality of the adapter bottleneck. 
Experiments are conducted using the same training schedule and evaluated on both SPair-71k and SPair-U. 
\noindent Three consistent trends emerge. 
First, fully fine-tuning the DINOv2 backbone harms generalization. 
While full fine-tuning increases SPair-71k accuracy to 67.0 \pckTen{}, the accuracy on SPair-U drops sharply to 43.9, compared to 54.9 when the backbone is kept frozen. 
This confirms that the pre-trained representation should remain largely frozen to preserve its semantic structure and generalization ability.
A lighter strategy that fine-tunes only the QKV projections performs better (81.1 \pckTen{} on SPair-71k and 59.6 on SPair-U), but still underperforms parameter-efficient adaptation. Second, among parameter-efficient strategies, AdaptFormer provides the best trade-off between in-domain accuracy and generalization.
Classical Adapters \cite{Houlsby:2019:Adapter} achieve 86.9 \pckTen{} on SPair-71k and 65.7 on SPair-U, while LoRA \cite{Hu:2022:LoRA} obtains slightly lower results (85.2 and 65.6 \pckTen{}, respectively).
AdaptFormer \cite{Chen:2022:AdaptFormer} improves this balance, reaching 87.2 \pckTen{} on SPair-71k and 67.5 on SPair-U. Third, the placement of adapters across the transformer depth plays an important role.
Adaptation is most effective when applied to the upper transformer blocks (Layers 12--24), where high-level semantic features are formed. Applying adapters earlier in the network (\eg, Layers 3--24) slightly reduces accuracy, while restricting adaptation to only the last few blocks also degrades accuracy.
Finally, the dimensionality of the adapter bottleneck controls the balance between model capacity and regularization.
A mid-sized bottleneck ($\times$0.5, used in \ours{}) achieves the best overall trade-off, while both larger ($\times$1) and smaller ($\times$0.3 or $\times$0.1) bottlenecks slightly reduce accuracy.

\begin{table}[t]
\centering
\caption{\textbf{Additional ablations} comparing fine-tuning \vs adaptation strategies, adapter placement, and bottleneck dimension.
All results reported as PCK@0.10 (in \%, $\uparrow$).}
\label{tab:ablations}
\tablesize
\vspace{-0.5em}
\setlength{\tabcolsep}{4pt}
\begin{tabularx}{\linewidth}{@{}>{\columncolor{black!8}[0pt][5pt]}X>{\columncolor{black!8}}c>{\columncolor{black!8}[5pt][0pt]}c@{}}
\toprule
\cellcolor{white} \textbf{Setting} & \cellcolor{white} \textbf{SPair-71k} & \cellcolor{white} \textbf{SPair-U} \\
\midrule
\cellcolor{white} DINOv2 \textit{frozen} & \cellcolor{white} 53.9 & \cellcolor{white} 54.9 \\
\midrule
\multicolumn{3}{@{}l}{\textit{Fine-tuning strategies}} \\
\cellcolor{white} \hspace{1em}Full FT      & \cellcolor{white} 67.0 & \cellcolor{white} 43.9 \\
\cellcolor{white} \hspace{1em}QKV-only FT  & \cellcolor{white} 81.1 & \cellcolor{white} 59.6 \\
\midrule
\multicolumn{3}{l}{\textit{Adapter type}} \\
\cellcolor{white} \hspace{1em}Adapter~\cite{Houlsby:2019:Adapter}  & \cellcolor{white} 86.9 & \cellcolor{white} 65.7 \\
\cellcolor{white} \hspace{1em}LoRA~\cite{Hu:2022:LoRA} & \cellcolor{white} 85.2 & \cellcolor{white} 65.6 \\
\hspace{1em}AdaptFormer~\cite{Chen:2022:AdaptFormer}  (\textbf{ours})& 87.2 & 67.5 \\
\midrule
\multicolumn{3}{@{}l}{\textit{Adapter placement (AdaptFormer)}} \\
\cellcolor{white} \hspace{1em}Layers 3--24   & \cellcolor{white} 85.5 & \cellcolor{white} 64.5 \\
\cellcolor{white} \hspace{1em}Layers 6--24   & \cellcolor{white} 85.6 & \cellcolor{white} 65.3 \\
\cellcolor{white} \hspace{1em}Layers 9--24   & \cellcolor{white} 86.2 & \cellcolor{white} 66.0 \\
\cellcolor{white} \hspace{1em}Layers 12--24  & \cellcolor{white} 86.7 & \cellcolor{white} 66.7 \\
\hspace{1em}Layers 15--24 (\textbf{ours})   & 87.2 & 67.5 \\
\cellcolor{white} \hspace{1em}Layers 18--24  & \cellcolor{white} 87.3 & \cellcolor{white} 65.3 \\
\cellcolor{white} \hspace{1em}Layers 21--24  & \cellcolor{white} 84.9 & \cellcolor{white} 63.8 \\
\midrule
\multicolumn{3}{@{}l}{\textit{Adapter bottleneck size (AdaptFormer, Layers 12--24)}} \\
\cellcolor{white} \hspace{1em}Bottleneck $\times$ 1  & \cellcolor{white} 86.2 & \cellcolor{white} 67.1 \\
\hspace{1em}Bottleneck $\times$ 0.5 (\textbf{ours})  & 87.2 & 67.5 \\
\cellcolor{white} \hspace{1em}Bottleneck $\times$ 0.3  & \cellcolor{white} 86.8 & \cellcolor{white} 66.5 \\
\cellcolor{white} \hspace{1em}Bottleneck $\times$ 0.1  & \cellcolor{white} 83.7 & \cellcolor{white} 65.7 \\
\bottomrule
\end{tabularx}
\end{table}

\begin{table*}[t]
\centering
\caption{
\textbf{Ablation of architectural and training components.}
Per-image PCK (in~\%, $\uparrow$) on SPair-71k (seen keypoints) and SPair-U (unseen keypoints).
Adapters and feature upsampling are added to a frozen DINOv2 backbone, while training objectives progressively include standard supervision (InfoNCE + $\ell_2$), the proposed coarse-to-fine loss, and dense self-distillation.
}
\tablesize
\vspace{-0.65em}
\setlength{\tabcolsep}{4pt}
\begin{tabularx}{\textwidth}{@{}X c c c c c c c@{}}
\toprule
\multirow{2}{*}{Architecture} & 
\multicolumn{3}{c}{\textbf{Training objective}} & 
\multicolumn{2}{c}{\textbf{SPair-71k (seen keypoints)}} & 
\multicolumn{2}{c@{}}{\textbf{SPair-U (unseen keypoints)}} \\
\cmidrule(lr){2-4} \cmidrule(lr){5-6} \cmidrule(lr){7-8}
& InfoNCE + $\ell_2$ \cite{Zhang:2024:Telling} & Coarse-to-fine & Dense Loss & \pckOne{} & \pckTen{} & \pckOne{} & \pckTen{} \\
\midrule

DINOv2 (frozen) & \xmark & \xmark & \xmark & 6.3 & 53.9 & 3.3 & 54.9 \\

\specialrule{0.1pt}{1.0pt}{1pt}

\multirow{3}{*}{Adapter + Upsample}
& \checkmark & \xmark & \xmark & 20.0 & 78.9 & 1.9 & 39.7 \\
& \xmark & \checkmark & \xmark & 26.8 & 85.6 & 2.1 & 42.0 \\
& \xmark & \checkmark & \checkmark & \textbf{27.0} & \textbf{87.2} & \textbf{4.7} & \textbf{67.5} \\

\bottomrule
\end{tabularx}
\label{tab:ablations_architecture}
\end{table*}

\myparagraph{Contribution of architectural and training components.}
\Cref{tab:ablations_architecture} complements the ablation analysis in the main paper (\cref{tab:ablations_unified}). 
While the ablation in the main paper evaluates each component independently, this table presents a \emph{progressive ablation} where the architectural and training components of \ours{} are introduced sequentially. Starting from a frozen DINOv2 backbone, adding the proposed adapters and feature upsampling and training with the objective of Geo-SC~\cite{Zhang:2024:Telling}, \ie InfoNCE+$\ell_2$, improves accuracy from 6.3 to 20.0 \pckOne{} and from 53.9 to 78.9 \pckTen{} on SPair-71k, while relying on a \emph{single} DINOv2 encoder instead of the DINOv2+Stable Diffusion pipeline used by prior work. Replacing this objective with the proposed coarse-to-fine supervision further improves localization precision to 26.8 \pckOne{} and 85.6 \pckTen{}. Finally, adding dense self-distillation substantially improves generalization to unseen keypoints, increasing SPair-U accuracy from 42.0 to 67.5 \pckTen{}, while also improving SPair-71k to 87.2 \pckTen{}.

\myparagraph{Pseudo-label coverage and noise.}
The dense self-distillation objective propagates correspondences across the object surface, extending supervision beyond the sparse annotated keypoints. As shown in \cref{tab:ablations_unified} in the main paper, enabling dense self-distillation already increases accuracy on unseen keypoints from 41.8 to 64.7 \pckTen{}, producing pseudo-labels that densely cover the object surface, \ie about \(17\mathrm{k}\) correspondences per object on average in SPair-71k. However, our goal is to maximize pseudo-label quality rather than raw coverage. Anchoring clusters using the GT keypoints improves accuracy to 67.5 \pckTen{}, while retaining an average coverage of about \(13\mathrm{k}\) correspondences. To estimate the quality of the pseudo-labels, we measure their sensitivity to noise by injecting Gaussian perturbations into the pseudo-label coordinates. As reported in \cref{tab:pseudo_noise}, accuracy remains stable for perturbations up to \(\sigma=5\) pixels and only begins to degrade around \(\sigma=10\) pixels, suggesting that the intrinsic noise of the pseudo-labels remains below roughly \(10\) pixels.

\myparagraph{Hyperparameters.}
Our solution is hyperparameter-free. In the flow-anchoring stage, flow vectors are grouped into \(k\) clusters to identify regions with coherent motion. While this step could require selecting the number of clusters, we instead initialize clustering with a large value (\eg, \(k=15\)) and use the Bayesian Information Criterion (BIC) to merge clusters with statistically consistent motion patterns. As shown in \cref{tab:clustering_sensitivity}, this over-segmentation followed by BIC-based merging automatically determines a suitable number of clusters, avoiding the need to tune \(k\).

\section{Use of Object Masks}
\label{sec:training}
Recent works in semantic correspondence frequently leverage instance masks to concentrate feature matching on foreground regions. 
In unsupervised settings, masks are used to suppress background responses when mining feature matches~\cite{Dunkel:2025:DIY, Fundel:2025:Distillation, Mariotti:2024:SphericalMaps}. 
Supervised methods also rely on masks, either to map object pixels to canonical 3D templates~\cite{Mariotti:2025:Jamais} or to restrict optimal-transport domains and sampling to the foreground region~\cite{Hartwig:2025:GECO}. 
Similarly, Geo-SC performs mask-based pose alignment during inference through pose augmentation~\cite{Zhang:2024:Telling}. Masks are obtained by prompting SAM \cite{Kirillov:2023:SAM} with annotated keypoints. In \ours{}, we follow this common practice: importantly, masks are used \emph{only} as a weak spatial prior during pseudo-label generation, not as a direct supervision signal.
Specifically, masks enter our pipeline in a single step:
\begin{quote}
    After extracting mutual nearest-neighbor (MNN) matches, we restrict candidate locations to pixels inside the object mask. The MNN matches are used to construct the Delaunay triangulation for flow estimation. 
\end{quote}

\noindent Their role is, therefore, identical to prior SOTA methods, acting purely as a spatial prior that filters out background regions. However, we observe that, in the absence of masks, we can simply derive a tight bounding box from the ground-truth keypoints and restrict MNN mining to this region.  As reported in \cref{tab:benchmark_spair_singlecol_full}, the accuracy difference between the two variants is negligible:
\begin{itemize}
    \item on SPair-71k, from 27.0 to 26.6 \pckOne{} and from 87.2 to 86.7 \pckTen{};
    \item on SPair-U, from 67.5 to 66.9 \pckTen{}.
\end{itemize}
In all cases, \ours{} \emph{without SAM-based masks} still exceeds all prior methods by a considerable margin. This shows that masks merely provide a mild foreground prior during pseudo-label mining and can be removed with almost no degradation. In other words, \ours{} performs competitively even without any supervision beyond the sparse landmarks. For completeness, \cref{tab:benchmark_spair_singlecol_full} summarizes the supervision used by each competing approach (keypoints, object masks, and depth) together with their accuracy on SPair-71k and SPair-U. 

\newcolumntype{C}{>{\centering\arraybackslash}X}

\begin{table}[t]
\centering
\caption{
\textbf{Pseudo-label noise estimation.}
SPair-U \pckTen{} (\%, $\uparrow$) when Gaussian noise with standard deviation $\sigma$ (px) is added to pseudo-label coordinates. Performance degrades near $\sigma{=}10$.
}
\label{tab:pseudo_noise}
\vspace{-0.5em}
\tablesize
\setlength{\tabcolsep}{4pt}
\renewcommand{\arraystretch}{0.9}

\begin{tabularx}{\linewidth}{@{}XCCCCC@{}}
\toprule
$\sigma$ (px) & 0 & 0.5 & 1 & 5 & 10 \\
\midrule
SPair-U & 67.5\unc{0.2} & 67.7\unc{0.3} & 67.2\unc{0.2} & 67.1\unc{0.4} & 65.0\unc{0.5} \\
\bottomrule
\end{tabularx}

\end{table}
\begin{table}[t]
\centering
\caption{\textbf{Clustering sensitivity.} Performance on SPair-U (\pckTen{}, in \%, $\uparrow$) for different initial numbers of clusters \(k\). Initializing with a larger \(k\) and merging clusters using BIC yields the best result, avoiding the need to tune \(k\).}
\label{tab:clustering_sensitivity}
\vspace{-0.5em}
\tablesize
\setlength{\tabcolsep}{4pt}
\renewcommand{\arraystretch}{0.9}

\begin{tabularx}{\linewidth}{@{}XCCCCC@{}}
\toprule
$k$ & 3 & 5 & 10 & 15 & $15+\mathrm{BIC}$ \\
\midrule
SPair-U & 66.6 & 66.9 & 66.5 & 66.0 & \textbf{67.5} \\
\bottomrule
\end{tabularx}

\end{table}
\begin{table}[t]
\centering
\caption{\textbf{Comparison of supervision levels and acuracy.}
Per-image PCK (in~\%, $\uparrow$) on SPair-71k and SPair-U.
All recent methods use SAM masks during training. 
Geo-SC uses masks during inference ($\dagger$). 
Jamais Vu additionally requires depth supervision. 
\ours{} gives accurate results even with no supervision beyond the sparse keypoints.
}
\label{tab:benchmark_spair_singlecol_full}
\tablesize
\vspace{-0.5em}
\setlength{\tabcolsep}{3pt}

\begin{tabular}{lccccccc}
\toprule
 & \multicolumn{3}{c}{\textbf{Supervision}} 
 & \multicolumn{2}{c}{\textbf{SPair-71k}}
 & \multicolumn{2}{c}{\textbf{SPair-U}} \\
\cmidrule(lr){2-4} \cmidrule(lr){5-6} \cmidrule(lr){7-8}
\textbf{Method} 
& {Keypoint} & {Mask} & {Depth} 
& 0.01 & 0.10 
& 0.01 & 0.10 \\
\midrule
\noalign{
  \begin{tikzpicture}[overlay, remember picture]
    \draw[line width=0.4pt] (1.03,-3.42) -- ++(0.,3.5cm);
    \draw[line width=0.4pt] (2.5,-3.42) -- ++(0.,3.5cm);
  \end{tikzpicture}
}
\multicolumn{4}{l}{\textit{Unsupervised / Weakly Supervised}}&& \\
DistillDIFT~\cite{Fundel:2025:Distillation} 
& -- & ✓ & -- 
& 8.9 & 65.1 
& -- & -- \\

DIY-SC~\cite{Dunkel:2025:DIY}  
& -- & ✓ & -- 
& 10.1 & 71.6 
& -- & -- \\

\midrule
\multicolumn{4}{l}{\textit{Supervised}} &&& \\
Geo-SC~\cite{Zhang:2024:Telling} 
& ✓ & ${\dagger}$ & -- 
& 21.7 & 83.2 
& 3.9 & 56.9 \\

GECO~\cite{Hartwig:2025:GECO} 
& ✓ & ✓ & -- 
& 14.2 & 73.6 
& 3.2 & 55.2 \\

Jamais Vu~\cite{Mariotti:2025:Jamais} 
& ✓ & ✓ & ✓ 
& 20.5 & 82.5 
& \underline{4.2} & 62.4 \\

\textbf{MARCO} \emph{(ours)} 
& ✓ & ✓ & -- 
& \textbf{27.0} & \textbf{87.2} 
& \textbf{4.7} & \textbf{67.5} \\

\textbf{MARCO} \emph{(ours)} 
& ✓ & -- & -- 
& \underline{26.6} & \underline{86.7} 
& 4.3 & \underline{66.9} \\
\bottomrule
\end{tabular}

\vspace{-0.5em}
\end{table}

\section{Computational Cost}
\label{sec:compute}
\Cref{tab:compute} compares the computational footprint of \ours{} with respect to state-of-the-art solutions. Geo-SC~\cite{Zhang:2024:Telling} and Jamais Vu~\cite{Mariotti:2025:Jamais} use the same dual-encoder architecture combining Stable Diffusion and DINOv2, resulting in \(950\)M parameters. In contrast, \ours{} relies on a single DINOv2 backbone with adapters, totaling \(323\)M parameters. We measure inference speed on a single NVIDIA RTX4090 GPU. To ensure a fair comparison, all methods follow the same evaluation protocol: feature extraction at \(840\)p resolution, batched reference--target image pairs, and the same soft-argmax prediction \cite{Zhang:2024:Telling}. For Geo-SC and Jamais Vu we use the original Geo-SC implementation. Under these conditions, \ours{} runs at \(8.3\) FPS, compared to \(0.85\) FPS for Geo-SC and Jamais Vu, corresponding to roughly a \(10\times\) speedup.

\begin{table}[t]
\centering
\caption{\textbf{Compute comparison.} Model size and inference speed measured on an RTX4090 GPU. All methods use the same evaluation protocol: feature extraction at 840p, batched reference--target pairs, and the same soft-argmax keypoint prediction.}
\label{tab:compute}
\vspace{-0.5em}
\tablesize
\setlength{\tabcolsep}{12pt}
\begin{tabularx}{\columnwidth}{@{}Xccc@{}}
\toprule
Method & Backbone & Params & FPS $\uparrow$ \\
\midrule
Geo-SC~\cite{Zhang:2024:Telling} & SD + DINOv2 & 950M & 0.85 \\
Jamais Vu~\cite{Mariotti:2025:Jamais} & SD + DINOv2 & 950M & 0.85 \\
\ours{} & DINOv2 & 323M & \textbf{8.30} \\
\bottomrule
\end{tabularx}
\vspace{-0.5em}
\end{table}

\begin{table*}
\centering
\caption{\textbf{Statistics on MP-100 benchmark.}}
\label{tab:spair_mp100_categories}
\vspace{-0.5em}
\tablesize
\setlength{\tabcolsep}{4pt}
\begin{tabularx}{\textwidth}{@{}p{0.21\textwidth}|p{0.21\textwidth}|X@{}}
\toprule
\textbf{SPair-71k training categories} &
\textbf{Split type} &
\textbf{MP-100 categories used in our benchmark} \\
\midrule

\begin{tabular}[t]{@{}cl@{}}
ID & Category \\
\midrule
 1 & aeroplane \\
 2 & bicycle \\
 3 & bird \\
 4 & boat \\
 5 & bottle \\
 6 & bus \\
 7 & car \\
 8 & cat \\
 9 & chair \\
10 & cow \\
11 & dog \\
12 & horse \\
13 & motorbike \\
14 & person \\
15 & plant \\
16 & sheep \\
17 & train \\
18 & tv/monitor \\
\end{tabular}
&

\begin{tabular}[t]{@{}l@{}}
\textbf{Unseen keypoints} \\
\hspace*{0.5em}Human face \\
\hspace*{1.4em}Categories: 1 \\
\hspace*{1.4em}Avg. keypoints/pair: 68 \\
\hspace*{1.4em}\#Keypoint definitions: 68 \\[0.3em]

\hspace*{0.5em}Apparel item \\
\hspace*{1.4em}Categories: 12 \\
\hspace*{1.4em}Avg. keypoints/pair: 27 \\
\hspace*{1.4em}\#Keypoint definitions: 282 \\[0.6em]

\textbf{Unseen categories} \\
\hspace*{0.5em}Animal body \\
\hspace*{1.4em}Categories: 32 \\
\hspace*{1.4em}Avg. keypoints/pair: 15 \\
\hspace*{1.4em}\#Keypoint definitions: 101 \\[0.3em]

\hspace*{0.5em}Animal face \\
\hspace*{1.4em}Categories: 26 \\
\hspace*{1.4em}Avg. keypoints/pair: 9 \\
\hspace*{1.4em}\#Keypoint definitions: 9 \\[0.3em]

\hspace*{0.5em}Home furniture \\
\hspace*{1.4em}Categories: 4 \\
\hspace*{1.4em}Avg. keypoints/pair: 12 \\
\hspace*{1.4em}\#Keypoint definitions: 45 \\
\end{tabular}
&

\begin{tabular}[t]{@{}l@{}}
\textbf{Human face:} \\
\hspace*{0.5em}Human face. \\[0.35em]

\textbf{Apparel item:} \\
\hspace*{0.5em}short sleeved outwear, short sleeved shirt, skirt, short sleeved dress, vest dress, \\
\hspace*{0.5em}long sleeved dress, long sleeved outwear, long sleeved shirt, \\
\hspace*{0.5em}sling, sling dress, trousers, vest. \\[0.35em]

\textbf{Animal body:} \\
\hspace*{0.5em}macaque body, locust body, fly body, antelope body, cheetah body, fox body, \\
\hspace*{0.5em}leopard body, panther body, rat body, squirrel body, beaver body, deer body, \\
\hspace*{0.5em}giraffe body, lion body, pig body, rhino body, weasel body, bison body, \\
\hspace*{0.5em}elephant body, gorilla body, otter body, polar bear body, skunk body, wolf body, \\
\hspace*{0.5em}hippo body, bobcat body, raccoon body, hamster body, panda body, \\
\hspace*{0.5em}rabbit body, spider monkey body, zebra body. \\[0.35em]

\textbf{Animal face:} \\
\hspace*{0.5em}alpaca face, californian sea lion face, chipmunk face, ferret face, gibbons face, \\
\hspace*{0.5em}guanaco face, proboscis monkey face, arctic wolf face, camel face, \\
\hspace*{0.5em}common warthog face, gentoo penguin face, grey seal face, klipspringer face, \\
\hspace*{0.5em}fennec fox face, blackbuck face, cape buffalo face, dassie face, gerbil face, \\
\hspace*{0.5em}grizzly bear face, olive baboon face, quokka face, bonobo face, capybara face, \\
\hspace*{0.5em}fallow deer face, onager face, pademelon face. \\[0.35em]

\textbf{Home furniture:} \\
\hspace*{0.5em}couch, table, bed, swivel chair. \\
\end{tabular}
\\
\bottomrule
\end{tabularx}
\vspace{-0.5em}
\end{table*}

\begin{figure*}[h]
\centering
\setlength{\tabcolsep}{1.5pt}
\begin{tabular}{ccccc}
\scriptsize Human face & \scriptsize Apparel item &
\scriptsize Animal face & \scriptsize Animal body & \scriptsize Home furniture \\[2pt]

\includegraphics[width=0.19\linewidth]{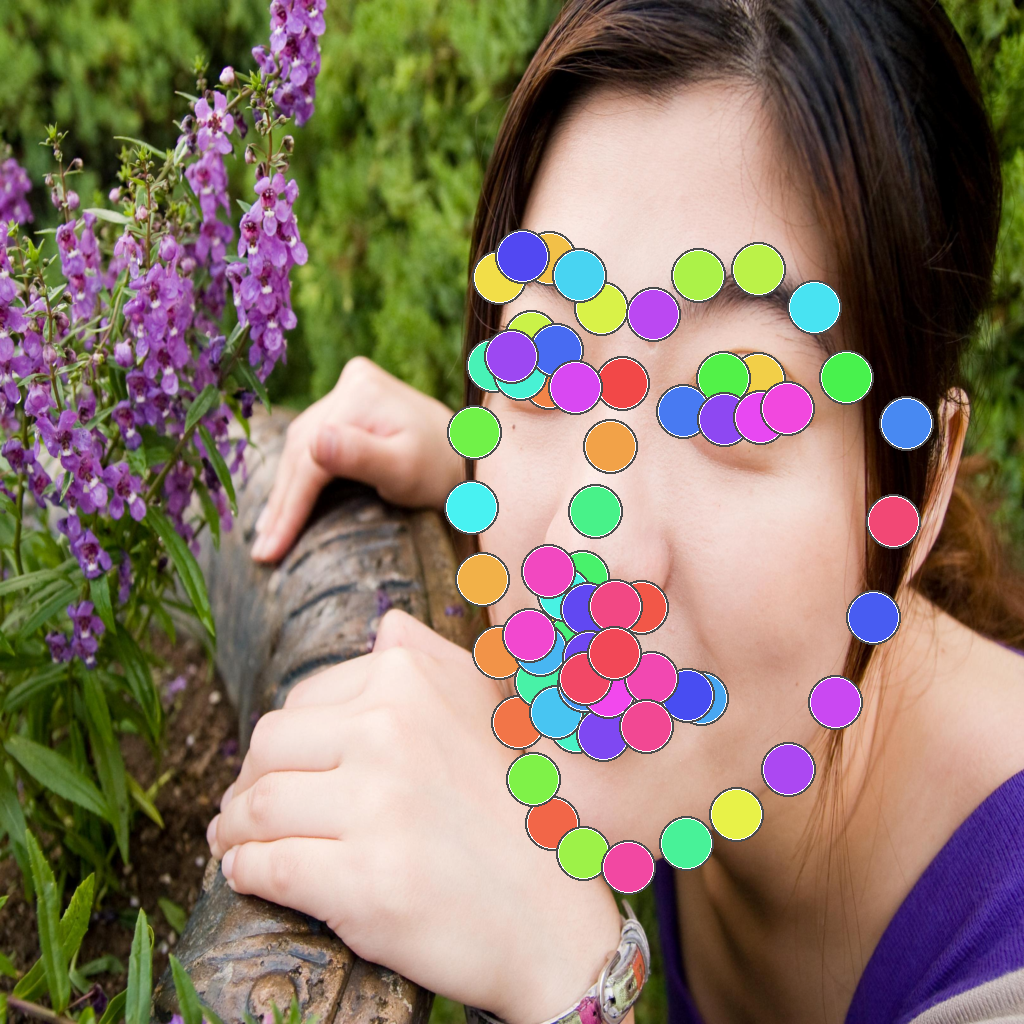} &
\includegraphics[width=0.19\linewidth]{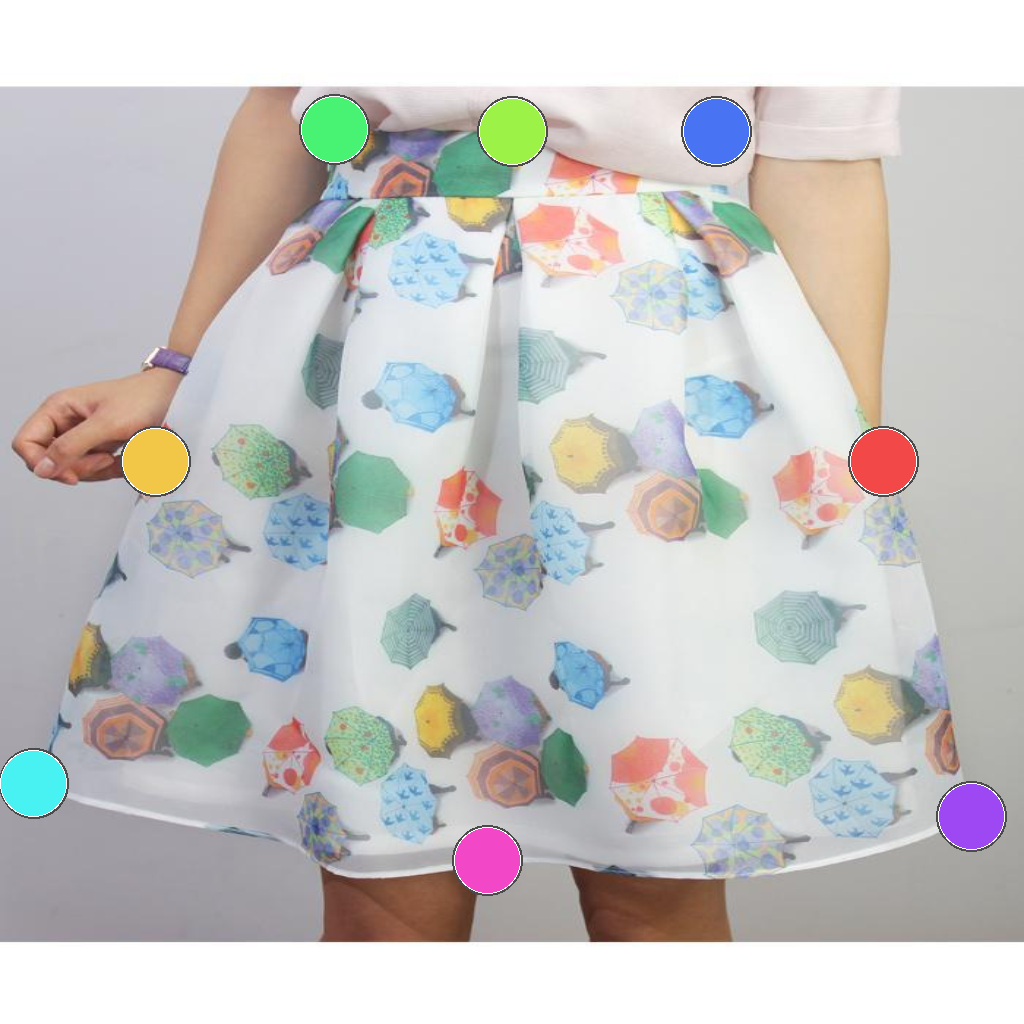} &
\includegraphics[width=0.19\linewidth]{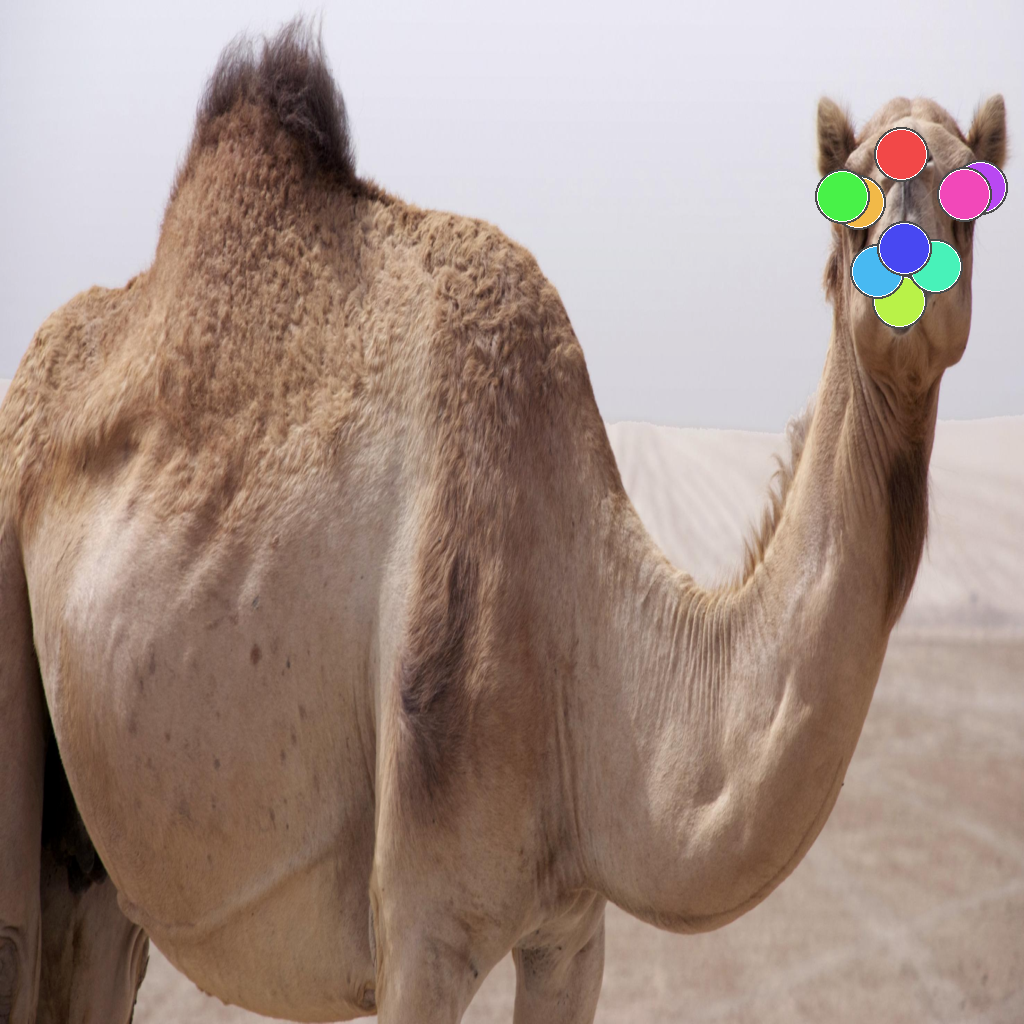} &
\includegraphics[width=0.19\linewidth]{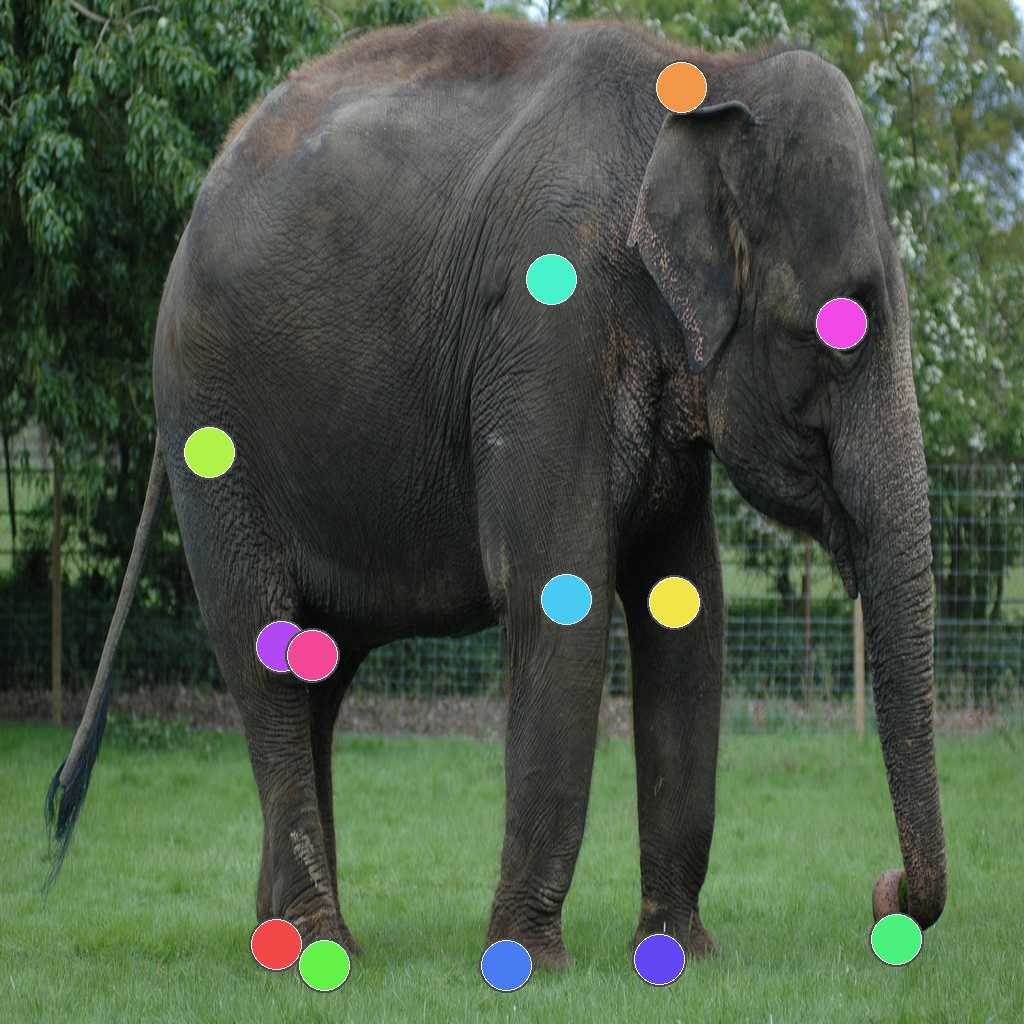} &
\includegraphics[width=0.19\linewidth]{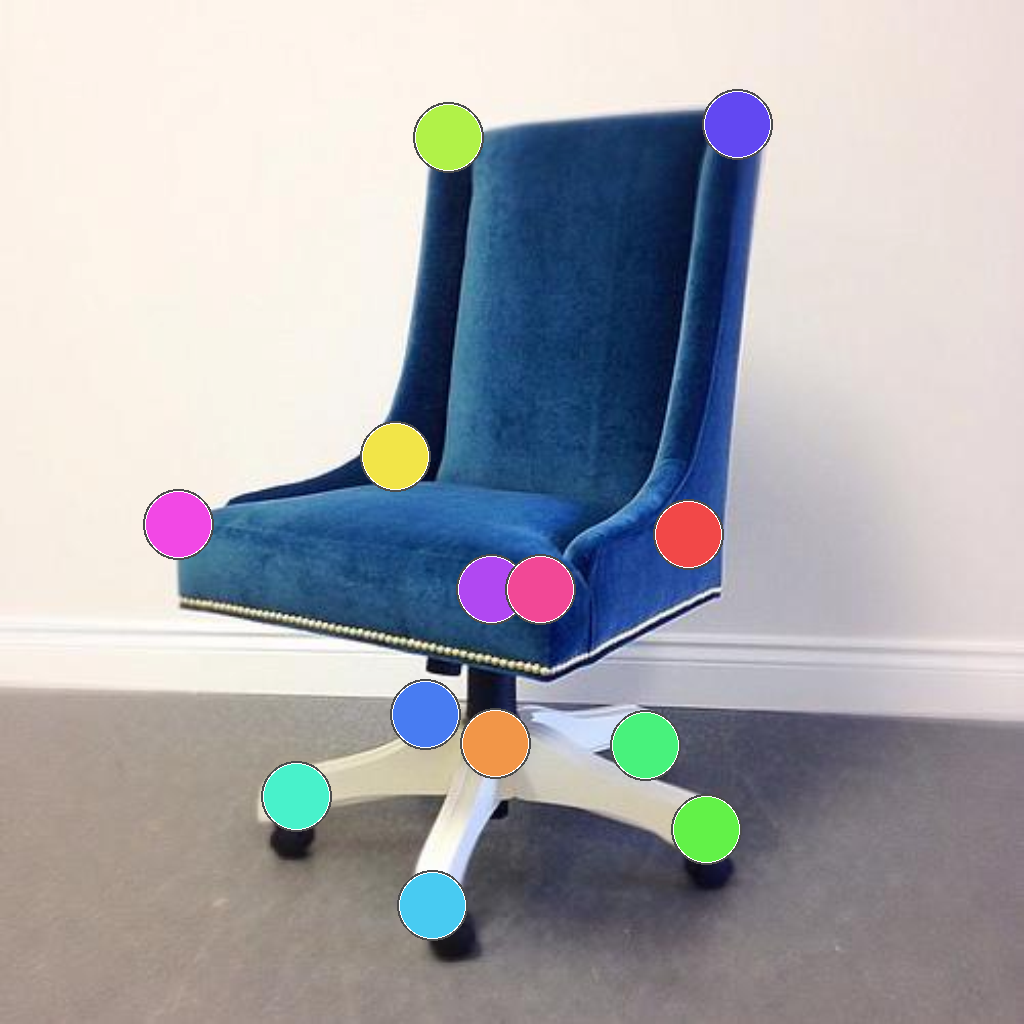} \\

\includegraphics[width=0.19\linewidth]{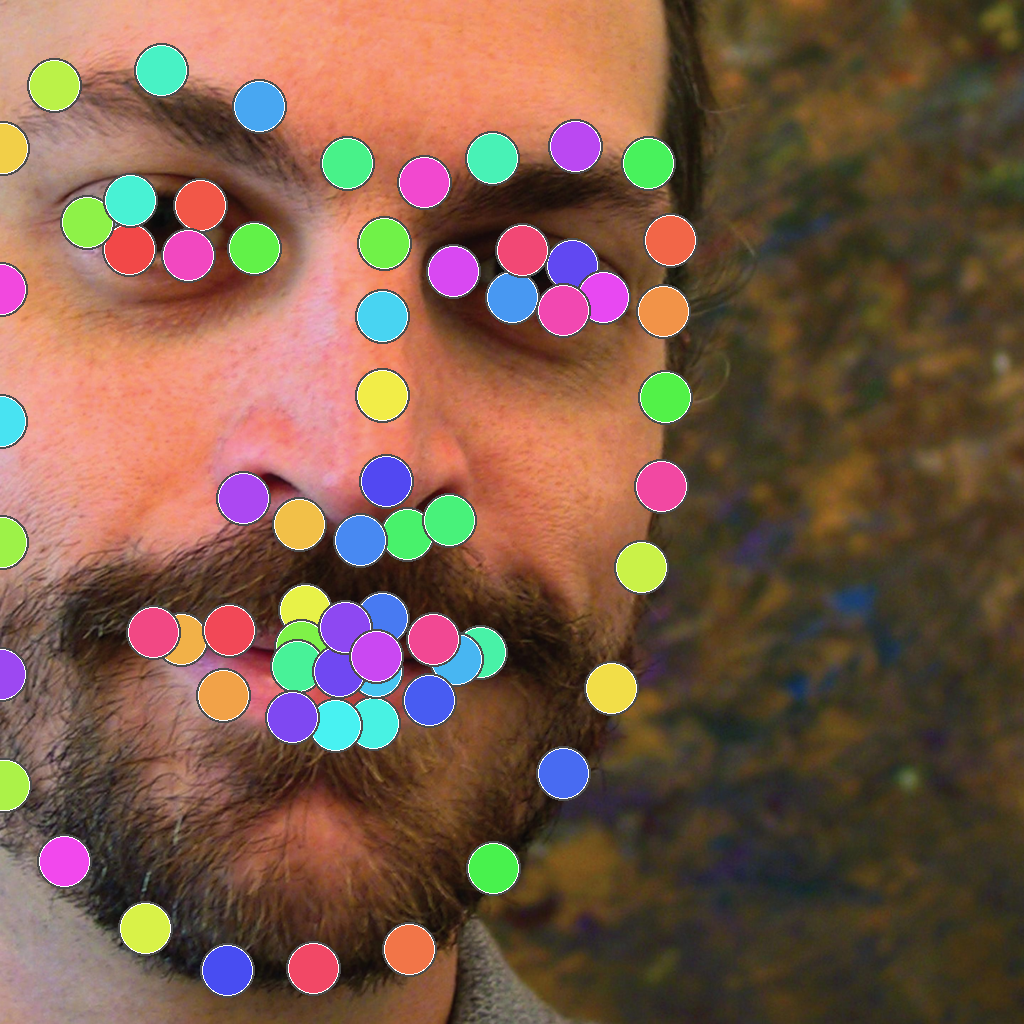} &
\includegraphics[width=0.19\linewidth]{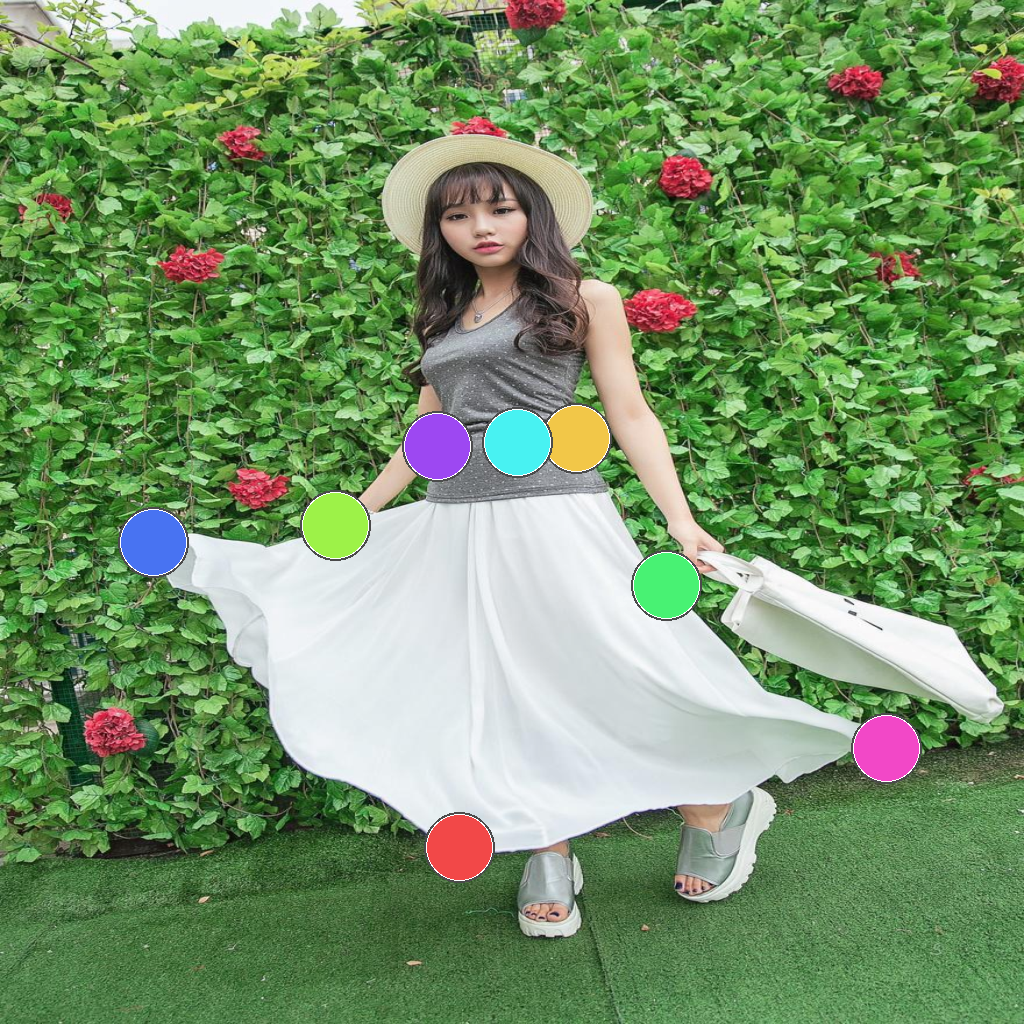} &
\includegraphics[width=0.19\linewidth]{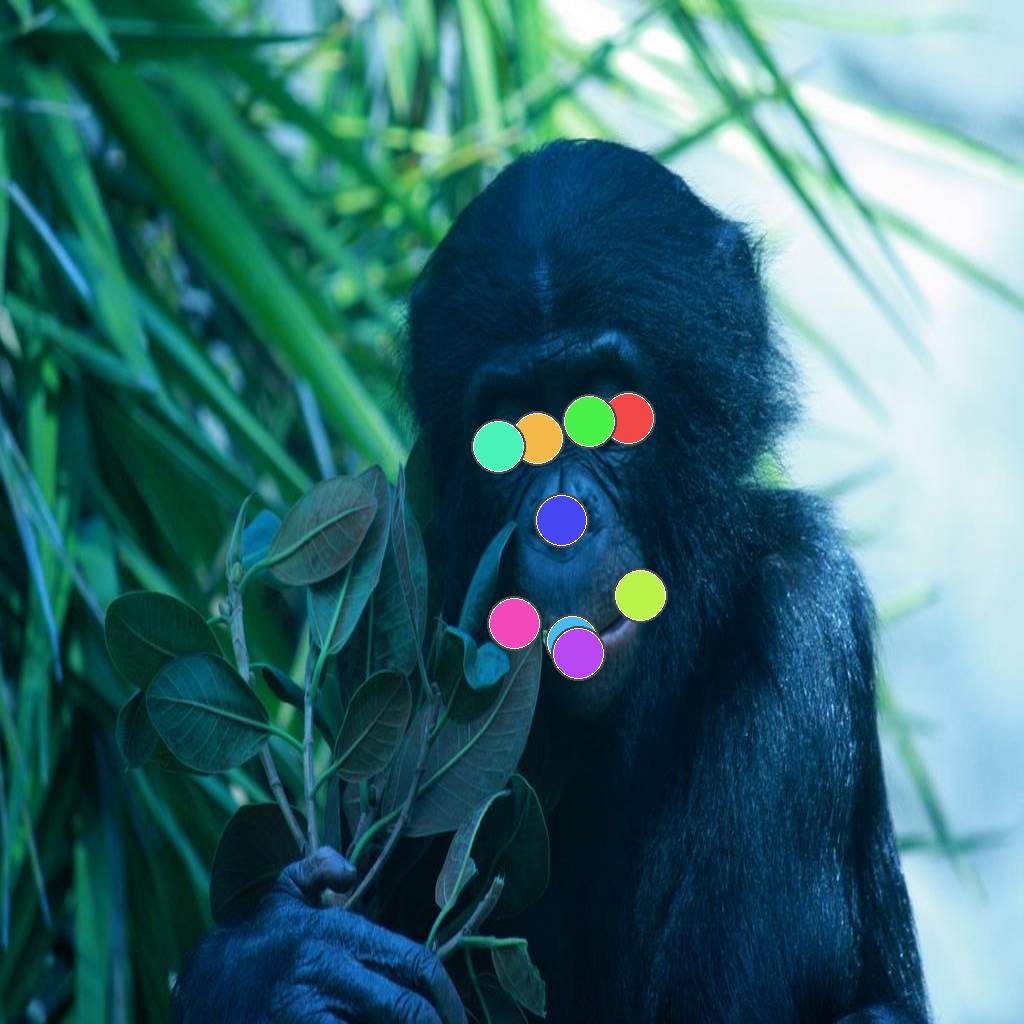} &
\includegraphics[width=0.19\linewidth]{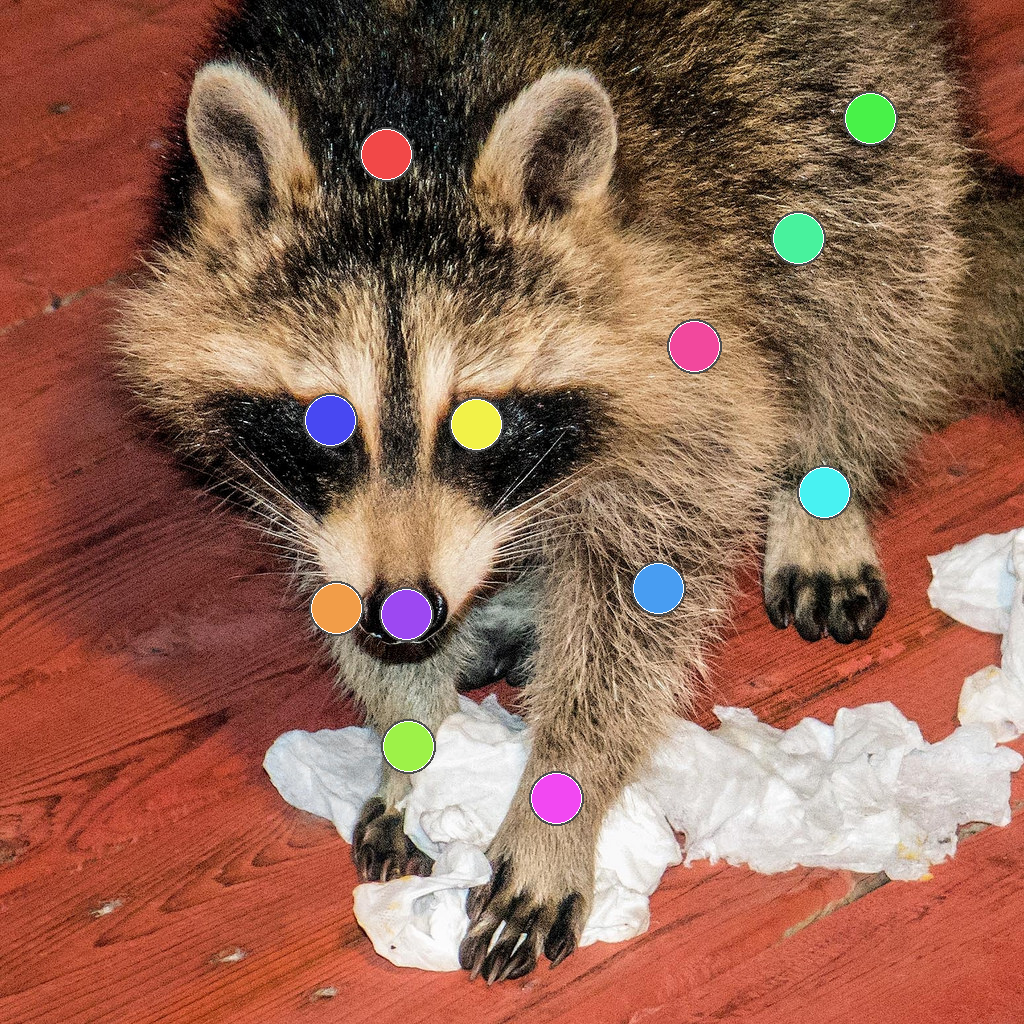} &
\includegraphics[width=0.19\linewidth]{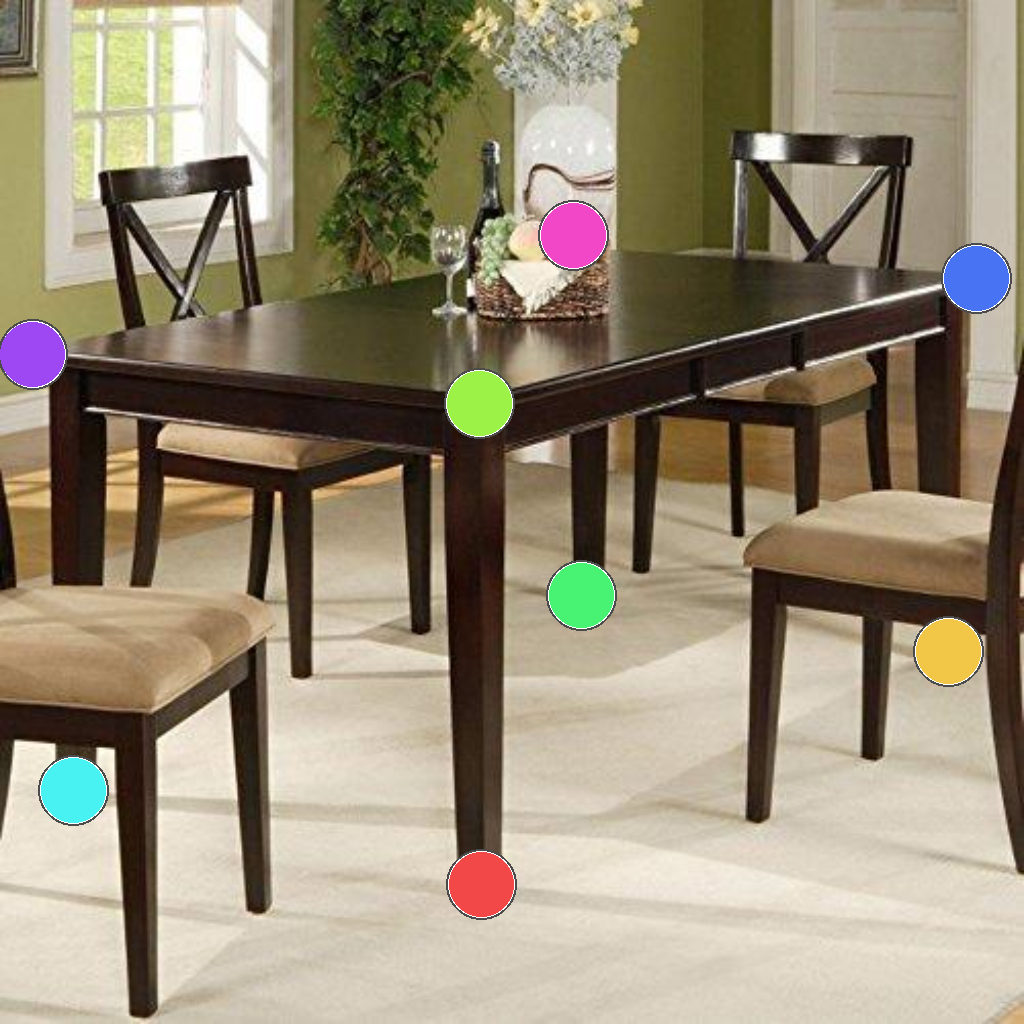} \\

\includegraphics[width=0.19\linewidth]{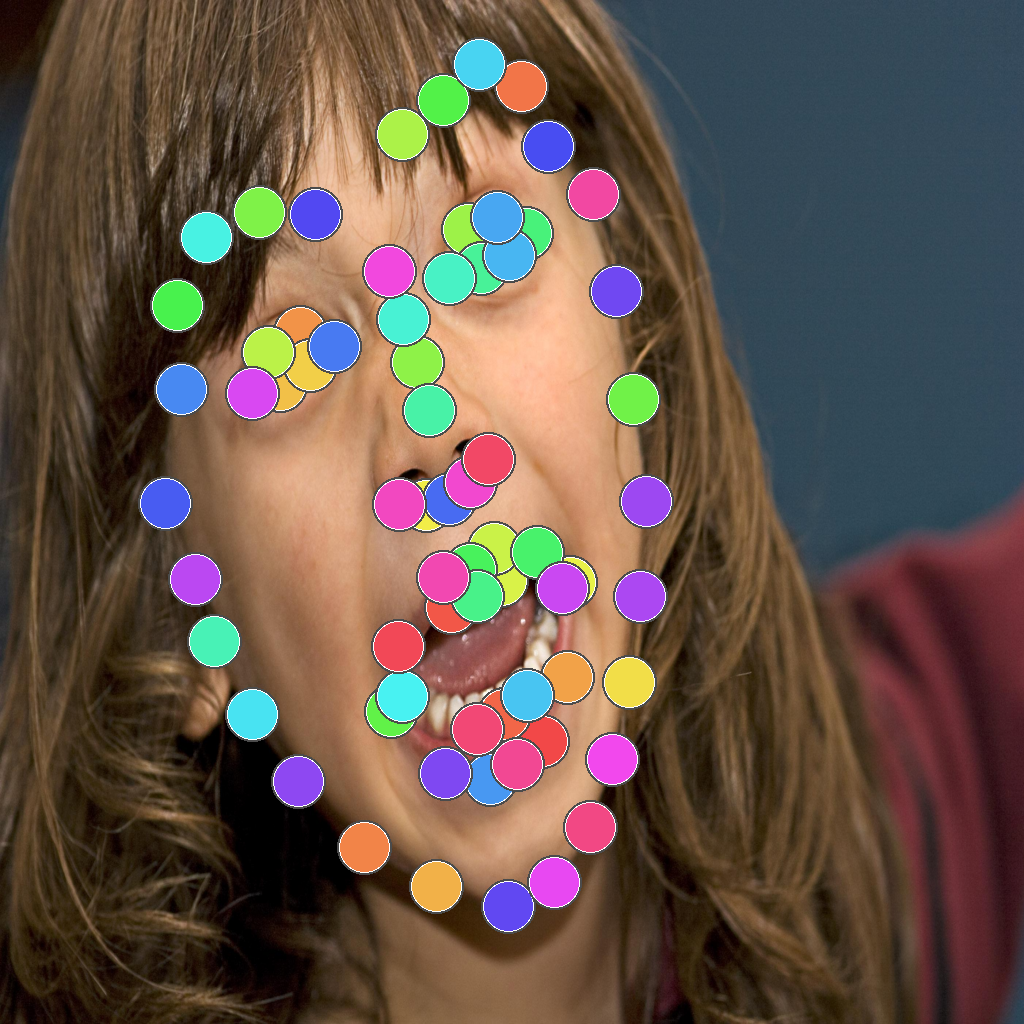} &
\includegraphics[width=0.19\linewidth]{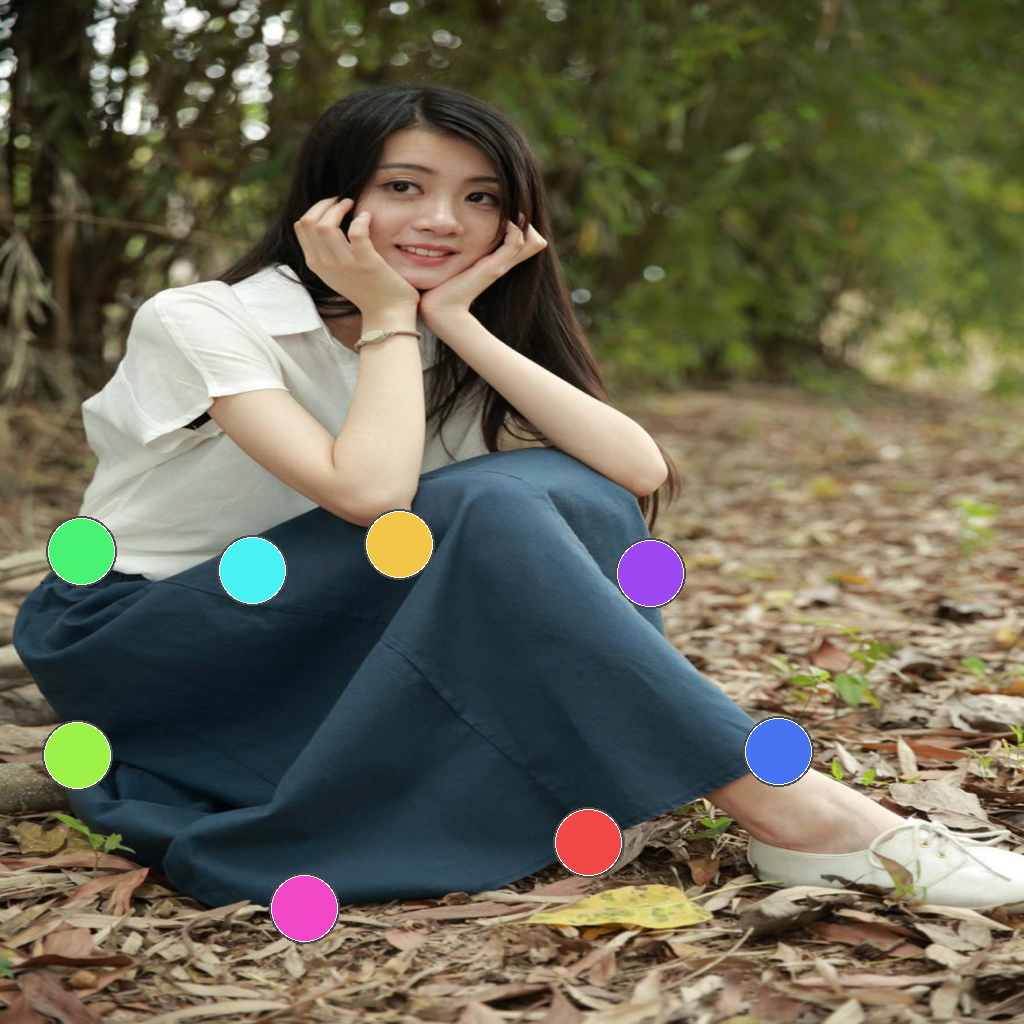} &
\includegraphics[width=0.19\linewidth]{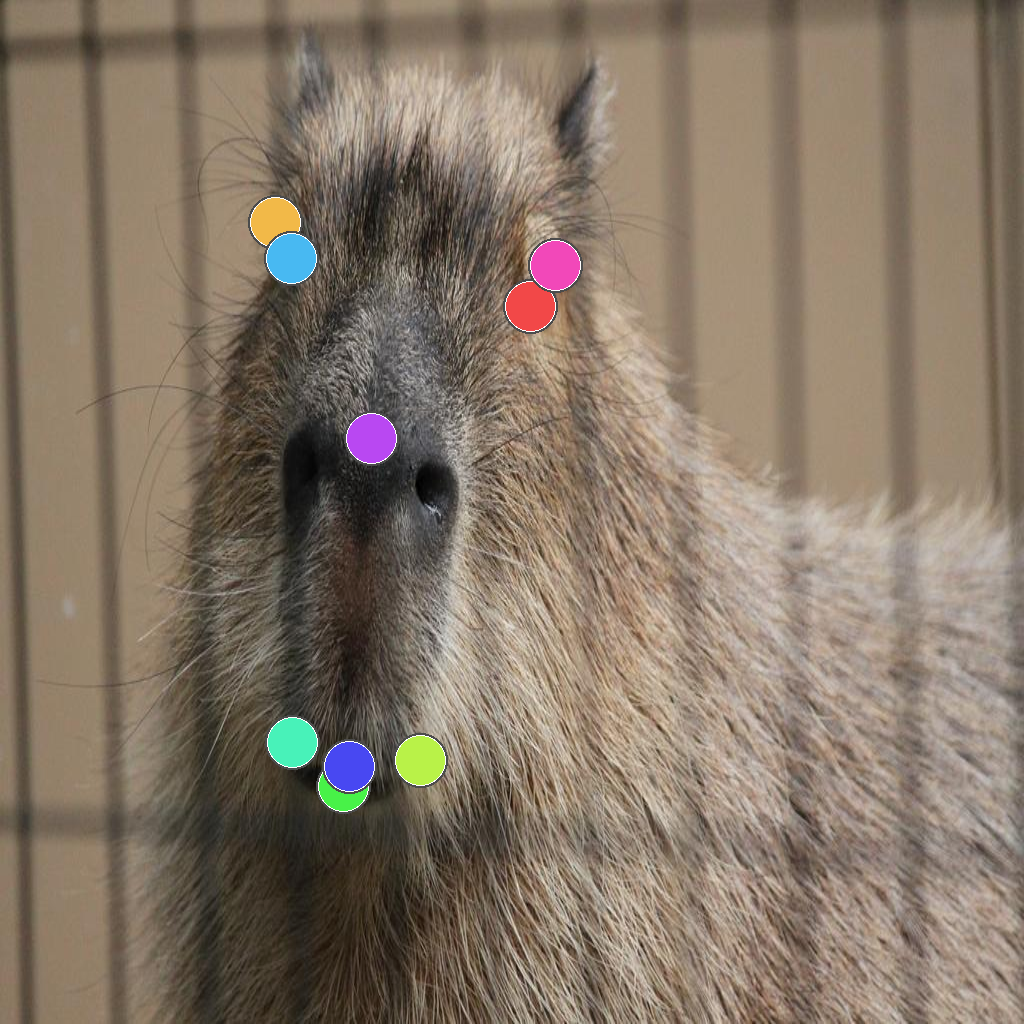} &
\includegraphics[width=0.19\linewidth]{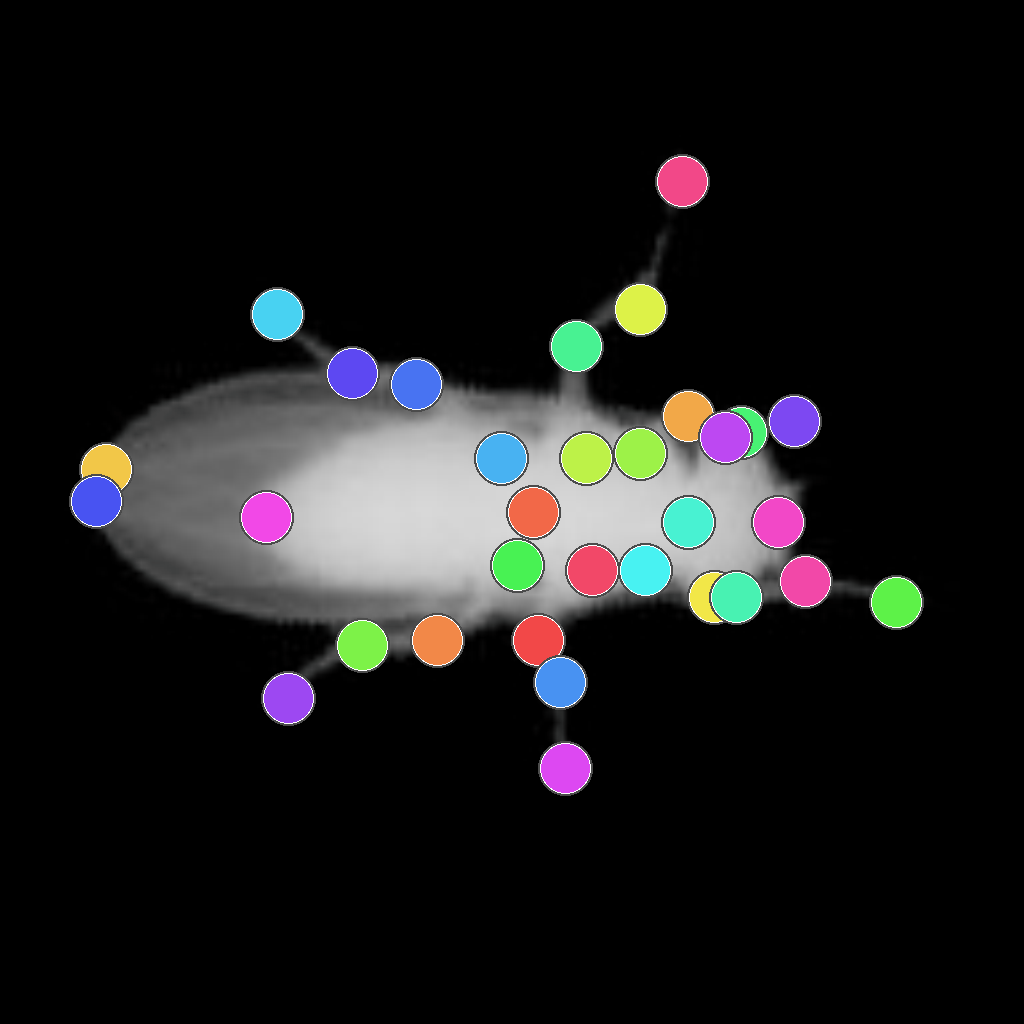} &
\includegraphics[width=0.19\linewidth]{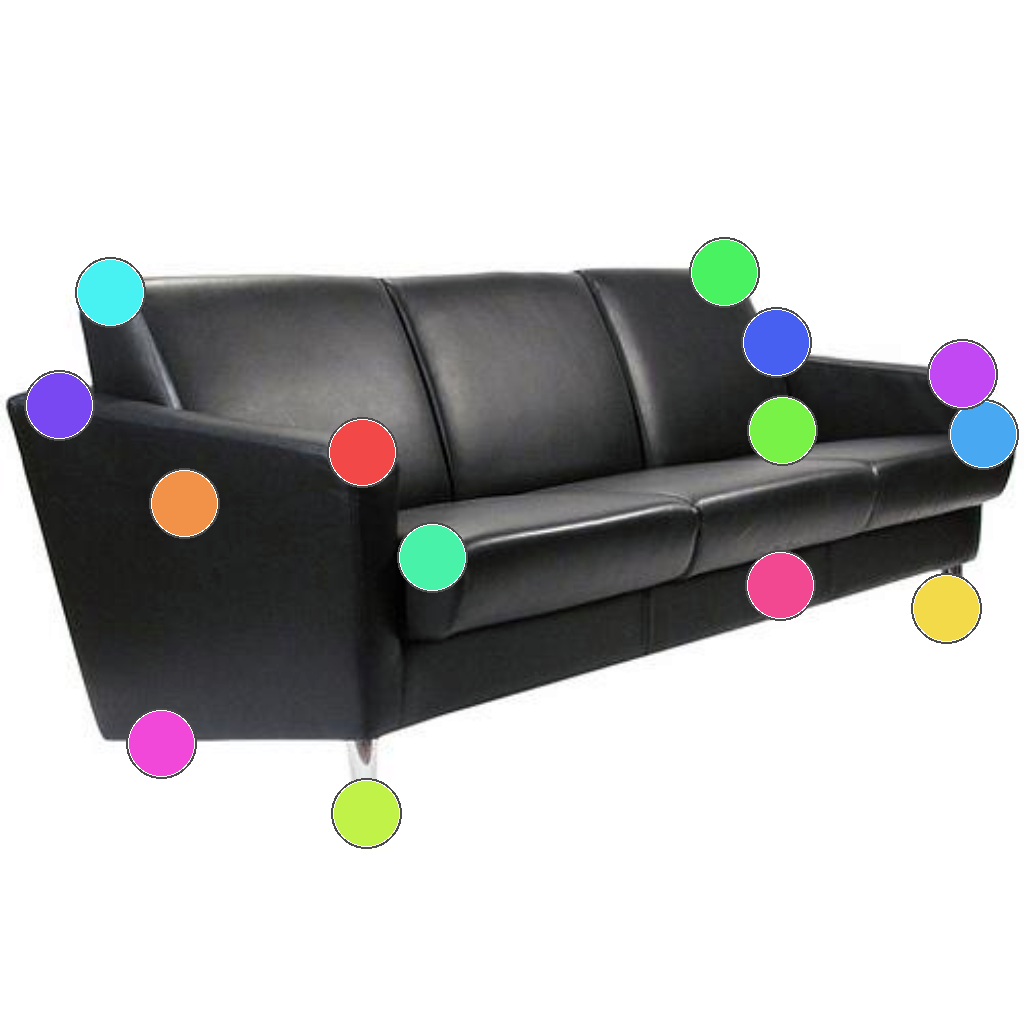} \\
\end{tabular}
\vspace{1pt}
\caption{\textbf{MP-100 samples used in our benchmark.}
Each column shows representative instances from the five macro-domains.}
\label{fig:mp100_examples}
\end{figure*}

\section{Pseudocode of Dense Self-Distillation}
\label{sec:pseudocode}
\begin{algorithm}[t]
\caption{\textbf{Dense self-distillation via flow anchoring}}
\label{alg:dense_self_distillation}
\small
\begin{algorithmic}[1]
\Require Source and target images $(\mathbf{I}^s,\mathbf{I}^t)$, sparse GT correspondences $\mathcal{E}$, student parameters $\theta_S$, teacher parameters $\theta_T$
\Ensure Self-distillation loss $\mathcal{L}_{\text{self}}$

\State \textbf{Teacher / student feature extraction}
\State $\mathbf{F}_T^s,\mathbf{F}_T^t \leftarrow \Phi_{\theta_T}(\mathbf{I}^s),\Phi_{\theta_T}(\mathbf{I}^t)$
\State $\mathbf{F}_S^s,\mathbf{F}_S^t \leftarrow \Phi_{\theta_S}(\mathbf{I}^s),\Phi_{\theta_S}(\mathbf{I}^t)$

\vspace{0.2em}
\State \textbf{Seed correspondence extraction}
\State Compute mutual nearest neighbors:
\[
\mathcal{P}_{\text{MNN}} =
\{(\mathbf{u},\mathbf{v}) \mid
\mathrm{NN}_{s\rightarrow t}(\mathbf{u})=\mathbf{v}
\wedge
\mathrm{NN}_{t\rightarrow s}(\mathbf{v})=\mathbf{u}\}
\]
\State Restrict $\mathcal{P}_{\text{MNN}}$ to pixels inside the object mask
\State Form seed correspondences: $\mathcal{P}_{\text{seed}} \leftarrow \mathcal{E} \cup \mathcal{P}_{\text{MNN}}$

\vspace{0.2em}
\State \textbf{Dense flow estimation}
\State Construct Delaunay triangulation $\mathcal{T}$ over source points $\mathcal{P}_{\text{seed}}$
\For{each triangle $\tau \in \mathcal{T}$}
    \State Define target triangle $\tau'$ from the matched vertices
    \State Estimate affine warp $\mathcal{W}_{\tau\rightarrow\tau'}$
\EndFor
\State Compose all triangle warps into piecewise-affine mapping $\hat{\mathcal{W}}$
\State Compute dense flow field \quad
$
\mathbf{D}(\mathbf{u}) = \hat{\mathcal{W}}(\mathbf{u}) - \mathbf{u}
$

\vspace{0.2em}
\State \textbf{Flow clustering and GT anchoring}
\State Cluster flow vectors $\mathbf{D}(\mathbf{u})$ using $k$-means
\State Merge clusters using BIC to obtain $\{\Omega_n\}$
\For{each cluster $\Omega_n$}
\State $C_n^s = \{\mathbf{u} \mid \mathbf{D}(\mathbf{u}) \in \Omega_n\}$
\State $C_n^t = \{\mathbf{u}+\mathbf{D}(\mathbf{u}) \mid \mathbf{u}\in C_n^s\}$
\EndFor
\State Retain clusters anchored by GT matches:
\[
\mathcal{P}_{\text{self}} =
\{(\mathbf{u},\mathbf{u}+\mathbf{D}(\mathbf{u})) \mid
\exists n,i:
\mathbf{u}\in C_n^s,
\mathbf{p}_i^s\in C_n^s,
\mathbf{p}_i^t\in C_n^t
\}
\]

\vspace{0.2em}
\State \textbf{Self-distillation loss}
\[
\mathcal{L}_{\text{self}} =
\frac{1}{|\mathcal{P}_{\text{self}}|}
\sum_{(\hat{\mathbf{u}},\hat{\mathbf{v}})\in\mathcal{P}_{\text{self}}}
\Bigl\|
\SoftArgmax_{\mathbf{u}\in\hat{\Lambda}} S(\hat{\mathbf{u}},\mathbf{u})
-
\hat{\mathbf{v}}
\Bigr\|_2^2
\]

\vspace{0.2em}
\State \textbf{EMA teacher update} \quad
$\theta_T \leftarrow \beta\theta_T + (1-\beta)\theta_S$

\State \Return $\mathcal{L}_{\text{self}}$
\end{algorithmic}
\end{algorithm}
For clarity, \Cref{alg:dense_self_distillation} summarizes the procedure used to generate dense pseudo-correspondences and compute the dense self-distillation loss described in \cref{sec:self_supervised}. 
\noindent The teacher network is maintained as an exponential moving average (EMA) of the student parameters and is used to generate stable pseudo-labels. Given a pair of images, the teacher features are used to mine reliable correspondences through mutual nearest-neighbor matches, which are combined with the available ground-truth keypoints. These correspondences are then \emph{densified} by estimating a piecewise-affine warp obtained from a Delaunay triangulation of the seed points, producing a dense flow field between the two images. Flow vectors are subsequently clustered to identify regions with coherent motion, and clusters consistent with the ground-truth correspondences are retained to form pseudo-labels. These pseudo-correspondences supervise the student network through a regression loss, while the teacher parameters are updated via EMA during training.

\section{Details on the MP-100 Benchmark}
\label{sec:benchmark_mp100}

Our goal is to establish an evaluation protocol to thoroughly assess the generalization ability of correspondence models beyond the specific landmarks and categories observed during training. Concurrently to our work, Jamais Vu \cite{Mariotti:2025:Jamais} highlights a similar limitation and augments SPair-71k with only four additional keypoint definitions per category. While this extension is useful, SPair-U remains limited both in the number of new keypoints and in its restriction to the original SPair-71k taxonomy.

To broaden this perspective, we turn to the ecosystem of 2D pose estimation \cite{Sun:2019:Pose}, where annotation schemes vary widely across object types. In particular, we repurpose the MP-100 dataset \cite{Xu:2022:MP100}, a large-scale collection spanning 100 categories and 18k images. For context, SPair-71k contains only 1.8k images in total. Similarly to us, Geo-SC \cite{Zhang:2024:Telling} adopts a pose estimation dataset, namely AP-10K, for semantic correspondence. However, their focus is primarily to provide a source of training, whereas our intent is to provide an evaluation benchmark to measure generalization under keypoint and category shift. We describe our curation protocol below.

\myparagraph{Data curation.} We first filter out images with fewer than three visible keypoints, accounting for occlusion or missing annotations. Next, to ensure a strict zero-shot setup for unseen-category evaluation, we remove all classes overlapping with the SPair-71k training categories, namely: \textit{cat body, sheep body, horse body, dog body, cow body, goldenretriever face, germanshepherddog face, bighornsheep face, przewalskihorse face, car, bus}. 
For categories that only partially overlap with SPair-71k but contribute a substantial number of novel keypoints, we retain them for the unseen-keypoints evaluation. In particular, the \textit{apparel items} super-category partially overlaps with the SPair-71k \textit{person} class, but provides many fine-grained garment-specific keypoint definitions that are entirely absent from SPair-71k. The same rationale applies to the \textit{human face} super-category, which offers a dense and diverse set of facial landmarks. After filtering, we group the categories into five domains, \ie, \textit{human face, apparel items, furniture, animal face, animal body}. We sample 2k image pairs within each domain using stratified sampling to avoid class imbalance.

\Cref{tab:spair_mp100_categories} summarizes the resulting benchmark splits. For acquisition details and annotation conventions, we refer readers to the original MP-100 paper \cite{Xu:2022:MP100}. To give an intuitive understanding of the visual diversity present across the domains used in our benchmark, we include representative samples in \cref{fig:mp100_examples}. The samples highlight the substantial variation in appearance, pose, texture, and structure across human faces, apparel items, animal species, and furniture categories, as well as the richness and heterogeneity of their keypoint definitions, which make this a challenging testbed for semantic correspondence.

\myparagraph{Unseen keypoints.}
This setting evaluates a model’s ability to generalize to \emph{new keypoint definitions} for object categories seen during training. For instance, \textit{human face} appears in SPair-71k as part of the broader \textit{person} class, but only with seven coarse keypoints (eyes, ears, nose, mouth, chin). In contrast, MP-100 provides a 68-point landmark definition capturing fine-grained facial geometry. A similar situation arises for apparel items: although clothing is implicitly present within the \textit{person} category in SPair-71k, MP-100 introduces rich keypoint annotations on garments (\eg., sleeve corners, skirt borders), with 282 novel keypoint definitions over 12 categories. These cases allow us to test fine-grained keypoint transfer under class-level familiarity.

\myparagraph{Unseen categories.}
Here, we evaluate generalization to object types for which \emph{no} keypoint annotations are available during training. We remove any category present in SPair-71k to enforce a strict zero-shot setting. The \textit{animal body} (32 categories) and \textit{animal face} (26 categories) domains include species entirely absent from SPair-71k. The \textit{home furniture} domain includes \textit{couch}, \textit{table}, \textit{bed}, and \textit{swivel chair}. Although \textit{swivel chair} may be loosely related to \textit{chair}, it does not appear in SPair-71k, and its keypoint definition (\eg, rotating base) differs meaningfully. We therefore choose to include it in the \textit{unseen-category} split.

\myparagraph{Split protocol.}
We organize the benchmark into five evaluation splits: two for the \textit{unseen-keypoint} scenario (human face and apparel items), and three for the \textit{unseen-category} scenario (animal body, animal face, and home furniture). This structure allows us to separately examine \emph{(i)} generalization to fine-grained keypoint definitions within familiar categories, and \emph{(ii)} generalization across entirely novel object types with no lexical or semantic overlap with the SPair-71k training set.

\subsection{Additional results on MP-100}
\Cref{tab:supp_mp100_generalization} extends the main paper evaluation by reporting the accuracy across PCK thresholds on all MP-100 splits. Overall, the trends observed at \pckTen{} remain stable at both finer (\pckFive{}) and coarser (\pckFifteen{}) thresholds. On \textit{unseen categories}, \ours{} remains the strongest method across all domains and all thresholds. Averaged across domains, it improves over the strongest prior method by +\SI{4.3}{\%} at \pckFive{}, +\SI{4.5}{\%} at \pckTen{}, and +\SI{5.4}{\%} at \pckFifteen{}. The gains are especially pronounced on \textit{Home furniture}, but remain consistent also on \textit{Animal body} and \textit{Animal face}, indicating robust transfer to categories never seen during training.
On \textit{unseen keypoints}, the comparison is more nuanced. For \textit{Human face}, the strongest competitor is the zero-shot DIFT baseline: \ours{} is slightly weaker at \pckFive{}, but becomes the best method at coarser thresholds, improving over DIFT by +\SI{0.2}{\%} at \pckTen{} and +\SI{1.4}{\%} at \pckFifteen{}. For \textit{Apparel items}, \ours{} consistently outperforms SD$+$DINO, with gains of +\SI{1.5}{\%}, +\SI{5.7}{\%}, and +\SI{9.0}{\%} at \pckFive{}, \pckTen{}, and \pckFifteen{}, respectively. Interestingly, several zero-shot methods remain highly competitive, and in some cases outperform supervised baselines. This supports our main observation: training with sparse keypoint annotations limits transfer to new landmark vocabularies. In contrast, \ours{} preserves the transferable structure of the pre-trained backbone, yielding more stable results across both unseen keypoints and unseen categories. Overall, these results reinforce MP-100 as a challenging and informative benchmark for assessing correspondence generalization.

\subsection{Per-category results on MP-100}

For completeness, \cref{tab:mp100_per_category_left,tab:mp100_per_category_right} report detailed per-category results on MP-100 using PCK@0.10. We compare \ours{} with representative prior approaches, namely DINOv2~\cite{Oquab:2023:Dinov2}, Geo-SC~\cite{Zhang:2024:Telling}, and Jamais Vu~\cite{Mariotti:2025:Jamais}. 
\Cref{tab:mp100_per_category_left} focuses on \textit{unseen categories}, while \cref{tab:mp100_per_category_right} reports results on \textit{unseen keypoints} within known categories. A clear pattern emerges on \textit{unseen categories}: \ours{} achieves the best accuracy on the large majority of classes, with especially strong gains on highly variable categories such as \textit{macaque} (+7.6 over Jamais Vu), \textit{rhino} (+6.0), \textit{cape buffalo} (+8.2), \textit{olive baboon} (+11.3), \textit{bed} (+14.7), and \textit{couch} (+9.0). These results indicate that the proposed dense self-distillation improves transfer not only across new animal species, but also to object families with very different geometry, such as home furniture. At the same time, the table also highlights genuinely difficult categories where all methods remain relatively close, or where a strong frozen foundation model remains competitive, such as \textit{table}, \textit{locust}, \textit{fly}, and \textit{polar bear}. This suggests that some categories are limited less by semantic transfer and more by intrinsic ambiguity, symmetry, or annotation difficulty. On \textit{unseen keypoints}, \ours{} is consistently best across all apparel categories and on \textit{human face}, often with large margins. In particular, the gains over Jamais Vu reach +10.0 on \textit{sling}, +14.6 on \textit{sling dress}, +11.1 on \textit{long sleeved shirt}, and +9.2 on \textit{short sleeved dress}, while remaining positive on all other categories. These improvements are notable because this setting introduces new landmark vocabularies within categories already seen during training, directly testing whether the model has learned dense semantic structure rather than memorizing the supervised keypoints. Interestingly, the unsupervised DINOv2 baseline is already fairly strong in some categories, especially \textit{human face}, confirming that foundation features contain substantial transferable structure; however, \ours{} consistently improves on top of this prior structure and yields the strongest overall generalization.

\begin{table*}[t]
  \caption{\textbf{Generalization on MP-100}~\cite{Xu:2022:MP100}. This table extends \cref{tab:mp100_generalization} of the main paper, reporting results across PCK thresholds on our proposed MP-100 benchmark. We evaluate methods  trained on SPair-71k, as well as zero shot baselines, indicated with $^*$. Supervised methods often fall short of zero-shot approaches, highlighting a generalization gap in existing approaches. In contrast, \ours{} maintains robust performances across domains, outside of the training distribution. Per-image PCK (in~\%, $\uparrow$), 
  best \textbf{bold}, second best \underline{underlined}.}
  \label{tab:supp_mp100_generalization}
  \centering
  \tablesize
  \vspace{-0.6em}
  \setlength{\tabcolsep}{5.5pt}
  \begin{tabularx}{\linewidth}{@{}Xccccccccccccccc@{}}
    \toprule
      &
      \multicolumn{6}{c}{\textbf{Unseen keypoints}} &
      \multicolumn{9}{c}{\textbf{Unseen categories}} \\
    \cmidrule(lr){2-7} \cmidrule(lr){8-16}
      & \multicolumn{3}{c}{\HumanFace}
      & \multicolumn{3}{c}{\Dress}
      & \multicolumn{3}{c}{\Elephant}
      & \multicolumn{3}{c}{\TableIcon}
      & \multicolumn{3}{c}{\AnimalFace} \\
    & \multicolumn{3}{c}{Human face}
      & \multicolumn{3}{c}{Apparel items}
      & \multicolumn{3}{c}{Animal body}
      & \multicolumn{3}{c}{Home furniture} & \multicolumn{3}{c}{Animal face} \\
    \cmidrule(lr){2-4} \cmidrule(lr){5-7}
    \cmidrule(lr){8-10} \cmidrule(lr){11-13} \cmidrule(lr){14-16}
      & 0.05 & 0.10 & 0.15 
      & 0.05 & 0.10 & 0.15 
      & 0.05 & 0.10 & 0.15 
      & 0.05 & 0.10 & 0.15 
      & 0.05 & 0.10 & 0.15  \\
    \midrule

DINOv2 $^*$~\cite{Oquab:2023:Dinov2} &
 41.8   &    66.2 &      76.9 &
 24.9   &    44.7 &      58.3 &
 22.6   &    36.1 &      46.2 &
 30.5   &    44.2 &      53.0 & 
 13.5   &    33.3 &      46.7
 \\

DIFT $^*$~\cite{Tang:2023:Dift} 
& 68.9 & 87.3 & 92.8 & 
29.3 & 48.2 & 59.0 & 
19.1 & 31.0 & 39.9 & 
34.9 & 46.9 & 54.1 & 
10.0 & 26.5 & 38.8 \\

SD + DINO $^*$~\cite{Zhang:2023:Tale} 
& 68.3   & 85.3  & 90.8 &
29.4  &   50.2    &  62.1 &
23.8   &  36.1    &  45.2 &
36.5   &  49.2   &   56.1 &
17.4   &  39.6    &  53.1\\

GECO \cite{Hartwig:2025:GECO}
  & 40.8 & 82.9 & 86.3
  & 23.4 & 41.7 & 56.6
  & 23.2 & 31.9 & 43.7
  & 37.9 & 48.1 & 58.2
  & 17.2 & 38.2 & 50.0 \\

Geo-SC~\cite{Zhang:2024:Telling} & 63.5 & 85.2 & 91.1 & 23.8 & 42.9 & 54.9 & 27.1 & 38.9 & 47.6 & 40.4 & 49.6 & 55.0 & 24.0 & 49.2 & 61.3 \\

 Jamais Vu~\cite{Mariotti:2025:Jamais} & 64.0 & 85.5 & 91.7 & 25.0 & 45.7 & 58.8 & 27.0 & 39.3 & 48.3 & 42.5 & 52.7 & 58.7 & 22.9 & 47.7 & 59.8 \\

\textbf{MARCO} \emph{(ours)} &  \textbf{64.3} & \textbf{87.5} & \textbf{94.2} & \textbf{30.9} & \textbf{55.9} & \textbf{71.1} & \textbf{29.9} & \textbf{42.3} & \textbf{51.5} & \textbf{48.8} & \textbf{60.4} & \textbf{66.9} & \textbf{26.6} & \textbf{52.6} & \textbf{64.7}  \\
    \bottomrule
  \end{tabularx}
  \vspace{-0.5em}
\end{table*}

\begin{figure*}[t]
  \centering
  \includegraphics[width=0.82\linewidth]{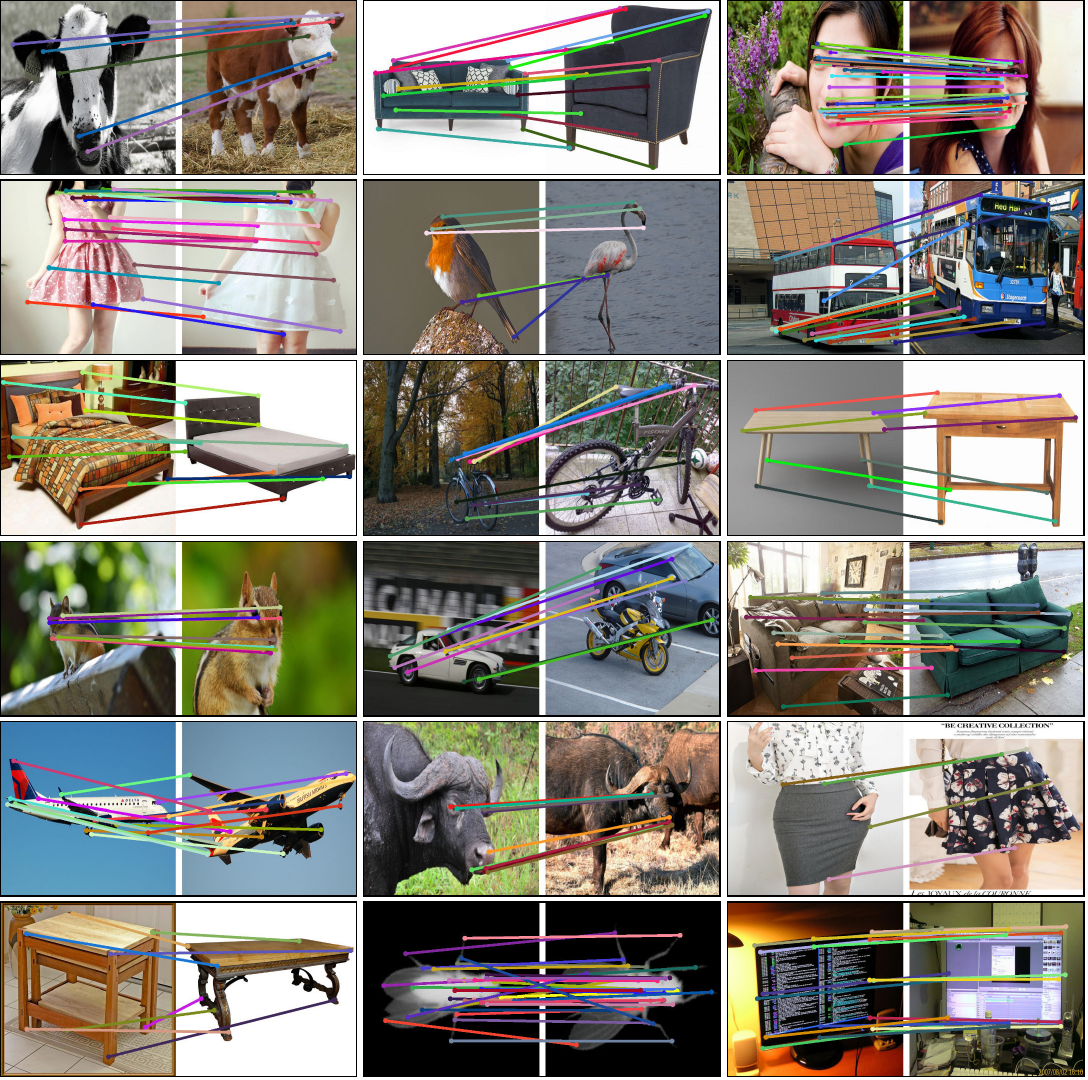}
    \vspace{-0.5em}
    \caption{\textbf{Qualitative results with \ours{}.} Examples from Spair-71k \cite{Min:2019:Spair} and MP-100 \cite{Xu:2022:MP100}.} 
    \label{fig:qualitatives_supp}
    \vspace{-0.5em}
\end{figure*}

\begin{table}[!t]
\caption{\textbf{Per-category evaluation on MP-100~\cite{Xu:2022:MP100}}: unseen categories. 
Per-image PCK@0.10 (in~\%, $\uparrow$).}
\label{tab:mp100_per_category_left}

\scriptsize
\setlength{\tabcolsep}{3pt}
\renewcommand{\arraystretch}{1.05}

\centering
\textbf{Unseen categories}

\vspace{0.3em}

\begin{tabularx}{\linewidth}{>{\raggedright\arraybackslash}p{0.09\linewidth} X cccc}
\toprule
\textbf{Domain} & \textbf{Category} & \textbf{DINOv2} & \textbf{Geo-SC} & \textbf{Jamais Vu} & \textbf{MARCO} \\
\midrule

Animal & macaque & 46.5 & \underline{53.6} & 54.3 & \textbf{61.9} \\
body & locust & 64.7 & \underline{70.4} & \textbf{71.1} & 65.8 \\
 & fly & 63.3 & \underline{70.8} & \textbf{73.4} & 66.8 \\
 & antelope & 37.9 & \underline{42.8} & 41.7 & \textbf{48.1} \\
 & cheetah & 36.8 & 40.3 & \underline{41.2} & \textbf{44.0} \\
 & fox & 46.0 & 50.5 & \underline{51.3} & \textbf{55.7} \\
 & leopard & 32.4 & 35.5 & \underline{37.8} & \textbf{39.4} \\
 & panther & 40.6 & \underline{41.1} & 40.0 & \textbf{46.8} \\
 & rat & 26.5 & 26.5 & \underline{27.3} & \textbf{29.1} \\
 & squirrel & 35.2 & \underline{41.4} & \underline{41.4} & \textbf{44.9} \\
 & beaver & \underline{29.8} & 21.6 & 19.4 & \textbf{28.9} \\
 & deer & 41.2 & 48.0 & \underline{50.1} & \textbf{53.7} \\
 & giraffe & 31.1 & \textbf{35.3} & 34.6 & \underline{35.2} \\
 & lion & 37.8 & 42.6 & \underline{43.4} & \textbf{47.9} \\
 & pig & 19.8 & \underline{22.4} & 22.3 & \textbf{26.2} \\
 & rhino & 44.3 & 51.7 & \underline{52.8} & \textbf{58.8} \\
 & weasel & 42.6 & 44.3 & \underline{45.4} & \textbf{50.7} \\
 & bison & 31.0 & 36.0 & \underline{37.2} & \textbf{40.0} \\
 & elephant & 23.4 & 25.5 & \underline{25.6} & \textbf{31.4} \\
 & gorilla & 33.4 & 31.8 & 32.0 & \textbf{36.1} \\
 & otter & 33.5 & \underline{34.0} & 32.9 & \textbf{35.9} \\
 & polar bear & 31.3 & \underline{35.8} & \underline{35.8} & \textbf{33.1} \\
 & skunk & 34.3 & 31.0 & \underline{34.5} & \textbf{38.6} \\
 & wolf & 35.4 & \underline{41.1} & 40.0 & \textbf{42.1} \\
 & hippo & 26.3 & 28.1 & \underline{28.3} & \textbf{30.5} \\
 & bobcat & 32.5 & 35.9 & \underline{36.8} & \textbf{40.6} \\
 & raccoon & 34.3 & 35.0 & \textbf{38.2} & \underline{37.5} \\
 & hamster & 38.2 & 40.5 & \underline{41.2} & \textbf{45.6} \\
 & panda & 24.1 & 25.0 & \underline{25.9} & \textbf{27.6} \\
 & rabbit & 39.2 & \underline{41.1} & 39.3 & \textbf{42.4} \\
 & spider monkey & 31.6 & 30.0 & 28.8 & \textbf{33.8} \\
 & zebra & 33.2 & 36.4 & \underline{37.4} & \textbf{37.9} \\

\midrule

Animal & alpaca & 22.1 & 28.9 & \underline{29.9} & \textbf{34.4} \\
face & californian sea lion & 23.7 & 30.3 & \underline{30.8} & \textbf{40.7} \\
 & chipmunk & 39.3 & \textbf{61.7} & \underline{56.9} & 60.0 \\
 & ferret & 38.7 & \textbf{69.8} & \underline{69.3} & 68.0 \\
 & gibbons & 23.7 & 44.1 & \underline{45.9} & \textbf{51.5} \\
 & guanaco & 21.0 & 26.8 & \underline{27.5} & \textbf{36.8} \\
 & proboscis monkey & 25.3 & \underline{40.2} & 39.8 & \textbf{48.1} \\
 & arctic wolf & 43.2 & 63.3 & \underline{63.4} & \textbf{65.8} \\
 & camel & 24.2 & 32.1 & \underline{31.6} & \textbf{39.0} \\
 & common warthog & 44.1 & \textbf{61.1} & \underline{55.4} & 58.3 \\
 & gentoo penguin & 28.2 & 33.3 & \underline{34.0} & \textbf{38.3} \\
 & grey seal & 32.1 & \underline{49.3} & 47.6 & \textbf{51.8} \\
 & klipspringer & 38.7 & \underline{49.1} & 47.9 & \textbf{53.8} \\
 & fennec fox & 37.1 & \textbf{63.7} & \underline{61.3} & 62.8 \\
 & blackbuck & 17.5 & \underline{29.9} & 29.1 & \textbf{34.8} \\
 & cape buffalo & 34.3 & \underline{53.2} & \underline{53.2} & \textbf{61.4} \\
 & dassie & 36.1 & \textbf{58.7} & \underline{55.2} & 57.2 \\
 & gerbil & 31.4 & \textbf{51.5} & 46.6 & \underline{51.1} \\
 & grizzly bear & 44.0 & \underline{57.8} & 56.4 & \textbf{60.4} \\
 & olive baboon & 39.4 & 48.3 & \underline{49.8} & \textbf{61.1} \\
 & quokka & 34.6 & \underline{51.6} & 48.2 & \textbf{56.5} \\
 & bonobo & 38.1 & \textbf{63.0} & \underline{61.5} & 62.7 \\
 & capybara & 36.3 & \underline{45.0} & 43.5 & \textbf{49.1} \\
 & fallow deer & 29.4 & \underline{42.6} & 38.6 & \textbf{44.4} \\
 & onager & 44.7 & \underline{53.0} & 50.2 & \textbf{53.5} \\
 & pademelon & 41.5 & \textbf{70.1} & \underline{68.1} & 67.5 \\

\midrule

Home & couch & 61.1 & 61.7 & \underline{69.1} & \textbf{78.1} \\
furniture & table & \underline{24.0} & \textbf{25.9} & \textbf{25.9} & 23.9 \\
 & bed & 37.7 & 44.9 & \underline{47.0} & \textbf{61.7} \\
 & swivel chair & 52.9 & 65.2 & \underline{67.3} & \textbf{75.8} \\
\bottomrule
\end{tabularx}
\vspace{-0.5em}
\end{table}

\begin{table}[t]
\caption{\textbf{Per-category evaluation on MP-100~\cite{Xu:2022:MP100}}: unseen keypoints. Per-image PCK@0.10 (in~\%, $\uparrow$).}
\label{tab:mp100_per_category_right}

\scriptsize
\setlength{\tabcolsep}{3pt}
\renewcommand{\arraystretch}{1.05}

\centering
\textbf{Unseen keypoints}

\vspace{0.3em}

\begin{tabularx}{\linewidth}{>{\raggedright\arraybackslash}p{0.12\linewidth} X cccc}
\toprule
\textbf{Domain} & \textbf{Category} & \textbf{DINOv2} & \textbf{Geo-SC} & \textbf{Jamais Vu} & \textbf{MARCO} \\
\midrule

Apparel & short sleeved outwear & 46.5 & 45.3 & \underline{48.2} & \textbf{58.9} \\
item & short sleeved shirt & 48.8 & 49.6 & \underline{52.7} & \textbf{61.3} \\
 & skirt & \underline{40.0} & 32.6 & 34.7 & \textbf{43.9} \\
 & short sleeved dress & 48.2 & 46.8 & \underline{50.6} & \textbf{60.0} \\
 & vest dress & 54.4 & 56.2 & \underline{60.7} & \textbf{69.9} \\
 & long sleeved dress & 43.7 & 42.2 & \underline{45.8} & \textbf{55.0} \\
 & long sleeved outwear & 42.5 & 42.4 & \underline{46.1} & \textbf{52.6} \\
 & long sleeved shirt & 42.5 & 41.9 & \underline{43.9} & \textbf{55.0} \\
 & sling & 42.6 & 33.7 & 33.8 & \textbf{51.5} \\
 & sling dress & \underline{48.6} & 45.9 & \underline{48.6} & \textbf{63.2} \\
 & trousers & 35.3 & 37.4 & \underline{39.4} & \textbf{44.7} \\
 & vest & 42.5 & 40.4 & \underline{43.4} & \textbf{52.6} \\

\midrule

Human face & human & 66.2 & \underline{85.2} & \underline{85.5} & \textbf{87.5} \\

\bottomrule
\end{tabularx}

\vspace{-0.5em}

\end{table}

\section{Qualitative Examples}
\label{sec:qualitatives}

To complement the quantitative analyses, \cref{fig:qualitatives_supp} provides a series of qualitative visualizations, illustrating the behavior of \ours{} across a wide range of settings. 
We include examples from \textbf{SPair-71k}, highlighting the model’s ability to produce accurate and spatially coherent correspondences under significant appearance changes, occlusions, and viewpoint variations.  
We further showcase results on our \textbf{MP-100 benchmark}, focusing on both the \emph{unseen-keypoint} and \emph{unseen-category} regimes. 
These examples demonstrate how \ours{} adapts to novel landmark definitions it has never been trained on, and how its predictions remain stable even for novel object types with distinct shapes and geometries. 
Overall, the qualitative results visually confirm the strong generalization capabilities encouraged by our dense self-distillation framework.

\end{document}